\documentclass[a4paper,USenglish]{tgdk-v2021}
\nolinenumbers

\newcommand{\concept}[1]{\texttt{#1}}

\newcommand{\myurl}[1]{\footnote{\url{#1}}}
\usepackage{todonotes}
\usepackage{bigfoot}

\author{Ansgar Scherp}{Ulm University, Germany \and \url{http://ansgarscherp.net}}{ansgar.scherp@uni-ulm.de}{https://orcid.org/0000-0002-2653-9245}{}

\author{Gerd Groener}{Carl Zeiss SMT GmbH, Germany \and
\url{http://www.gerd-groener.de}}{gerd.groener@gmx.de}{https://orcid.org/0009-0002-0259-9769}{}{}

\author{Petr \v{S}koda}{Department of Software Engineering, Faculty of Mathematics and Physics, Charles University, Prague, Czechia}{petr.skoda@matfyz.cuni.cz}{https://orcid.org/0000-0002-2732-9370}{}

\author{Katja Hose}{TU Wien, Austria \and \url{http://www.katja-hose.de}}{katja.hose@tuwien.ac.at} {https://orcid.org/0000-0001-7025-8099}{}

\author{Maria-Esther Vidal}{Leibniz University of Hannover and TIB-Leibniz Information Centre for Science and Technology, Germany}{maria.vidal@tib.eu}{https://orcid.org/0000-0003-1160-8727}{Partially funded by Leibniz Association, program "Leibniz Best Minds: Programme for Women Professors", project TrustKG-Transforming Data in Trustable Insights; Grant P99/2020}

\authorrunning{A. Scherp et\,al.}

\Copyright{Ansgar Scherp and Gerd Groener and Petr \v{S}koda and Katja Hose and Maria-Esther Vidal}



\SeriesVolume{2}
\SeriesIssue{1}
\SectionAreaEditor{Aidan Hogan, Ian Horrocks, Andreas Hotho, and Lalana Kagal}
\SpecialIssue{Trends in Graph Data and Knowledge -- Part 2}
\DateSubmission{~}
\DateAcceptance{~}
\DatePublished{\today}

\ArticleNo{2}

\category{Survey}

\relatedversion{Lifelong version}\relatedversiondetails{Full Version}{https://github.com/ascherp/semantic-web-primer}

\begin{document}
\title{Semantic Web: Past, Present, and Future}
\titlerunning{Semantic Web: Past, Present, and Future}
\subtitle{With ``Machine Learning on Knowledge Graphs'' and ``Language Models and Knowledge Graphs'', 2025-11-11, Original: TGDK 2(1): 3:1-3:37 (2024)}
\maketitle

\begin{abstract}

Ever since the vision was formulated, the Semantic Web has inspired many generations of innovations.
Semantic technologies have been used to share vast amounts of information on the Web, enhance them with semantics to give them meaning, and enable inference and reasoning on them. 
Throughout the years, semantic technologies, and in particular knowledge graphs, have been used in search engines, data integration, enterprise settings, and machine learning.

In this paper, we recap the classical concepts and foundations of the Semantic Web as well as modern and recent concepts and applications, building upon these foundations. 
The classical topics we cover include knowledge representation, creating and validating knowledge on the Web, reasoning and linking, and distributed querying.
We enhance this classical view of the so-called ``Semantic Web Layer Cake'' with an update of recent concepts.
These include provenance, security and trust, as well as a discussion of practical impacts from industry-led contributions.
We also provide an overiew of shallow and deep machine learning methods for knowledge graphs and discuss the relation of language models and knowledge graphs.
We conclude with an outlook on the future directions of the Semantic Web.

\end{abstract}

\clearpage

\section*{Changelog}

\begin{itemize}
\item \textbf{Update in November 2025}: Added an overview of \textit{machine learning methods for knowledge graphs} in Section~\ref{sec:machine-learning-on-knowledge-graphs}, particularly shallow graph embeddings and neural networks for graphs.
We discuss \textit{language models} and their relation to and use for knowledge graphs in Section~\ref{sec:language-models-and-knowledge-graphs}.
\end{itemize}

\section*{Preamble}

This article is a living document.
The first version was written by Jannik, Scherp, and Staab in 2011~\cite{DBLP:journals/insk/JanikSS11}.
It was translated to German and updated by Gröner, Scherp, and Staab in 2013~\cite{DBLP:books/ol/13/GronerSS13} and updated again by Gröner and Scherp in 2020~\cite{DBLP:books/degruyter/20/ScherpG20}.
This release in 2024 reflects on the latest developments in knowledge graphs and large language models.
It has been translated back to English and extended with contributions by Petr \v{S}koda, Katja Hose, and Maria-Esther Vidal.	

\textbf{If you like to contribute}, please contact the first author and visit: \url{https://github.com/ascherp/semantic-web-primer}

\textbf{Please cite this paper as}, see \url{https://dblp.org/rec/journals/tgdk/ScherpG0HV24.html?view=bibtex}:

\begin{verbatim}
	@article{DBLP:journals/tgdk/ScherpG0HV24,
		author = {Ansgar Scherp and Gerd Gr{\"{o}}ner and
			Petr Skoda and Katja Hose and
			Maria{-}Esther Vidal},
		title        = {Semantic Web: Past, Present, and Future},
		journal      = {{TGDK}},
		volume       = {2},
		number       = {1},
		pages        = {3:1--3:37},
		year         = {2024},
		url          = {https://doi.org/10.4230/TGDK.2.1.3},
		doi          = {10.4230/TGDK.2.1.3},
	}
	
\end{verbatim}

\vfill

\clearpage
\tableofcontents
\clearpage

\section{Introduction}

The vision of the Semantic Web as coined by Tim Berners-Lee, James Hendler, and Orla Lassila~\cite{TimBernersLee2001} in 2001 is to develop intelligent agents that can automatically gather semantic information from distributed sources accessible over the Web, integrate that knowledge, use automated reasoning~\cite{DBLP:journals/ki/GlimmS16a}, and solve complex tasks as such as schedule appointments in negotiation of the preferences of the involved parties.
We have come a long way since then.
In this paper, we reflect on the \textit{past}, i.\,e., the ideas and components developed in the early days of the Semantic Web.
Since the beginning, the Semantic Web has tremendously developed and undergone multiple waves of innovation.
The Linked Data movement has especially seen uptake by industries, governments, and non-profit organizations, alike.
We discuss those \textit{present} components and concepts that have been added over the years and shown to be very useful.
Although many concepts of the initial idea of the Semantic Web have been implemented and put into practice, still further research is needed to reach the full vision.
Thus, this paper concludes with an outlook to \textit{future} directions and steps that may be taken.

For the novice reader of the Semantic Web, we provide a brief historical overview of the developments and innovation waves of the Semantic Web:
At the beginning of the Semantic Web, we were mainly talking about publishing Linked Data on the Web~\cite{DBLP:books/crc/linked2014}, i.\,e., semantic data typically structured using the Resource Description Framework (RDF)\footnote{\url{http://www.w3.org/TR/rdf-primer/}} that is accessible on the Web using URIs/IRIs to identify entities, classes, predicates, etc.
By referencing entities from other websites and Web-accessible sources, i.\,e., dereferencable via HTTP, the data becomes naturally linked. 
By using standardized vocabularies and ontologies the information then becomes more aligned and easier to use across sources. 
These principles have allowed non-profit organizations, companies, governments, and individuals to publish and share large amounts of interlinked data, which has led to the success of the Linked Open Data cloud\footnote{\url{https://lod-cloud.net/}} since 2007. 
Since many of the large interconnected semantic sources are accessible via interfaces understanding structured query languages (SPARQL endpoints), federated query processing methods were developed that allow exploiting the strengths of structured query languages to precisely formulate an information need and optimize the query for efficient execution in a distributed setting.

When Google launched its Knowledge Graph in 2012\footnote{\url{https://blog.google/products/search/introducing-knowledge-graph-things-not/}}, semantic technologies experienced another wave of new applications in the context of searching information.
Whereas search engines before mainly relied on keyword search and string-based matches of the keywords in the websites' text, the knowledge graph enabled including semantics to capture the user's information need as well as the meaning of potentially relevant documents. 
To achieve this purpose, Google's knowledge graph integrates large amounts of machine-processable data available on the Web and uses this information not only to improve search results but also to display infoboxes for entities identified in the user's keywords. 
It is only since 2012 that we have widely used the term ``knowledge graph'' to refer to semantic data, where entities are connected via relationships and form large graphs of interconnected information, typically with RDF as a common standard language.
In recent years though (labeled) property graphs (LPG) have been used to manage knowledge graphs.
We refer to the literature for a detailed comparison of RDF graphs and LPGs~\cite{DBLP:journals/corr/abs-2003-02320} and also like to point out that they can be converted into each other~\cite{DBLP:phd/dnb/Blume22}.
In this article, we consider knowledge graphs from the perspective of the Semantic Web, i.\,e., we consider RDF graphs.

A few years later, semantic technologies found another novel application in enterprise settings as enterprise knowledge graphs \cite{DBLP:conf/iceis/GalkinAVS17,Noy19} and data integration \cite{Cheatham2017,DBLP:books/sp/10/Wood10}. 
Since all kinds of information can be structured as a graph, knowledge graphs can be used as a common structure to integrate heterogeneous information that is otherwise locked up in silos in different branches of a company.
Integrating the data into a knowledge graph then allows for an integrated view, efficiently retrieving relevant information from this view once needed, and integrating external information that is available in the form of knowledge graphs, for instance on the Semantic Web. 
After all, online analytical processing-style queries can also be formulated with SPARQL\footnote{\url{http://www.w3.org/TR/sparql11-query/}} and evaluated on (distributed) knowledge graphs.

In the past few years, learning on graph data became one of the fastest growing and most active areas in machine learning~\cite{DBLP:series/synthesis/2020Hamilton}.
Graph representation learning has created a new wave of graph embedding models and graph neural networks on knowledge graphs for tasks such as entity classification and link prediction.
Natural Language Processing (NLP) has been another important field of the Semantic Web since the early years to extract knowledge from textual data and make it machine readable. 
Another field where NLP meets the Semantic Web is user interfaces for search on structured data to enable intuitive, natural language querying for graph data~\cite{DBLP:journals/eswa/HabernalK13} similar to web search engines.
At the end of 2022, ChatGPT\footnote{\url{https://chat.openai.com/}} emerged as the first publicly available end-consumer tool based on a Large Language Model (LLM).
Since then, the GPT-based family of LLMs has stirred up research and business alike and demonstrated impressive performance on many NLP tasks, including generating structured queries from user prompts and extracting structured knowledge from text~\cite{DBLP:journals/corr/abs-2305-04676}.

The capabilities of tools like Bing Chat\myurl{http://bing.com} with underlying access to the World Wide Web are reminding one of the intelligent agents that were envisioned 20 years before.
For example, at the time of writing, the GPT4-based tool Bing\footnote{\url{https://www.bing.com/}} can internally generate SPARQL queries and execute them, while the structured response is seamlessly embedded into its natural language outputs to the users.\footnote{Based on a sequence of prompts ran on January 22, 2024, using the GPT-4 model provided on the Bing Chat mobile app.
The prompt sequence is: ``do you have access to DBpedia'', ``how do you access DBpedia'', ``please give me an example where you access DBpedia in response''.}
While it already addresses some of the early visions of the Semantic Web, particularly the complex planning and reasoning capabilities of LLMs are -- due to their nature of focusing on generating and processing text -- still limited.
We hypothesize that advances in neuro-symbolic AI and semantic technologies will be key for improving LLMs and bringing generative AI tools like Bing and the Semantic Web further together.
We are keen to witness this next era of the Semantic Web.

In this paper, we provide a comprehensive overview of the Semantic Web with its semantic technologies and underlying principles that have been inspiring and driving the multiple waves of innovations in the past two decades.
Section~\ref{sec:musicscenario} provides a motivating example for the classic Semantic Web. 
We refer back to this example throughout the paper.
Section~\ref{sec:swarchitecture} presents the principles and the general architecture of the Semantic Web along with the basic semantic technologies it is founded upon. 
Besides classical components, we are also describing recent developments, and pointing out components that are still being researched and developed.
Section~\ref{sec:representation} shows how to represent distributed knowledge on the Semantic Web.
The creation and maintenance of graph data is described in Section~\ref{sec:creation}.
Section~\ref{sec:reasoning} discusses the principle of reasoning and logical inference.
Section~\ref{sec:distributeddataquerying} then shows how to query over the (distributed) graph data on the Semantic Web.
We discuss the trustworthiness and provenance of data on the Semantic Web in Section~\ref{sec:trustprovenance}.
Machine learning methods for knowledge graphs are discussed in Section~\ref{sec:machine-learning-on-knowledge-graphs}, particularly shallow graph embeddings and neural networks for graphs.
We discuss language models and their relation and use for knowledge graphs in Section~\ref{sec:language-models-and-knowledge-graphs}.
We provide extensive examples of applications based on and using the Semantic Web and its technologies in Section~\ref{sec:examples}.
Finally, we reflect on the impact the Semantic Web has on practitioners in Section~\ref{sec:impact}.
Finally, we conclude with a brief outlook on future developments for the Semantic Web.

\section{Motivating Example}
\label{sec:musicscenario}

On the Semantic Web, knowledge components from different sources can be intelligently integrated with each other.
As a result, complex questions can be answered, questions like ``What types of music are played on British radio stations? At which time and day of the week?'' or ``Which radio station plays songs by Swedish artists?''
In this section, we provide an overview of how the Semantic Web can be employed to answer those questions. We provide details of the components of the Semantic Web in the following sections.

\begin{figure}[ht]
  \centering
   \includegraphics[scale=0.7]{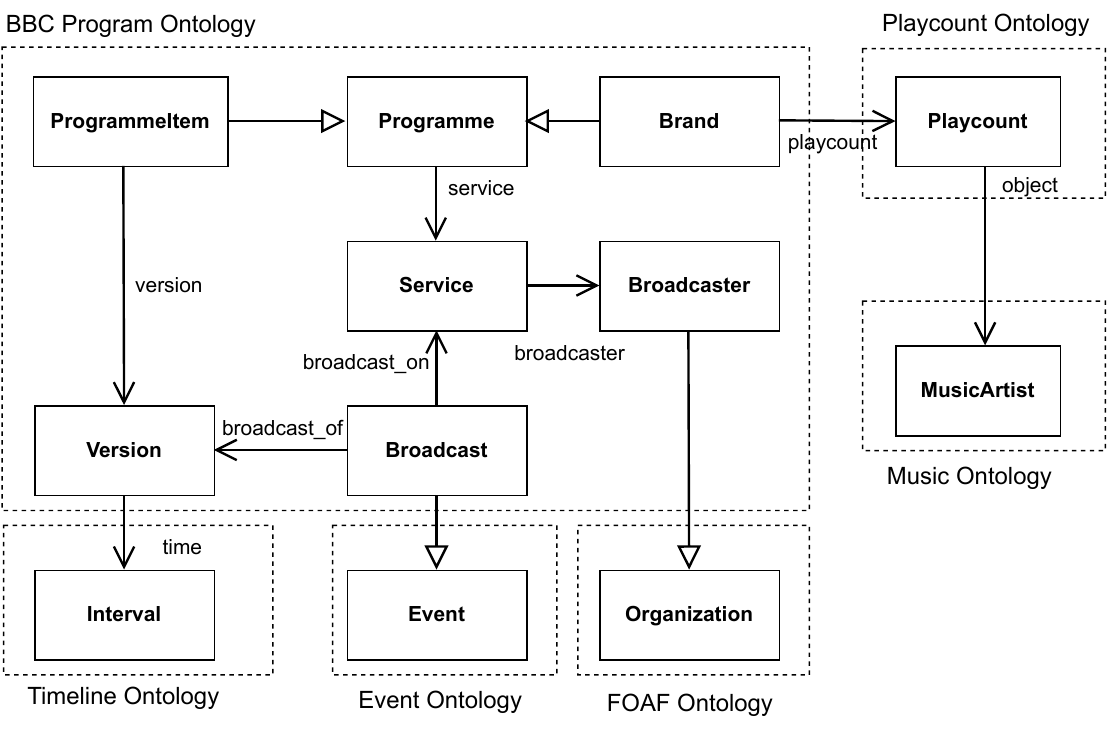}
   \caption{Example of the BBC ontology with links to other ontologies (notation based on UML, here without prefix for namespaces).}
   \label{fig:bbc-ontology-example}
\end{figure}

We consider the example of the BBC program ontology with links to various other ontologies such as for music, events, and social networks as shown in Figure~\ref{fig:bbc-ontology-example}.
We start with the BBC playlists of its radio stations.
The playlists are published online in Semantic Web formats.
We can leverage the playlist to get unique identifiers of played artists and bands. For example, the music group ``ABBA'' has a unique identifier in the form of a URI (\url{https://www.bbc.co.uk/programmes/b03lyzpr}). 
This URI can be used to link the music group to information from the MusicBrainz\myurl{http://musicbrainz.org/} music portal.
MusicBrainz knows the members of the band, such as Benny Andersson, as well as the genre and songs.
In addition, MusicBrainz is linked to Wikipedia\myurl{https://www.wikipedia.org/} (not shown in the figure), e.\,g., to provide information about artists, such as biographies on DBpedia~\cite{key:dbpedia}.
Information about British radio stations can be found in the form of lists on Web pages such as Radio UK\myurl{https://www.radio-uk.co.uk}, which can also be converted into a representation in the Semantic Web.

We can see that the required information is distributed across multiple knowledge components, e.\,g., BBC Program, MusicBrainz, and others.
Each knowledge component can in principle provide different access to the data and utilize various ways to describe the data.
Consequently, to answer the questions the data must be integrated.
On the Semantic Web, data integration relies on ontologies describing data and the meaning of relations in data.

Colloquially, an ontology is a description of concepts and their relationships.
Ontologies are used to formally represent knowledge on the Semantic Web.\footnote{An ontology definition is provided in Section~\ref{sec:ontologies:definition}.}
For example, Dublin Core\myurl{https://www.dublincore.org/specifications/dublin-core/dc-rdf/} provides a metadata schema for describing common properties of objects, such as the creator of the information, type, date, title, usage rights, and so on.
Figure~\ref{fig:bbc-ontology-example} presents ontologies used to describe data in our example.
For example, the Playcount ontology\myurl{http://dbtune.org/bbc/playcount/} of the BBC is used to model which artist was played and how many times in the programs.
Ontologies can be interconnected in the Semantic Web.
For example, the MusicBrainz ontology is connected to the BBC ontology using the Playcount ontology.
Different ontologies with varying degrees of formality and different relationships to each other are used by the BBC to describe their data (see also~\cite{Raimond2009IMR}).

As all this data is available and interconnected by ontologies, a user of the Semantic Web can directly ask for answers to questions in this and other scenarios.
To make this possible, the Semantic Web requires generic software components, languages, and protocols that can interact seamlessly with each other.
We introduce the classical and modern components of the architecture of the Semantic Web in Section~\ref{sec:swarchitecture}.

In addition to the above example, the Semantic Web can be used for a variety of other applications (see examples in Section~\ref{sec:userinterfaces}).
Apart from technical aspects, the Semantic Web should also be understood as a socio-political phenomenon.
Similar to the World Wide Web, various individuals and organizations publish their data on the Semantic Web and collaborate to link and improve this data.
This impact on practitioners is discussed in Section~\ref{sec:impact}.

\section{Architecture of the Semantic Web}
\label{sec:swarchitecture}

The example in Section~\ref{sec:musicscenario} describes \emph{what} the Semantic Web is as an infrastructure, but not \emph{how} this is achieved.
In fact, the capabilities of the Semantic Web \emph{in a small scale} have already been implemented by some knowledge-based systems originating from artificial intelligence research, e.\,g., Heinsohn et al~\cite{DBLP:journals/ai/HeinsohnKNP94}.
However, for the implementation of the vision \emph{on a large scale}, i.\,e, the Web, these knowledge-based systems lacked flexibility, robustness, and scalability.
In part, this was due to the complexity of the algorithms used.
For example, knowledge bases in description logic in the 1990s, which serve as the basis of Web ontologies, were limited regarding their size such that they could handle at the most some hundred concepts~\cite{DBLP:journals/ai/HeinsohnKNP94}.

In the meantime, enormous improvements have been achieved.
Greatly increased computational power and optimized algorithms allow a practical handling of large ontologies like 
Simple Knowledge Organization System (SKOS)\myurl{https://www.w3.org/TR/skos-reference/}, 
Gene Ontology\myurl{https://geneontology.org/}, Schema.org, and SNOMED-CT\myurl{https://www.snomed.org/}.
However, there are some fundamental differences between traditional knowledge-based systems and the Semantic Web.
Data management in traditional knowledge-based systems has weaknesses in terms of handling large amounts of data and data sources, among other things because of 
different underlying formalisms,
distributed locations, 
different authorities,
different data quality, and 
a high frequency of change in the data used.

The Semantic Web applies fundamental principles to deal with these problems; they represent the basis for the architecture of the Semantic Web.
This architecture's building blocks can roughly be categorized into groups, covering the entire life cycle of handling and managing graph data on the Web.
These groups are
graph data representation,
creation and validation of graph data,
reasoning over and linking of graph data,
(distributed) querying of graph data, 
crypto, provenance, and trustworthiness of graph data, and
user interfaces and applications.

\begin{figure}[ht]
  \centering
 \includegraphics[width=0.75\textwidth]{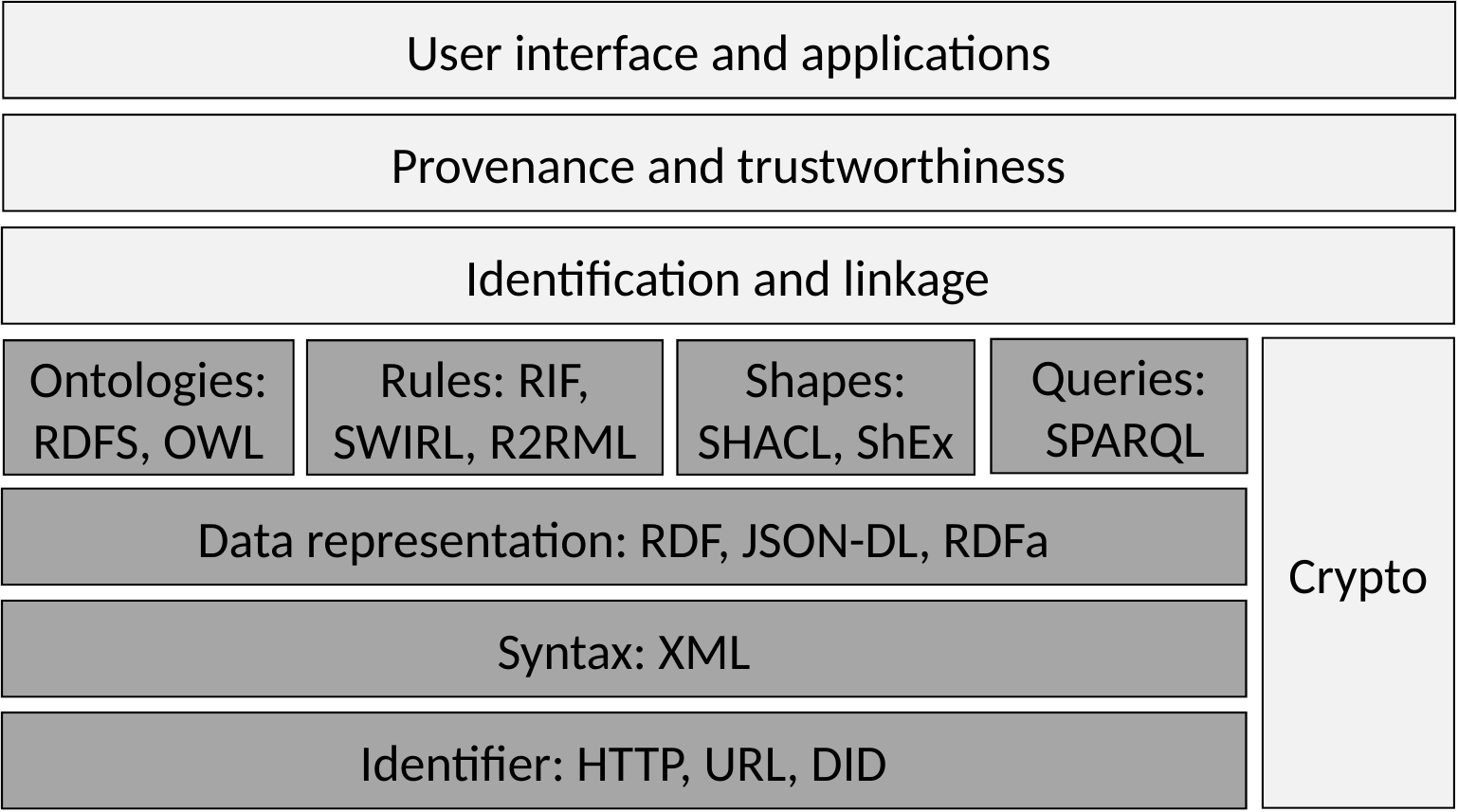}
 \caption{Representation of the components of the so-called ``Semantic Web Layer Cake''. W3C language standards are shown in dark gray. Current developments are shown in light gray.}
  \label{fig:layercake}
\end{figure}

Below, we first introduce the principles of the Semantic Web, from which we derive the architecture and its main components.
Subsequently, we describe the groups of the architecture.
The principles of the Semantic Web are:

\begin{enumerate}
\item \emph{Explicit and simple data representation:}
A general data representation abstracts from the underlying formats and captures only the essentials.

\item \emph{Distributed systems:}
A distributed system operates on a large set of data sources without centralized control that regulates which information belongs where and to whom.

\item \emph{Cross-references:}
The advantages of a network of data in answering queries are not based solely on the sheer quantities of data but on their interconnection, which allows reusing data and data definitions from other sources.

\item \emph{Loose coupling with common language constructs:}
The World Wide Web and likewise the Semantic Web are mega-systems, i.\,e., systems consisting of many subsystems, which are themselves large and complex.
In such a mega-system, individual components must be loosely coupled in order to achieve the greatest possible flexibility.
Communication between the components is based on standardized protocols and languages, whereby these can be individually adapted to specific systems. 

\item \emph{Easy publishing and easy consumption:}
Especially in a mega-system, participation, i.\,e., publishing and consumption of data, must be as simple as possible.
\end{enumerate}
These principles are achieved through a mix of protocols, language definitions, and software components.
Some of these components have already been standardized by the W3C, which has defined both syntax and formal semantics of languages and protocols. 
Other components are not yet standardized, but they are already provided for the so-called \emph{Semantic Web Layer Cake}\index{Semantic Web Layer Cake} by Tim Berners-Lee (cf.~{\tt http://www.w3.org/2007/03/layerCake.png}).
We present a variant of the Semantic Web architecture, distinguishing between standardized languages and current developments.
A graphical representation of the architecture can be found in Figure~\ref{fig:layercake}.

\paragraph*{Identifier for Resources: HTTP, URL, DID}
Entities (also called resources) are identified on the Internet by so-called Uniform Resource Identifiers (URIs)~\cite{rfc1630}.
When a URI holds a dereferenceable location of the resource, in other words, it can be employed to get access to the resource via HTTP, it is called a Uniform Resource Locator (URL)~\cite{rfc1738,rfc3986}.
Furthermore, Internationalized Resource Identifiers (IRIs)~\cite{rfc3987} supplement URIs with international character sets from Unicode/ISO10646.
URIs are globally and universally used but are usually not under our control.
A recent W3C recommendation, the Decentralized Identifiers (DIDs), introduces an alternative approach to the above identifiers~\cite{w3c-did}.
A DID is by default decentralized and allows for self-sovereign management of the identity, i.\,e., the control of a DID and its associated data is with the users.

In our example in Section~\ref{sec:musicscenario}, a URI \footnote{\url{http://www.bbc.co.uk/music/artists/2f031686-3f01-4f33-a4fc-fb3944532efa#artist}} describes the musician Benny Andersson of the Swedish pop group ABBA.
A user can dereference a URI that refers to ABBA, e.\,g., by performing a so-called look-up using HTTP to obtain a detailed description of the URI.
We refer to the referenced standards for details.

For a detailed discussion of the role of dereferenceable URIs on the Semantic Web, we refer to the Linked Data principles described in Section~\ref{sec:linked-graph-data}.
 
\paragraph*{Syntax for Data Exchange: XML, JSON-LD, RDFa} 
The Extensible Markup Language (XML)\myurl{https://www.w3.org/TR/xml/} is used to structure documents and enables the specification and serialization of structured data. In addition, other data formats were introduced to facilitate for serialization of RDF data, often replacing XML.
We can view those formats as forming two groups.
The first group consists of formats designed specifically for RDF data, such as Turtle\myurl{https://www.w3.org/TR/turtle/}, N-triple\footnote{\url{https://www.w3.org/TR/n-triples/}}, and TRIG\myurl{https://www.w3.org/TR/trig/}.
These are easier to view in a text editor, compared to XML, and thus easier to understand and modify.
While initially not included in standards, their popularity has led to them being official W3C recommendations since 2014.
The other group of formats is built by extending existing data formats.
As a result, those can be employed to add RDF to existing systems.
Examples of such formats are JSON-LD\myurl{https://www.w3.org/TR/json-ld/}, CSV on the Web (CSVW)\myurl{https://www.w3.org/TR/tabular-data-primer/}, and RDFa\myurl{https://www.w3.org/TR/rdfa-primer/} extending JSON, CSV, and (X)HTML, respectively.

\paragraph*{Graph Data Representation: RDF}

In addition to the referencing of resources and a uniform syntax for the exchange of data, a data model is required that allows resources to be described both individually and in their entirety and how they are linked~\cite{DBLP:books/crc/linked2014,DBLP:reference/bdt/HartigHS19}.
An integrated representation of data from multiple sources is provided by a data model based on directed graphs~\cite{key:oem}.
The corresponding W3C standard language is RDF (Resource Description Framework)\footnote{\url{https://www.w3.org/RDF/}}.

An RDF graph consists of a set of RDF triples, where a triple consists of a subject, predicate (property), and object.
An RDF graph can be given an identifier, such a graph is called a named graph~\cite{DBLP:conf/semweb/CarrollBHS04}.
RDF graphs can be serialized in several ways (see Syntax for Data Exchange above).
It is important to note that some formats (e.\,g., Turtle) do not support named graphs. 
Finally, RDF-star (aka RDF*)\footnote{\url{https://w3c.github.io/rdf-star/}}$^,$\footnote{\url{https://blog.liu.se/olafhartig/2019/01/10/position-statement-rdf-star-and-sparql-star/}} was introduced to allow nesting of triples and thus enables an efficient way to make statements about statements while avoiding reification and the increased number of triples and complexity that come along with it.

The representation of graph data is further discussed in Section~\ref{sec:representation}.

\paragraph*{Creation and Validation of Graph Data: RIF, SWRL, [R2]RML, and SHACL}

In the RDF context, a rule is a logical statement employed to infer new facts from existing graph data or to validate the data itself.
RIF (Rule Interchange Format)\myurl{http://www.w3.org/2005/rules/wiki/RIF_Working_Group} is a W3C recommendation format designed to facilitate the seamless interchange of rules between different rule engines.
This enables the extraction of rules from one engine, their translation into RIF, publication, and subsequent conversion into the native syntax of another rule engine for execution.
SWRL\myurl{https://www.w3.org/submissions/SWRL/} is a rule-based language designed for representing complex relationships 
and reasoning.

Rules can also be used to state the correspondence between data sources and RDF graphs. 
The RDB to RDF Mapping Language (R2RML)\myurl{https://www.w3.org/TR/r2rml/} and the RDF Mapping Language (RML)\myurl{https://rml.io/specs/rml/} correspond to rule-based mapping languages for the declarative definition of RDF graphs. R2RML is the W3C recommendation for representing mappings from relational databases to RDF datasets, while RML extends R2RML to express rules not only from relational databases but also from data in the format of CSV, JSON, or XML.

Validating constraints, representing syntactic and semantic restrictions in RDF graphs, is essential for ensuring data quality.
In addition to rule-based languages, shapes allow for the specification of conditions to meet data quality criteria and integrity constraints.
A shape encompasses a conjunction of constraints representing conditions that nodes in an RDF graph must satisfy~\cite{DBLP:series/synthesis/2021Hogan}. 
A shapes graph is a labeled directed graph where nodes correspond to shapes, and edges denote interrelated constraints. 
The Shapes Constraint Language (SHACL)~\cite{SHACL} and Shape Expressions (ShEx)~\cite{SHEX}) are two W3C-recommendations  to express shapes graphs over RDF~\cite{DBLP:conf/www/RabbaniLH22}.

The creation and validation of graph data are described in detail in Section~\ref{sec:creation}.
We introduce OWL and reasoning on graph data below.
 
\paragraph*{Reasoning and Linking of Graph Data: RDFS, OWL}

Data from different sources may be heterogeneous. 
In order to deal with this heterogeneity and to model the semantic relationships between resources, the RDF Schema (RDFS)\myurl{https://www.w3.org/TR/rdf11-schema/} vocabulary extends RDF by modeling types of resources (so-called RDF classes) and semantic relationships on types and properties in the form of generalizations and specializations.
Likewise, it can be used to model the domain and range of properties.

RDFS is not expressive enough to merge data from different sources and define consistency criteria about it, such as the disjointness of classes or the equivalence of resources.
The Web Ontology Language (OWL)\myurl{https://www.w3.org/OWL/}~\cite{key:owl2} is an ontology language with formally defined meaning based on description logic.
This allows for reasoning services to be provided by knowledge-based systems for OWL ontologies. 
OWL can be exchanged using RDF data formats.
Compared to RDFS, OWL provides more expressive language constructs.
For example, OWL allows the specification of equivalences between classes and cardinality constraints on properties~\cite{key:owl2}.

Reasoning over RDF graphs, incorporating RDFS and OWL  models enhances semantic expressiveness and inferential capabilities. 
This involves making implicit information explicit, inferring new triples, and validating the RDF graph's consistency against defined ontological constraints. RDFS provides basic entailment regimens, creating hierarchies and simple inferencing via sub-class and sub-property relationships. In contrast, OWL introduces advanced constructs such as property characteristics (e.\,g., functional, inverse, symmetric properties), cardinalities, and disjointness axioms, enabling more expressive and complex modeling. Integrating RDFS and OWL reasoning mechanisms empowers applications to derive insights, discover implicit knowledge, and ensure adherence to specified ontological constraints within RDF-based knowledge representations. 

Graph data aggregated from many data sources, such as in our example in Section~\ref{sec:musicscenario}, may contain many different identities.
But those identities may represent the same set of real-world objects.
Integration and linkage mechanisms allow references to be made between data from different sources.
A popular approach to state the identity of two resources $v$ and $w$ is the \texttt{owl:sameAs} feature of OWL.

We discuss the reasoning over and linking of graph data in Section~\ref{sec:reasoning}.

\paragraph*{Querying of Graph Data: SPARQL}
Since RDF makes it possible to integrate data from different sources, a query language is needed that allows formulating queries over individual RDF graphs as well as over the combination of multiple RDF graphs across multiple sources. 
SPARQL\myurl{http://www.w3.org/TR/rdf-sparql-query/} (a recursive acronym for SPARQL Protocol and RDF Query Language) is a declarative query language for RDF graphs that enables us to formulate such queries. 
SPARQL~1.1\myurl{http://www.w3.org/TR/sparql11-query/} is the current version of SPARQL, which includes the capability to formulate federated queries over distributed data sources. 

The basic building blocks of a SPARQL query are triple and graph patterns. 
A triple pattern corresponds to an RDF triple but where one, two, or all three of its components are replaced by variables (denoted with a leading ``?''). 
These triple patterns with variables are to be matched in the queried graph.
Multiple triple patterns can be combined into more complex graph patterns describing the connections between multiple nodes in the graph. 
The solution to such a SPARQL query then corresponds to all the subgraphs in an RDF graph matching this pattern.

Finally, there is RDF-star (aka RDF*)\footnote{\url{https://w3c.github.io/rdf-star/}}$^,$\footnote{\url{https://blog.liu.se/olafhartig/2019/01/10/position-statement-rdf-star-and-sparql-star/}} -- along with the corresponding SPARQL-star/ SPARQL* extension -- was proposed and since then was implemented by several triple stores~\cite{AbuAebAgl23} that often provide publicly accessible SPARQL endpoints.
The key idea with RDF/SPARQL-star is to allow the nesting of triples to enable an efficient way to allow statements about statements while avoiding reification and the increased number of triples and complexity that come along with it.

We describe SPARQL and federated querying in Section~\ref{sec:distributeddataquerying}.

\paragraph*{Crypto, Provenance, and Trustworthiness of Graph Data}

Other aspects of the Semantic Web are encryption and authentication to ensure that data transmissions cannot be intercepted, read, or modified.
Crypto modules, such as SSL (Secure Socket Layer), verify digital certificates and enable data protection and authentication.
In addition, there are digital signatures for graphs that integrate seamlessly into the architecture of the Semantic Web and are themselves modeled as graphs again~\cite{DBLP:conf/esws/KastenSS14}.
This allows graph signatures to be applied iteratively and enables to building trust networks.
The Verifiable Credentials Data Model, a recent W3C recommendation, introduces a standard to model trustworthy credentials for graphs on the web\myurl{https://www.w3.org/TR/vc-data-model/}.
Data on the Semantic Web can be augmented with additional information about its trustworthiness and provenance.

Aspects of trustworthiness and provenance of graph data as well as crypto are discussed in Section~\ref{sec:trustprovenance}.

\paragraph*{User Interfaces and Applications}
A user interface enables users to interact with data on the Semantic Web.
From a functional perspective, some user interfaces are generic and operate on the graph structure of the data, whereas others are tailored to specific tasks, applications, or ontologies. 
New paradigms are exploring the spectrum of possible user interfaces between generality and specific end-user requirements.

Semantic Web applications are discussed in Section~\ref{sec:userinterfaces}.
The impact on practitioners is described in Section~\ref{sec:impact}.

\section{Representation of Graph Data}
\label{sec:representation} 

The Linked Open Data principles are notably the most successful and widely adopted choice for representing RDF graph data on the web.
Thus, we first introduce the reader to how to represent graph data as Linked Data.
Subsequently, we introduce the notion of ontologies. 
This is followed by a more detailed analysis of the different types of ontologies.
We give examples of ontologies throughout the sections.
With this background in mind, we reconsider our running example from Section~\ref{sec:musicscenario} and analyze the given distributed network of ontologies. 
In this context, we also introduce and discuss the notion of ontology design patterns.

\subsection{Linked Graph Data on the Web}
\label{sec:linked-graph-data}
The \emph{Linked Data} principles\myurl{http://www.w3.org/DesignIssues/LinkedData.html}\index{Linked Data principles} define the methods for representing, publishing, and using data on the Semantic Web.
They can be summarized as follows:
\begin{enumerate}
\item URIs are used as names for entities.
\item The HTTP protocol's GET method is used to retrieve descriptions for a URI.
\item Data providers shall return relevant information in response to HTTP GET requests on URIs using standards, e.\,g., in RDF.
\item Links to other URIs shall be used to facilitate knowledge discovery and use of additional information.
\end{enumerate}
Publishing data using Linked Data principles allows easy access to data via HTTP.
This allows exploration of resources and navigation across resources on the Semantic Web.
URIs (see 1.) are dereferenced using HTTP requests~(2.) to obtain additional information about a given resource.
In particular, using standardized syntax~(3.), this information may also contain links to other resources~(4.).

\begin{figure}[ht]
  \centering
  \includegraphics[width=\textwidth]{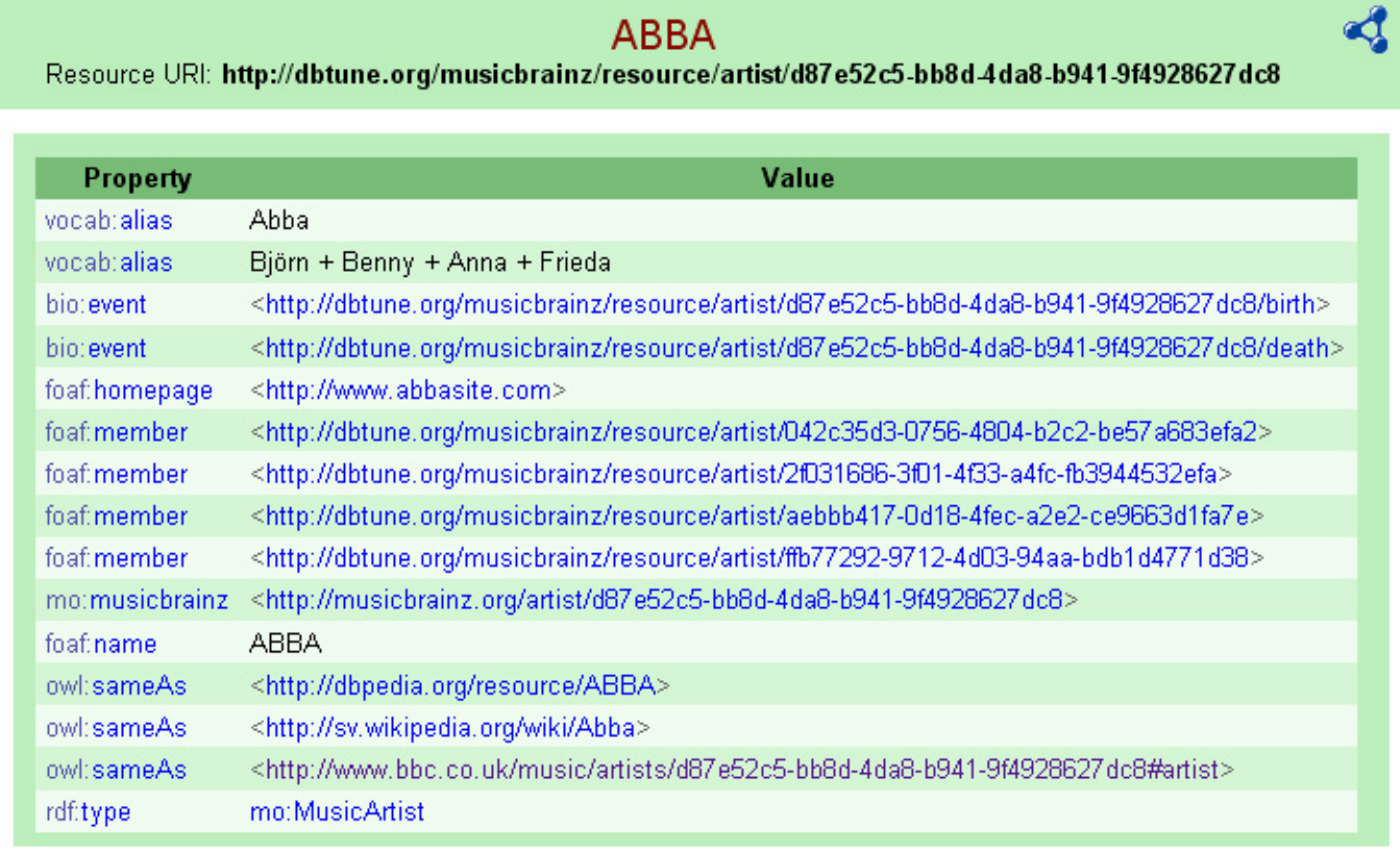}
\caption{Linked Data example for ABBA.}\label{fig:abba_linked_data}
\end{figure}

Figure~\ref{fig:abba_linked_data} represents an example of Linked Data about the pop group ABBA.
The example describes several relationships linking entities to ABBA's URI, such as {\tt foaf:member} and {\tt rdf:type}.
In the figure ``ABBA'', or more precisely the URI of ABBA, is the subject, ``Property'' refers to relationships, and ``Value'' represents objects of the RDF triples.
The relation {\tt owl:sameAs} will be explained in more in Section \ref{sec:reasoning}. 
The prefixes \texttt{foaf}, \texttt{rdf}, and \texttt{owl} refer to vocabularies of the FOAF ontology\myurl{http://xmlns.com/foaf/spec/}, and the W3C language specifications of RDF and OWL, respectively.

\subsection{Ontologies}
\label{sec:ontologies:definition}

An ontology is commonly defined as a formal, machine-readable representation of key concepts and relationships within a specific domain~\cite{GuarinoOberleStaabWhatIsAnOntology2009,OberleMiddleware2006}.
In essence, ontologies capture a shared perspective~\cite{GuarinoOberleStaabWhatIsAnOntology2009} that is, the formal conceptualization of ontologies expresses a consensus view among different stakeholders. 
Visualizing ontologies is akin to viewing a spectrum, with a specificity of concepts, their relationships, and the granularity of meaning representation varying along this continuum
~\cite{DBLP:conf/dagstuhl/McGuinness03,DBLP:journals/sigmod/UscholdG04,DBLP:journals/ao/Uschold15}.
A controlled vocabulary corresponds to the less expressive form of ontology, comprising a restrictive list of words or terms used for labeling, indexing, or categorization. The Clinical Data Interchange Standards Consortium (CDISC) Terminology is an exemplary vocabulary that harmonizes definitions in clinical research.\myurl{https://datascience.cancer.gov/resources/cancer-vocabulary/cdisc-terminology} 
A thesaurus is located next in the spectrum; they enhance controlled vocabularies with information about terms and their synonyms and broader/narrower relationships. The Unified Medical Language System (UMLS) integrates medical terms and their synonyms.\myurl{https://www.nlm.nih.gov/research/umls/index.html}
Next, taxonomies are built over controlled vocabularies to provide a hierarchical structure, e.\,g., parent/child relationship. 
SNOMED-CT\myurl{https://www.snomed.org/value-of-snomedct} (Systematized Nomenclature of Medicine Clinical Terms) provides a terminology and coding system used in healthcare and medical fields; medical concepts organized in a hierarchical structure enabling a granular representation of clinical information. 
The Simple Knowledge Organization System (SKOS)\myurl{https://www.w3.org/TR/skos-reference/} is a W3C standard to describe knowledge about organizational systems. 
Lastly, ontologies are at the highest extreme of the spectrum, integrating sets of concepts with attributes and relationships to define a domain of knowledge.

Note, SKOS is a popular standard for modeling domain-specific taxonomies in the different scientific communities such as economics, social sciences, etc. to represent concepts and their relationships, most importantly narrower, broader, and related. However, it does not have the expressiveness of OWL with its complex expressions on classes and relations. For a detailed discussion, we refer to the literature such as~\cite{DBLP:conf/owled/JuppBS08} and the W3C on using OWL and SKOS\myurl{https://www.w3.org/2006/07/SWD/SKOS/skos-and-owl/master.html}.

\subsection{Types and Examples of Ontologies}
\label{sec:ontologies}

A network of ontologies, such as the example shown in Figure~\ref{fig:bbc-ontology-example}, may consist of a variety of ontologies created by different actors and communities. 
Ontologies may be the result of a transformation or reengineering activity of a legacy system, such as a relational database or existing taxonomy such as the Dewey Decimal Classification\myurl{https://www.oclc.org/en/dewey.html} or Dublin Core.
Other ontologies are created from scratch.
This involves applying existing methods and tools for ontology engineering and choosing an appropriate representation language for the ontology (see Section~\ref{sec:reasoning}). 

Ontology engineering deals with the methods for creating ontologies~\cite{GomezPerezFernandezLopezOntologicalEngineering} and has its origins in software engineering in the creation of domain models and in database design in the creation of conceptual models.
A good overview of ontology engineering can be found in several reference books~\cite{GomezPerezFernandezLopezOntologicalEngineering}.
Ontologies vary greatly in their structure, size, development methods applied, and application domains considered. 
Complex ontologies are also distinguished in terms of their purpose and granularity.

\textbf{Domain Ontologies} represent knowledge specific to a particular domain~\cite{Euzenat2007a,OberleMiddleware2006}. 
Domain ontologies are used as external sources of background knowledge~\cite{Euzenat2007a}.
They can be built on foundational ontologies~\cite{Oberle:2007:DES:1290204.1290385} or core ontologies~\cite{ScherpEtAlDesigningCoreOntologiesIOS2011}, which provide precise structuring to the domain ontology and thus improve interoperability between different domain ontologies.
Domain ontologies can be simple such as the FOAF ontology or the event ontology mentioned above, or very complex and extensive, having been developed by domain experts, such as the SNOMED medical ontology.

\textbf{Core Ontologies} represent a precise definition of structured knowledge in a particular domain spanning multiple application domains~\cite{ScherpEtAlDesigningCoreOntologiesIOS2011,OberleMiddleware2006}. 
Examples of core ontologies include core ontologies for software components and web services~\cite{OberleMiddleware2006}, for events and event relationships~\cite{emf-mtap}, or for multimedia metadata~\cite{www-m3o}.
Core ontologies should thereby build on foundational ontologies to benefit from their formalization and strong axiomatization~\cite{ScherpEtAlDesigningCoreOntologiesIOS2011}.
For this purpose, new concepts and relations are added to core ontologies for the application domain under consideration and are specialized by foundational ontologies.
 
\textbf{Foundational Ontologies} have a very wide scope and can be reused in a wide variety of modeling scenarios~\cite{BorgoMasoloDOLCE2009}. 
They are therefore used for reference purposes~\cite{OberleMiddleware2006} and aim to model the most general and generic concepts and relations that can be used to describe almost any aspect of our world~\cite{BorgoMasoloDOLCE2009,OberleMiddleware2006}, such as objects and events.
An example is the Descriptive Ontology for Linguistic and Cognitive Engineering (DOLCE)~\cite{BorgoMasoloDOLCE2009}.
Such basic ontologies have a rich axiomatization that is important at the developmental stage of ontologies.
They help ontology engineers to have a formal and internally consistent conceptualization of the world, which can be modeled and checked for consistency.
For the use of foundational ontologies in a concrete application, i.\,e., during the runtime of an application, the rich axiomatization can often be removed and replaced by a more lightweight version of the foundational ontology.

In contrast, domain ontologies are built specifically to allow automatic reasoning at runtime.
Therefore, when designing and developing ontologies, completeness and complexity on the one hand must always be balanced with the efficiency of reasoning mechanisms on the other.
In order to represent structured knowledge, such as the scenario depicted in Figure~\ref{fig:bbc-ontology-example}, interconnected ontologies are needed, which are spanned in a network over the Internet. 
For this purpose, the ontologies used must match and be aligned with each other.

\subsection{Distributed Network of Ontologies and Ontology Patterns}

A network of ontologies must be flexible with respect to the functional requirements imposed on it. 
This is because systems are modified, extended, combined, or integrated over time.
In addition, the networked ontologies must lead to a common understanding of the modeled domain.
This common understanding can be achieved through a sufficient level of formalization and axiomatization, and through the use of ontology patterns.
An ontology pattern, similar to a design pattern in software engineering, represents a generic solution to a recurring modeling problem~\cite{ScherpEtAlDesigningCoreOntologiesIOS2011}. 
Ontology patterns allow to select parts from the original ontology.
Either all or only certain patterns of an ontology can be reused in the network.
Thus, to create a network of ontologies, e.\,g., existing ontologies and ontology patterns can be merged on the Web.
The ontology engineer can drive or explicitly provide for the modularization of ontologies using ontology patterns.
Core ontologies represent one approach to designing a network of ontologies (see in detail~\cite{ScherpEtAlDesigningCoreOntologiesIOS2011}).
They allow to capture and exchange structured knowledge in complex domains.
Well-defined core ontologies fulfill the properties mentioned in the previous section and allow easy integration and smooth interaction of ontologies (see also~\cite{ScherpEtAlDesigningCoreOntologiesIOS2011}).
The networked ontologies approach leads to a flat structure, as shown in Figure~\ref{fig:bbc-ontology-example}, where all ontologies are used on the same level.
Such structures can be managed up to a certain level of complexity.

\begin{figure}[ht]
  \centering
   \includegraphics[scale=0.8]{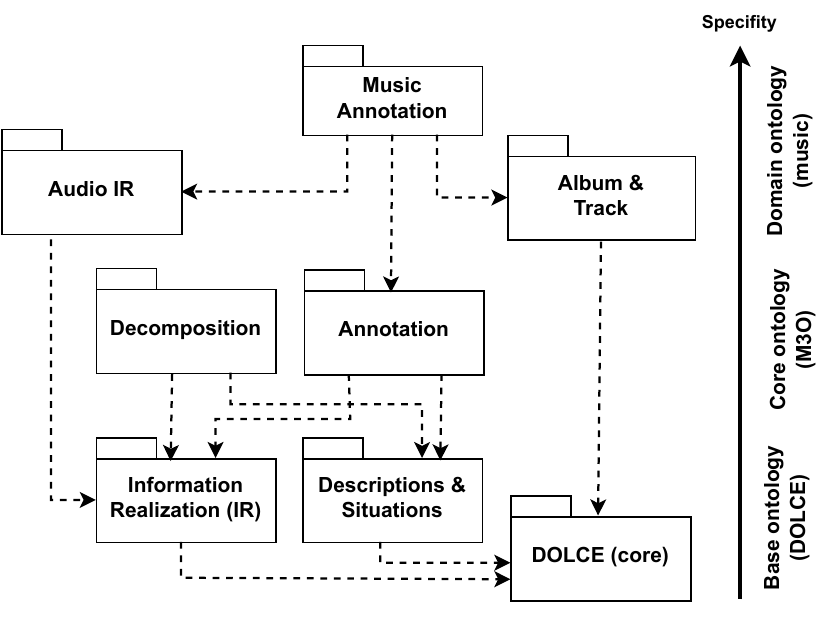}
   \caption{Ontology layers combining the foundational ontology DOLCE, the multimedia metadata ontology M3O, domain-specific extensions to M3O for annotating audio data and music, and a domain ontology for albums and tracks.}
   \label{fig:ontology-stack}
\end{figure} 

The approach of networked core ontologies is illustrated by the example of ontology layers starting from foundational to core to domain ontologies.
As shown in Figure~\ref{fig:ontology-stack}, DOLCE is the foundational ontology at the bottom layer, the Multimedia Metadata Ontology (M3O)~\cite{www-m3o} as the core ontology for multimedia metadata, and an extension of M3O for the music domain.
Core ontologies are typically defined in description logic and cover a field larger than the specific application domain requires~\cite{GangemiPresuttiODP2009}. 
Concrete information systems will typically use only a subset of core ontologies.
To achieve modularization of core ontologies, they should be designed using ontology patterns.
By precisely matching the concepts in the core ontology with the concepts provided in the foundational ontology, they provide a solid foundation for future extensions.
New patterns can be added and existing patterns can be extended by specializing the concepts and roles.
Figure~\ref{fig:ontology-stack} shows different patterns of the M3O and DOLCE ontologies.

Ideally, the ontology patterns of the core ontologies are reused in the domain ontologies~\cite{GangemiPresuttiODP2009}, as shown in Figure~\ref{fig:ontology-stack}.
However, since it cannot be assumed that all domain ontologies are aligned with a foundational or core ontology, the option that domain ontologies are developed and maintained independently must also be considered.
In this case, domain knowledge can be reused in core ontologies by applying the Descriptions and Situations (DnS) ontology pattern of the foundational ontology DOLCE.
The DnS ontology pattern is an ontological formalization of context~\cite{OberleMiddleware2006} by defining different views using roles.
These roles can refer to domain ontologies and allow a clear separation of the structured knowledge of the core ontology and domain-specific knowledge.
To model a network of ontologies, such as the example described above, the Web Ontology Language (OWL) and its ability to axiomatize using description logic~\cite{DBLP:conf/dlog/2003handbook} is used.
In addition to being used to model a distributed knowledge representation and integration, OWL, is also used in particular to derive inferences from this knowledge, which is described in Section~\ref{sec:reasoning}.

\section{Creation and Validation of Graph Data}
\label{sec:creation}
In this section, we describe the creation of graph data from legacy data. 
Many tools are available for this task, which support various mappings and transformations.
Subsequently, we discuss data quality and the validation of knowledge graphs, including the recent approaches on shapes.
We also reflect on the role of the open-world versus closed-world assumption with respect to validating data.

\subsection{Graph Data Creation}
Graph data can be created by transforming legacy data via a data integration system~\cite{lenzerini2002data}, which consists of a unified schema, data sources, and mapping rules. 
These mapping rules define the concepts within the schema and establish links to the data sources.
By employing declarative definitions, knowledge graph creation promotes modularity and reusability.
This approach allows users to trace the entire graph creation process, leading to improved transparency and ease of maintenance.

To enable comprehensive and extensive graph specification, mappings and transformations have been developed to convert data from various storage models into Semantic Web data models like RDF. 
These mappings and transformations facilitate the mapping of data into RDF, thereby supporting the integration of diverse data sources into the Semantic Web.

\begin{figure}[ht]
  \centering
  \includegraphics[width=1\textwidth]{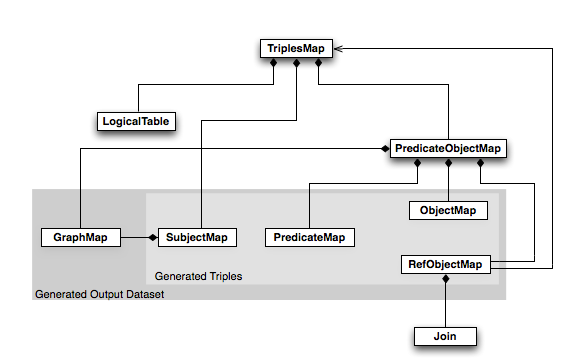}
\caption{Structure of a relational data mapping (source: \cite{W3CR2RML}).}\label{fig:r2rml}
\end{figure}

The mapping language R2RML~\cite{W3CR2RML} defines mapping rules from relational databases (relational data models) to RDF graphs.
These mappings themselves are also RDF triples~\cite{W3TURTLE}. 
Because of its compact representation, Turtle is considered a user-friendly notation of RDF graphs.
The structure of R2RML is illustrated in Figure \ref{fig:r2rml}; essentially, table contents are mapped to triples by the classes \texttt{SubjectMap}, \texttt{PredicateMap}, and \texttt{ObjectMap}.
If the object is a reference to another table, this reference is called \texttt{RefObjectMap}. 
Here, \texttt{SubjectMap} contains primary key attributes of the corresponding table. 
Thus, there exists a mapping rule representable in RDF graphs by means of which tables of relational databases can be represented as RDF graphs.

The RDF Mapping Language (RML)\cite{DimouSCVMW14} extends R2RML to encompass the definition of logical sources in various formats, including CSV, JSON, XML, and HTML.
This enhancement enables RML to introduce new operators that facilitate the integration of data from diverse sources into the Semantic Web.
Thus, instead of \texttt{LogicalTable}, RML includes the tag \texttt{LogicalSource}, to allow for the retrieval of data in several formats.
Additionally, RML resorts to W3C-standardized vocabularies and enables the definition of retrieval procedures to collect data from Web APIs or databases. R2RML and RDF mapping rules are expressed in RDF, and their graphs document how classes and properties in one or various ontologies that are part of an RDF graph are populated from data collected from potentially heterogeneous data sources. 
 
Over time, the Semantic Web community has actively contributed to addressing the challenge of integrating heterogeneous datasets, resulting in the development of several frameworks for executing declarative mapping rules~\cite{chebotko2009semantics,calvanese2017ontop,DBLP:conf/www/PriyatnaCS14}.
A rich spectrum of tools (e.\,g., RMLMapper~\cite{DimouSCVMW14}, RocketRML~\cite{csimcsek2019rocketrml}, CARML\footnote{\url{https://github.com/carml/carml}}, SDM-RDFizer~\cite{iglesias2020sdm}, Morph-KGC~\cite{arenas2022morph}, and RMLStreamer~\cite{oo2022rmlstreamer}) offers the possibility of executing R2RML and RML rules and efficiently materializing the transformed data into RDF graphs.
Van Assche et\,al.~\cite{DBLP:journals/ws/AsscheDHHMD23} provided an extensive survey detailing the main characteristics of these engines.
Despite significant efforts in developing these solutions, certain parameters can impact the performance of the graph creation process~\cite{Chaves-FragaEIC19}.
Existing engines may face challenges when handling complex mapping rules or large data sources.
Nonetheless, the community continues to collaborate and address these issues.
An example of such collaboration is the Knowledge Graph Construction Workshop 2023 Challenge\footnote{\url{https://zenodo.org/record/7689310}} that took place at ESWC 2023.
This community event aims to understand the strengths and weaknesses of existing approaches and devise effective methods to overcome existing limitations.
 
RDF graphs can also be dynamically created through the execution of queries over data sources. These queries involve the rewriting of queries expressed in terms of an ontology, based on mapping rules that establish correspondences between data sources and the ontology. Tools such as Ontop~\cite{calvanese2017ontop}, Ultrawrap~\cite{DBLP:journals/ws/SequedaM13}, Morph~\cite{DBLP:conf/www/PriyatnaCS14}, Squerall~\cite{mami2019squerall}, and Morph-CSV~\cite{chaves2020enhancing} exemplify systems that facilitate the virtual creation of RDF graphs.

\subsection{Quality and Validation of Graph Data}
\label{sec:quality-and-validation}

Quality and validation of the graph data are crucial to maintaining the integrity of the Semantic Web~\cite{DBLP:journals/corr/abs-2308-14217,DBLP:journals/semweb/Debattista0AC18,Zaveri2015surveyQuality}. The evaluation of integrity constraints allows for the identification of inconsistencies, inaccuracies, or contradictions within the data. They also help maintain consistency by ensuring related data elements remain coherent. Constraints are logical statements -- expressed in a particular language -- that impose restrictions on the values taken for target nodes in a given property.

Constraints can be expressed using OWL~\cite{DBLP:conf/aaai/TaoSBM10}, SPARQL queries~\cite{DBLP:conf/edbt/LausenMS08}, or using shapes. However, the interpretation of the results depends on the semantics followed to interpret the failure of an integrity constraint. For example, constraints expressed in OWL are validated using an Open-World Assumption (OWA) (i.\,e., a statement cannot be inferred to be false based on failures to prove it) and under the absence of the Unique Name Assumption (UNA) (i.\,e., two different names may refer to the same object). These two features make it difficult to validate data in applications where data is supposed to be complete. Definitions of integrity constraint semantics in OWL using the Closed-World Assumption~\cite{Motik2007,Motik2009,DBLP:conf/aaai/TaoSBM10} overcome these issues.

Contrarily, constraints expressed using SPARQL queries or shapes will be evaluated under the Closed-World Assumption (CWA) and following the Unique Name Assumption (UNA). Nevertheless, some constraints may be difficult to express in SPARQL, and the specification process is prone to errors and difficult to maintain.
 
\begin{figure}[!ht]
\centering
\includegraphics[width=1\textwidth]{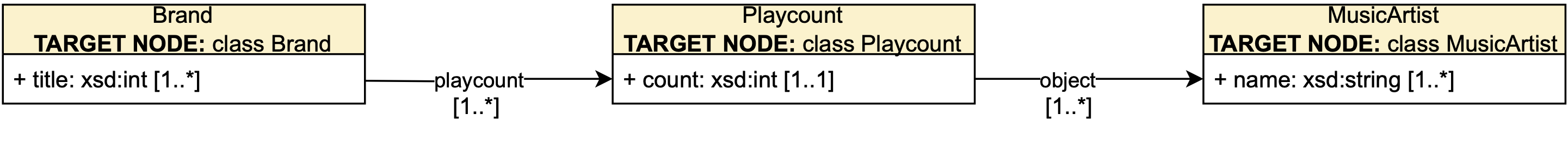}
\caption{\textbf{Shapes for Graph Data.} A shapes graph comprises three shapes interlinked by the properties \texttt{playcount} and \texttt{objects} between the target classes \texttt{Brand}, \texttt{Playcount}, and \texttt{MusicArtist}.}
\label{fig:shape}
\end{figure}

Data quality conditions and integrity constraints can also be expressed as graphs of shapes or the so-called \emph{shapes schema}. 
A shape corresponds to a conjunction of constraints that a set of nodes in an RDF graph must satisfy~\cite{DBLP:series/synthesis/2021Hogan}.
These constraints can restrict the types of nodes, the cardinality of certain properties, and the expected data types or values for specific properties. A shape can target the instances of a class, the domain or range of a property, or a specific node in the RDF graph. A shape or node in a shapes graph is validated in an RDF graph, if and only if, all the target nodes in the RDF graph satisfy all the constraints in the shape.
Figure~\ref{fig:shape} presents a shapes graph of three shapes targeting the classes  
\texttt{Brand}, \texttt{Playcount}, and \texttt{MusicArtist}. Each of the shapes comprises one constraint. In the shapes \texttt{Brand} and \texttt{MusicArtist} the properties title and name can take more than one value, while the shape \texttt{Playcount} states that each instance of the class \texttt{Playcount} must have exactly one value of the property \texttt{count}. Additionally, the instances of the class \texttt{Brand} must be related to valid instances of the class \texttt{Playcount} which should also be related to valid instances of the class \texttt{MusicArtist}.

There are two standards for defining shapes, ShEx (Shape Expressions)~\cite{SHEX}) and SHACL (Shapes Constraint Language)~\cite{SHACL}).
Both define shapes over the attributes (i.\,e., \texttt{owl:Datatype\-Properties}), and constraints on incoming/outgoing arcs, cardinalities, RDF syntax, and extension mechanism. These inter-class constraints induce a shape network used to validate the integrity and data quality properties of an RDF graph. 
 
SHACL and ShEx, although sharing a common goal, adopt distinct approaches. ShEx seeks to offer a language serving as a grammar or schema for RDF graphs, delineating RDF graph structures for validation. On the other hand, SHACL is positioned as the W3C recommendation for validating RDF graphs against a conjunction of constraints, emphasizing a constraint language for RDF. Despite their analogous roles in specifying shapes and constraints for RDF data, ShEx and SHACL differ in syntax, expressiveness, and community adoption \cite{DBLP:series/synthesis/2017Gayo}.

The evaluation results of a SHACL shape network over an RDF graph are presented in validation reports using a controlled vocabulary. A validation report includes explanations about the violations, the severity of the violation, and a message describing the violation. SHACL is the language selected by the International Data Space (IDS) to express the restrictions that state the integrity over RDF graphs \cite{DBLP:books/sp/22/0002LB22}. Besides the integrity validation of an RDF graph, SHACL can be utilized 
to describe data sources and the certification of a query answer \cite{DBLP:conf/vldb/Rohde21},
as metadata to enhance the performance of a SPARQL query engine \cite{DBLP:conf/edbt/RabbaniLH21},
to certify access policies \cite{DBLP:conf/esws/RohdeV23}, and 
to provide provenance as a result of the validation of integrity constraints~\cite{DBLP:conf/edbt/DelvaDJB23}. 

In the context of a quality assessment pipeline, one crucial step involves validating the shape schema against a graph. It is important to mention that the validation of recursive shape schemas is not explicitly addressed in the SHACL specification~\cite{SHACL}. 
To address this gap, Corman et al.~\cite{Corman2018} introduce a semantic framework for validating recursive SHACL. They also demonstrated that validating full SHACL features is an NP-hard problem. Building on these insights, they proposed specific fragments of SHACL that are computationally tractable, along with a fundamental algorithm for validating shape schemas using SPARQL~\cite{Corman2019}.
In a related vein, Andresel et al.~\cite{Andresel2020} propose a stricter semantics for recursive SHACL, drawing inspiration from stable models employed in Answer Set Programming (ASP). This innovative approach enables the representation of SHACL constraints as logic programs and leverages existing ASP solvers for shape schema validation. Importantly, this approach allows for the inclusion of negations in recursive validations. Further, Figuera et al.~\cite{DBLP:conf/www/FigueraRV21} present Trav-SHACL, an approach that focuses on query optimization techniques aimed at enhancing the incremental behavior and scalability of shape schema validation.

While SHACL has been adopted in a broad range of use cases, given a large graph it remains a challenge how to define shapes efficiently~\cite{DBLP:conf/www/RabbaniLH22}.
In many industrial settings with billions of entities and facts~\cite{Noy19} creating shapes manually simply is not an option. 
The current state of the art can automatically extract shapes on WikiData (ca. 2 Billion facts) in less than 1.5 hours while filtering shapes based on the well-established notions of support and confidence to avoid reporting thousands of shapes that are so rare or apply to such a small subset of the data that they become meaningless~\cite{DBLP:journals/pvldb/RabbaniLH23}.
Still, more work is needed to increase scalability further and also to help users make good use of the mined shapes~\cite{DBLP:conf/sigmod/RabbaniLH23} and, e.\,g., interactively use them to correct and improve the quality of their graphs.

\section{Reasoning over and Linking of Graph Data}
\label{sec:reasoning}

Section~\ref{sec:swarchitecture} introduced several formal languages for knowledge representation on the Semantic Web.
RDF allows the description of simple facts (statements with subject, predicate, and object, so-called RDF triples), e.\,g., ``Anni-Frid Lyngstad'' ``is a member of'' ``ABBA''.
RDFS allows the definition of types of entities (classes), relationships between classes, and a subclass and superclass hierarchy between types (analogously for relations).
OWL is even more expressive than RDF and RDFS.
For example, OWL allows the definition of disjoint classes or the description of classes in terms of intersection, union, and complement of other classes.

Below, we first introduce the reasoning over RDFS and OWL at the example of our BBC scenario from Section~\ref{sec:musicscenario}.
Subsequently, we discuss works on linking data objects and concepts.

\subsection{Reasoning over Graph Data}

Based on formal languages representing graph data and their semantics, further (implicit) facts can be derived from the knowledge base by deductive inference.
In the following, we exemplify the derivation of implicit facts from a set of explicitly given facts using the RDFS construct \texttt{rdfs:subClassOf} and the OWL construct \texttt{owl:sameAs}.
The property \texttt{rdfs:subClassOf} describes hierarchical relationships between classes and with \texttt{owl:sameAs} two resources can be defined as identical.

\begin{figure}[ht]
  \centering
  \includegraphics[width=0.9\textwidth]{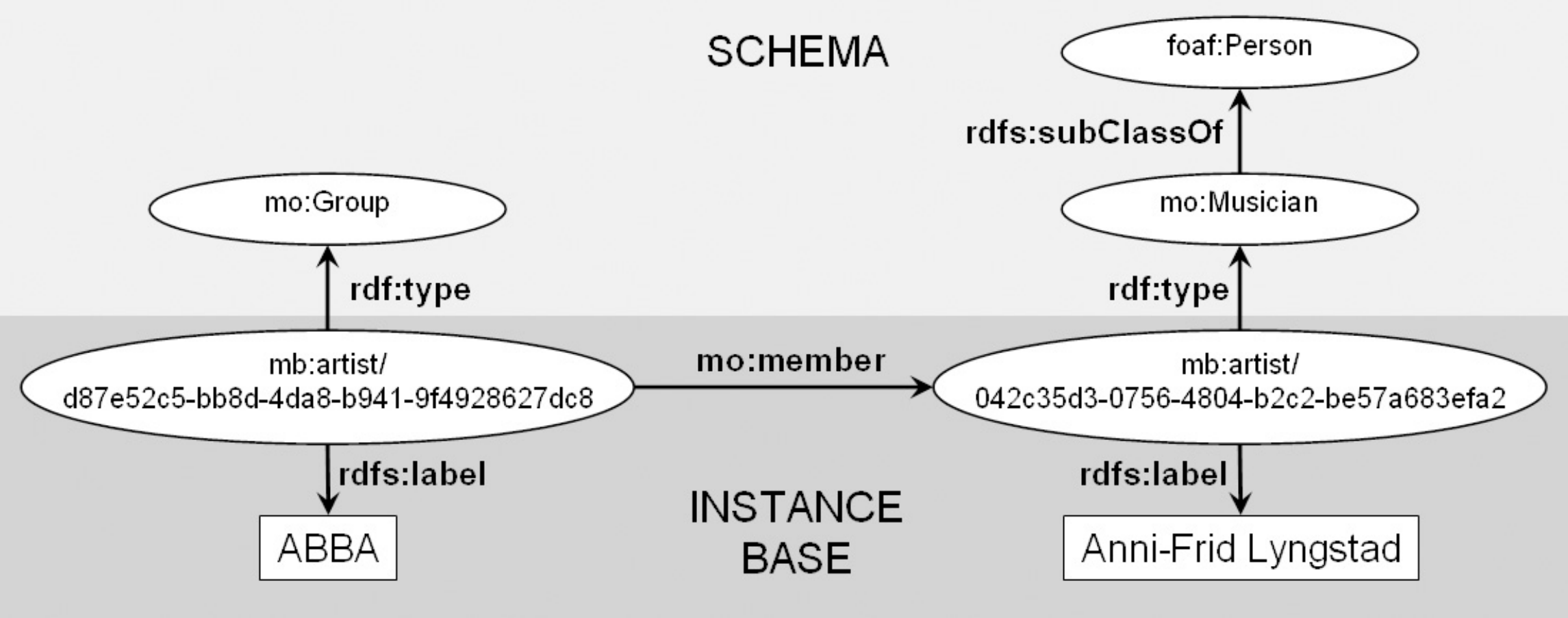}
\caption{Visualization of RDF sample data about ABBA and Anni-Frid Lyngstad to illustrate inference in RDFS.}\label{fig:rdf-example-graph}
\end{figure}

As a first example, we consider the class \texttt{foaf:Person}, which is defined in the FOAF ontology, and the classes \texttt{mo:Musician} and \texttt{mo:Group}, which are defined in the music ontology.
In the music ontology, there is an additional axiom that defines \texttt{mo:Musician} as a subclass of \texttt{foaf:Person} using \texttt{rdfs:subClassOf}.
Given this axiom, it can be deduced by deductive inference that instances of \texttt{mo:Musician} are also instances of \texttt{foaf:Person}.
Now if there is such a hierarchy of classes and in addition a statement that Anni-Frid Lyngstad is of type \texttt{mo:Musician}, then it can be inferred by inference that Anni-Frid Lyngstad is also of type \texttt{foaf:Person}.
This means that all queries asking for entities of type \texttt{foaf:Person} will also include Anni-Frid Lyngstad in the query result, even if that entity is not explicitly defined as an instance of \texttt{foaf:Person}.
Figure~\ref{fig:rdf-example-graph} represents these facts and the corresponding class hierarchy in RDFS as a directed graph.

In the second example, the OWL construct \texttt{owl:sameAs} is used to define two resources as identical, for example
\url{http://www.bbc.co.uk/music/artists/d87e52c5-bb8d-4da8-b941-9f4928627dc8#artist} and \url{http://dbpedia.org/resource/ABBA}.
Identical here means that these two URIs represent the same real-world object.
By inference, information about ABBA from different sources can now be linked.
Since ontologies are created independently on the web, and URIs are subject to local naming conventions, a real-world object may be represented by multiple URIs (in different ontologies).  

OWL offers a variety of other constructs for the description of classes, relationships, and concrete facts. 
For example, OWL allows the declaration of transitive relations and inverse relations. 
For example, the relation ``is-member'' is inverse to ``has-member''.
OWL reasoning allows, among other things, consistency checking of an ontology or checking the satisfiability of classes~\cite{DBLP:reference/sp/HorrocksP11}.
A class is satisfiable if there can be instances of that class.

For a detailed discussion about OWL reasoning, we refer to the literature such as~\cite{DBLP:reference/sp/HorrocksP11,DBLP:journals/sLogica/BaaderS01}.
Different reasoners for OWL have seen widespread adoption in the community such as the well-known Pellet\footnote{\url{https://github.com/stardog-union/pellet}} and Hermit~\cite{DBLP:journals/jar/GlimmHMSW14}.
Finally, a combination of description logic and rules is also possible.
For example, Motik et al.~\cite{DBLP:conf/semweb/MotikSS04} presented a combination of description logic and rules that allows tractable inference on OWL ontologies. 

\subsection{Linking of Objects and Concepts}
\label{sec:identity}

In the Semantic Web, it cannot be assumed that two URIs refer to two different real-world objects (cf. unique name assumption in Section~\ref{sec:quality-and-validation}).
A URI by itself, or in itself, has no identity~\cite{HalpinP09}.
Rather, the identity or interpretation of a URI is revealed by the context in which it is used on the Semantic Web.
Determining whether or not two URIs refer to the same entity is not a simple task and has been studied extensively in data mining and language understanding in the past.
For example, to identify whether or not the author names of research papers refer to the same person, it is often not sufficient to resolve the name, venue, title, and co-authors~\cite{PallikaEtAlCoreference2007}.
The process of determining the identity of a resource is often referred to as entity resolution~\cite{PallikaEtAlCoreference2007}, coreference resolution~\cite{WickEtAlCoreference}, object identification~\cite{RendleS06}, and normalization~\cite{WickEtAlCoreference,WickKDD2008}.
Correctly determining the identity of entities on the Web is important as more and more records appear on the Web and this presents a significant hurdle for very large Semantic Web applications~\cite{GlaserLOD2009}.

To address this, a number of services exist that can recognize entities and determine their identity:
Thomson Reuters offers OpenCalais\myurl{https://www.refinitiv.com/en/products/intelligent-tagging-text-analytics}, a service that can link natural language text to other resources using entity recognition. 
Another commercial tool that allows for extracting knowledge graphs from text is provided by DiffBot.\myurl{https://www.diffbot.com/}
Recently, the language model ChatGPT has been compared to the specialized entity and relation extraction tool REBEL~\cite{DBLP:conf/emnlp/CabotN21} for the task of creating knowledge graphs from sustainability-related text~\cite{DBLP:journals/corr/abs-2305-04676}.
The experiments suggest that large language models improve the accuracy of creating knowledge graphs~\cite{DBLP:journals/corr/abs-2305-04676}.
The sameAs\myurl{https://www.sameas.cc/} service aims to detect duplicate resources on the Semantic Web using the OWL relationship \concept{owl:sameAs}.
This can be used to resolve coreferences between different datasets.
For example, for the query with the URI \url{http://dbpedia.org/resource/ABBA}, a list of over 100 URIs is returned that also references the music group ABBA.
One of them is BBC with the resource \url{http://www.bbc.co.uk/music/artists/d87e52c5-bb8d-4da8-b941-9f4928627dc8#artist}.

Furthermore, the problem of schema matching~\cite{WickKDD2008} is very related to the problem of entity resolution, co-reference resolution, and normalization.
The goal of schema matching is to address the question of how to integrate data~\cite{WickKDD2008}, which is non-trivial even for small schemas. 
In the Semantic Web, schema matching means the matching of different ontologies, respectively the concepts defined in these ontologies.
Various (semi-)automatic or machine learning techniques for matching ontologies have been developed in the past~\cite{euzenat2007b,Ehrig07,Blomqvist09}.
Core ontologies as illustrated in Figure~\ref{sec:ontologies} represent generic modeling frameworks for integration and alignment with other ontologies. 
In addition, core ontologies can also integrate Linked Open Data, which typically contains no or very little schema information.
The YAGO ontology~\cite{YagoWWW07} was generated from the fusion of Wikipedia and Wordnet using rule-based and heuristic methods.
A manual assessment showed an accuracy of 95\%.

Manual matching of different data sources is also pursued in the Linked Open Data project of the German National Library\myurl{http://www.d-nb.de/}.
For example, the database containing the authors of all documents published in Germany was manually linked with DBpedia and other data sources.
A particular challenge was to identify the authors, as described above.
For example, former German Chancellor Helmut Kohl has a namesake whose work should not be linked to the chancellor's DBpedia entry.
Relationships between keywords used to describe publications are asserted using the SKOS (Simple Knowledge Organization System) vocabulary.\myurl{https://www.w3.org/TR/2009/REC-skos-reference-20090818/}
For example, keywords are related to each other using the relation \concept{skos:related}.
Hyponyms and hypernyms are expressed by the relations \concept{skos:narrower} and \concept{skos:broader}.
Finally, the Ontology Alignment Evaluation Initiative\myurl{http://oaei.ontologymatching.org/} should be mentioned, which aims to achieve an established consensus for evaluating ontology matching methods.

\section{Querying of Linked Data}
\label{sec:distributeddataquerying}

Queries over Linked Data can be processed using link traversal~\cite{DBLP:conf/esws/Hartig11}, i.\,e., the query processor would use one of those IRIs given directly in the query as starting point and query the respective source for more triples involving the IRI. 
By iteratively doing this for more IRIs and with respect to the graph pattern defined in the query, a local set of triples is collected over which the given query can be evaluated.

More conveniently, queries over RDF and Linked Data can be formulated in SPARQL\footnote{\url{http://www.w3.org/TR/sparql11-query/}}, if a corresponding endpoint to the graph data is made available. 
Whereas such queries can target graphs that are stored in a single graph store, Linked Data often requires formulating and executing queries across multiple graphs that are stored at distributed data sources.

Below, we first introduce the basic query processing of SPARQL queries along with our running example.
This is followed by discussing RDFS/OWL entailment regimes and querying.
Finally, we present approaches for distributed querying over multiple SPARQL endpoints.

\subsection{Basic Query Processing}
\label{subsec:querying}

In principle, a SPARQL query is evaluated by comparing the graph pattern defined in the query to the RDF graph and reporting all matches as results. The set of results can be restricted by additional criteria, such as filters, i.\,e., conditions on variables and triple patterns that additionally need to be fulfilled.

\begin{figure}[ht]
  \centering
  \includegraphics[width=0.8\textwidth]{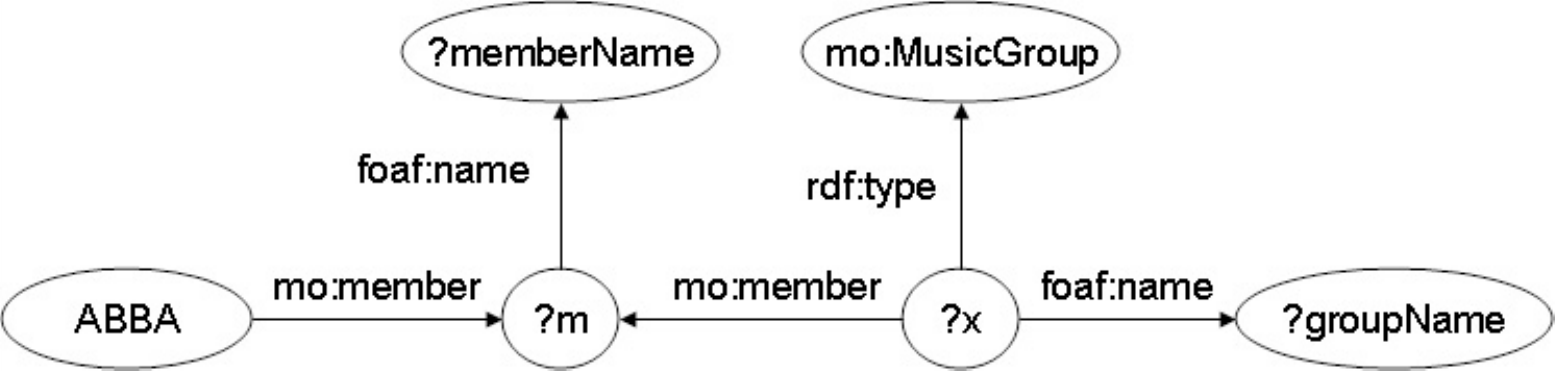}
\caption{Graphical representation of a query for music groups (represented by the variable \textit{?groupName}),
whose members are also members of ABBA. The variable \textit{?m} refers to the members of ABBA.
The vertex labeled ``ABBA'' represents the URI for ABBA.
The prefix \textit{mo} refers to the music ontology, \textit{foaf} to the FOAF ontology, and \textit{rdf} to the vocabulary of the RDF specification.
}
\label{fig:spaqrl_abba_members_graph}
\end{figure}

As an example, let us consider the query illustrated in Figures~\ref{fig:spaqrl_abba_members_graph} and~\ref{fig:spaqrl_abba_members} that we want to execute over our example MusicBrainz graph from Section~\ref{sec:musicscenario}. 
We are now interested in the musicians of ABBA who are also members of other bands.
If we follow the Linked Data principles and evaluate the query using link traversal~\cite{DBLP:conf/esws/Hartig11}, this would mean first querying for triples including the IRI that represents ABBA, then navigating to the individual band members, and then following the links to all of the members' bands and query more relevant triples.

\begin{figure}[ht]
{\small
\begin{verbatim}
PREFIX rdf: <http://www.w3.org/1999/02/22-rdf-syntax-ns#>
PREFIX mo: <http://purl.org/ontology/mo/>
PREFIX foaf: <http://xmlns.com/foaf/0.1/>
PREFIX bbc: <http://www.bbc.co.uk/music/>
SELECT ?memberName ?groupName
WHERE  { bbc:artists/d87e52c5-bb8d-4da8-b941-9f4928627dc8#artist mo:member ?m .
         ?x mo:member ?m .
         ?x rdf:type mo:MusicGroup .
         ?m foaf:name ?memberName .
         ?x foaf:name ?groupName }
FILTER (?groupName <>  "ABBA")
\end{verbatim}
}
\caption{SPARQL query for music groups whose members are also members of ABBA.
In the first triple pattern of the WHERE part, the URI of ABBA is the subject.
}
\label{fig:spaqrl_abba_members}
\end{figure}

Similar to relational database systems, there exist several dedicated graph stores (aka triple stores) that are optimized for RDF graphs and evaluating SPARQL queries.
Some of the most popular triple stores are RDF4J~\cite{iswc-Broekstra02sesame}, Jena~\cite{DBLP:conf/semweb/WilkinsonSKR03}, Virtuoso\myurl{http://virtuoso.openlinksw.com}, and GraphDB\myurl{https://graphdb.ontotext.com/}. 
They are building upon concepts and techniques known from relational database systems~\cite{DBLP:conf/rweb/HoseSTW11,DBLP:journals/pvldb/NeumannW08} and expand them with graph-specific optimizations~\cite{DBLP:conf/esws/VidalRLMSP10,DBLP:conf/www/StockerSBKR08,DBLP:conf/edbt/Gubichev014,DBLP:conf/kcap/LoustaunauH21,DBLP:conf/semweb/FerradaBH20}.

\subsection{Entailment Regimes and Query Processing}
\label{subsec:entailment}

In addition to explicitly querying existing facts, SPARQL provides inferencing support through so-called \emph{entailment regimes}. They correspond to logical consequences describing the relationship between the statements that are true when one statement logically follows from one or more statements.
Entailment regimes specify an entailment relation between well-formed RDF graphs, assuming that a graph $G$ entails another graph $E$ (denoted $G$ $\models$  $E$) if there is a logical consequence from $G$ to $E$. A regime extends the query of explicitly existing facts with facts that can be inferred using RDFS and OWL constructs (cf. Section~\ref{sec:reasoning}), such as the extension of facts about subclasses using \texttt{rdfs:subClassOf}.

Depending on the feature set of the respective SPARQL triple store, different (or even no) entailment regimes are supported.
They differ in terms of their power in the supported inference capabilities over RDFS/OWL classes and relationships.
SPARQL query engines such as GraphDB adopt a materialized approach, wherein they compute the closure of the input RDF graph $G$ over a set of relevant entailment rules $R$. Conversely, approaches grounded in query rewriting expand the SPARQL query itself rather than altering the RDF graph. Sub-queries are aligned with the entailment rules through backward chaining, and when the consequent of an entailment rule is matched, the antecedent of the rule is added to the query in the form of disjunctions.

Although both approaches yield equivalent answers for a given SPARQL query, their performance can diverge significantly. 
Materialized RDF query processing may outperform on-the-fly execution of the rewritten query, but it may consume more memory~\cite{DBLP:conf/aaai/GlimmKKS15}. Nevertheless, various optimization techniques have been proposed to mitigate the overhead caused by the on-the-fly evaluation of entailment regimes~\cite{DBLP:journals/ws/TrivelaSCS15}. These optimizations are required particularly in the presence of the \texttt{owl:sameAs}. This predicate corresponds to logical equivalence and involves the application of the Leibniz Inference Rule~\cite{DBLP:books/sp/GriesS93} to deduce all the equivalent triples entailed by equivalent resources based on \texttt{owl:sameAs} relation. 
This process may lead to many intermediate results, impacting the query engine's performance. Xiao et al.~\cite{DBLP:conf/esws/XiaoHBRGC18} propose query rewriting techniques to efficiently evaluate SPARQL queries with \texttt{owl:sameAs} employing equivalent SQL queries.

\subsection{Federated Query Processing}
\label{subsec:federatedQueries}

\begin{figure}[!ht]
\centering
\subfloat[Federated Query]{
\includegraphics[width=0.50\textwidth]{figs/QueryPlanA.pdf}}
 \subfloat[Bushy-Tree Plan]{
 \includegraphics[width=0.50\textwidth]{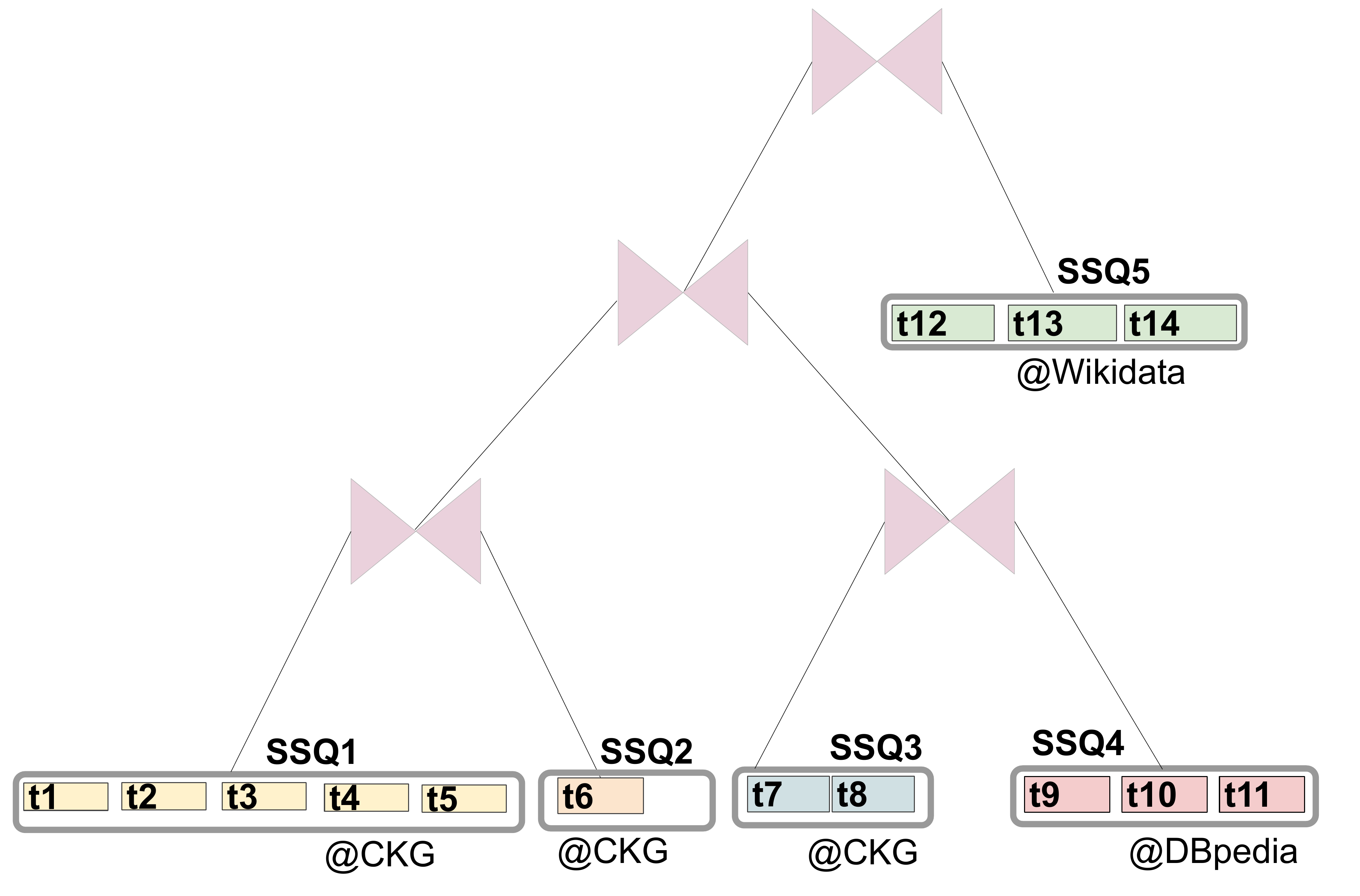}}
\caption{\textbf{Federated query.} a) SPARQL query comprising 14 triple patters to be executed over a federation including Cancer Knowledge Graph (CKG), DBpedia, and Wikidata. b) A query plan composed of five star-shaped subqueries SSQ1, SSQ2, SSQ3, SSQ4, and SSQ5 corresponding to the query decomposition. 
Each SSQ is executed over the graph that can answer the SSQ. The execution engine follows the query plan; the execution of four joins merges the SSQ answers and produces the federated query answer.}
\label{fig:federatedQuery}
\end{figure}

Federations provide another perspective on querying linked data over multiple sources.
A \emph{federation of knowledge graphs} shares common entities while potentially providing different perspectives on those entities. 
Each knowledge graph within the federation operates autonomously and can be accessed through various Web interfaces, such as SPARQL endpoints or Linked Data Fragments (LDFs)~\cite{DBLP:journals/ws/VerborghSHHVMHC16}. 
SPARQL endpoints offer users the ability to execute any SPARQL query against multiple SPARQL endpoints.
In contrast, LDFs enable access to specific graph patterns, such as triple patterns~\cite{DBLP:conf/semweb/VerborghHMHVSCCMW14a} or star-shaped graph patterns~\cite{DBLP:journals/corr/abs-2002-09172}, allowing retrieval of fragments from an RDF knowledge graph.
A \emph{star-shaped subquery} is a conjunction of triple patterns in a SPARQL query that share the same subject variable~\cite{DBLP:conf/esws/VidalRLMSP10}.
An LDF client can submit requests to a server, which then delivers results based on a data shipping policy and partitions results into batches of specified page sizes.
Query processing in a federation of graphs differs from querying a single source because it enables real-time data integration of graphs from multiple sources.
For example, \autoref{fig:federatedQuery} depicts a SPARQL query whose execution requires the evaluation of subqueries over three knowledge graphs: a Cancer Knowledge Graph (CKG)~\cite{DBLP:journals/semweb/AisoposJNPRSIVM23}, DBpedia, and Wikidata.
This query could not be executed over a single data source unless the three knowledge graphs were physically materialized into one. Subqueries with a specific shape (e.\,g., star-shaped subqueries) need to be identified and posed against the knowledge graph(s) that is able to answer a particular part of the query. The federated query engine has to decompose input queries into these subqueries, find a plan to execute them and collect and merge the answers from the subqueries to produce a federated answer.

A federated query engine is a system designed to execute queries over a federation of graphs. 
These engines leverage advanced query optimization methods to effectively decompose an input query into subqueries that can be executed over one or more graphs within the federation. 
Furthermore, they identify the most appropriate graphs for executing each subquery.
In addition to query execution, these engines also play a crucial role in discovering and managing efficient plans that minimize the data merging costs associated with collecting information from the selected knowledge graphs.

A federated SPARQL query engine typically follows a mediator and wrapper architecture, which has been established in previous research~\cite{Wiederhold92,ZadorozhnyRVUB02}.
Wrappers play a crucial role in translating SPARQL subqueries into requests sent to SPARQL endpoints, while also converting the endpoint responses into internal structures that the query engine can process.
The mediator, on the other hand, is responsible for rewriting the original queries into subqueries that can be executed by the data sources within the federation.
Additionally, the mediator collects and merges the results obtained from evaluating the subqueries to produce the final answer to the federated query.
Essentially, the mediator consists of three main components:
\begin{itemize}
  \item Source selection and query decomposition.
    This component decomposes queries into subqueries and selects the appropriate graphs (sources) capable of executing each subquery.
    Simple subqueries typically consist of a list of triple patterns that can be evaluated against at least one graph.
    Formally, source selection corresponds to the problem of finding the minimal number of knowledge graphs from the federation that can produce a complete answer to the input query.
    On the other hand, query decomposition requires partitioning the triple patterns of a query into a minimal number of subqueries, such that each subquery can be executed over at least one of the selected knowledge graphs.
    Commonly, federated query engines follow heuristic-based methods to solve these two problems.
    For example, for query decomposition, heuristics based on exclusive groups~\cite{DBLP:conf/semweb/SchwarteHHSS11} or star-shaped subqueries~\cite{DBLP:conf/esws/VidalRLMSP10,DBLP:journals/tlsdkcs/VidalCAMP16,DBLP:conf/semweb/MontoyaSH17} enable to efficiently solve source selection and query decomposition in queries free of general predicates (e.\,g., \texttt{owl:sameAs} or \texttt{rdf:type}).

Moreover, more general approaches (e.\,g., Endris et al.~\cite{DBLP:journals/tlsdkcs/EndrisGLMVA18}) resort to metadata describing the star-shaped patterns existing in a knowledge graph to perform query decomposition and source selection accurately.
An \emph{exclusive group} corresponds to a set of triple patterns (corresponding to a conjunction of the triples) in a SPARQL query that can be exclusively executed over one data source, as in FedX~\cite{DBLP:conf/semweb/SchwarteHHSS11}. 
Such a technique is implemented in the SPARQL engine FedX~\cite{DBLP:conf/semweb/SchwarteHHSS11}, which also optimizes the execution of \emph{star-shaped subqueries} by identifying groups of conjunctive triple patterns that a data source can execute independently from others.

  \item Query optimizer.
    This component identifies execution plans by combining star-shaped subqueries (SSQs) and utilizing physical operators implemented by the query engine.
    Formally, optimizing a query corresponds to the problem of finding a physical plan for the query that minimizes the values of a utility function (e.\,g., execution time or memory consumption). 
    To maximize the utility function, query optimizers consider plans with different orders of executing operators, alternative implementations of operators, such as joins, as well as particular execution alternatives for certain query types, e.\,g., queries involving aggregation~\cite{DBLP:conf/esws/IbragimovHPZ15}.
    In general, finding an optimal solution is computationally intractable~\cite{DBLP:journals/tods/IbarakiK84}, while the problems of constructing a \emph{bushy tree plan}~\cite{DBLP:conf/pods/ScheufeleM97} and finding an optimal query decomposition over the graphs~\cite{DBLP:journals/tlsdkcs/VidalCAMP16} are NP-Hard. 
    A bushy tree plan is a query execution plan that represents a query as a tree structure with multiple branches or subqueries, which can also be bushy-tree plans.
    Query plans can be generated following the traditional optimize-then-execute paradigm or re-optimize and adapt a plan on the fly according to the conditions and availability of selected graphs~\cite{DBLP:series/lncs/EndrisVG20}. 
    Alternatively, the query optimizer may resort to a cost model to guide the search on the space of query plans and identify the one that minimizes the values of the utility function~\cite{DBLP:conf/semweb/MontoyaSH17}.
  \item \emph{Query engine.}
    This component of a federated query engine implements the physical operators necessary to combine tuples obtained from the graphs.
    These physical operators are designed to support logical SPARQL operations such as JOIN, UNION, or OPTIONAL~\cite{DBLP:journals/tods/PerezAG09}.
    Physical operators can be empowered to adapt execution schedulers to the current conditions of a group of selected graphs. 
    Thus, adaptivity can be achieved at the intra-operator level, where the operators can detect when graphs become blocked or data traffic bursts.
    Additionally, intra-operator opportunistically produce results as quickly as data arrives from the graphs, and can produce results incrementally. 
    Some opportunistic approaches~\cite{DBLP:conf/semweb/AcostaVLCR11,DBLP:conf/sigmod/HoseS12,DBLP:conf/i-semantics/GalkinEACVA17,DBLP:conf/semweb/AcostaV15} combine producing results quickly in an incremental fashion with greedy source selection so that the system stops querying additional graphs once the user's wishes, e.\,g., in terms of the minimum number of obtained results, are fulfilled. 
    The physical operators implemented by the SPARQL federated query engine ANAPSID~\cite{DBLP:conf/semweb/AcostaVLCR11} implement intra-operator adaptivity, enabling results generation even when the graphs became blocked.
    On the other hand, inter-operator adaptive strategies can produce an answer as soon as it is computed and can keep producing intermediate results even when data from a source becomes blocked. 
    Acosta and Vidal~\cite{DBLP:conf/semweb/AcostaV15} propose adaptive query processing techniques that enable the re-ordering of a query execution plan to adjust execution schedulers to unexpected environmental conditions (e.\,g., changes at rates at which tuples arrive from graphs or a graph's availability).
\end{itemize}

During the query optimization process, a plan is generated as a bushy tree that comprises four join operators.  
This is shown in Figure~\ref{fig:federatedQuery}.

The problem of query execution is defined in terms of data shipping strategies that are able to distribute query load between a server-client architecture of Linked Data Fragments.
State of the art approaches tackle this issue by employing graph partitioning techniques to generate graph pattern fragments.
These fragments aim to improve query performance, scalability, and workload balance. 
SAGE~\cite{DBLP:conf/f-ic/MinierSM19}, smart-KG~\cite{DBLP:conf/www/AzzamFABP20}, brTPF~\cite{DBLP:journals/corr/HartigA16}, SkyTPF~\cite{DBLP:conf/semweb/KelesH19}, SPF~\cite{DBLP:journals/corr/abs-2002-09172}, TPF~\cite{DBLP:journals/ws/VerborghSHHVMHC16}, and WiseKG~\cite{DBLP:conf/www/AzzamAMKPH21} are exemplary approaches for accessing graph patterns of varying shapes.
These interfaces for Linked Data Fragments may significantly accelerate query execution, reduce workload, and minimize data transfer.
Moreover, query rewriting based on these fragments correctly produces all the answers of the rewritten query, i.\,e., the methods are sound and complete.
However, it is important to note that there exists a trade-off between memory consumption and execution time improvement so that storage requirements (e.\,g., smart-KG) can be twice as high in comparison to other approaches (e.\,g., SAGE, TPF, or SPF), due to the need to store partitions and their associated metadata. 

Because of the variety of available interfaces, some works have proposed methods for heterogeneous federations~\cite{DBLP:conf/www/HelingA22,DBLP:conf/semweb/MontoyaKH19,DBLP:conf/semweb/MontoyaAH18a,DBLP:conf/esws/ChengH22}, where the goal is to find efficient query execution plans in consideration of the constraints as well as strengths of the available interface.

\section{Trustworthiness and Provenance of Graph Data}
\label{sec:trustprovenance}
Trustworthiness of web pages and data on the web can be detected by various indicators, e.\,g., by certificates, by the placement of search engine results, and by links (forward and backward links) to other pages.
However, on the Semantic Web, there are few ways for users to assess the trustworthiness of individual data. 
Rules can be utilized to define policies and business logic over the web of data, and transparently used to infer data that validate or do not validate these policies.
The trustworthiness of inferred data can be assessed through its provenance, which encompasses metadata detailing how the data was acquired and verified~\cite{DBLP:conf/rr/KhandelwalJK11}.

The trustworthiness of data on the web can be inferred from the trustworthiness of other users (``Who said that?''), the temporal validity of facts (``When was a fact described?''), or in terms of uncertainty of statements (``To what degree is the statement true?'').
Artz and Gil~\cite{DBLP:journals/ws/ArtzG07} summarize trustworthiness as follows:
``Trust is not a new research topic in computer science, spanning areas as diverse as security and access control in computer networks, reliability in distributed systems, game theory and agent systems, and policies for decision-making under uncertainty. The concept of trust in these different communities varies in how it is represented, computed, and used.''
Although trustworthiness has long been considered in these areas, the provision and publication of data by many users to multiple sources on the Semantic Web introduces new and unique challenges.

One way of facilitating trust on the Semantic Web is to capture and provide the provenance of data with the PROV ontology (PROV-O)\footnote{\url{https://www.w3.org/TR/prov-o/}}.
It captures information about which \emph{Agents} cause data \emph{Entities} to be processed by which \emph{Activities}.
Capturing such information requires the use of known tools for modeling metadata for RDF data, e.\,g., reification, singleton properties, named graphs, or RDF-star\footnote{\url{https://w3c.github.io/rdf-star/}}.
While some approaches use these constructs to capture provenance information for each triple individually~\cite{DBLP:conf/semweb/GalarragaMH17}, 
others exploit the fact that typically multiple triples share the same provenance~\cite{DBLP:conf/semweb/HansenLGLTH20} so that they can be combined into the same named graph encoding the provenance information only once for a set of triples.
Delva et al.~\cite{DBLP:conf/edbt/DelvaDJB23} introduce the notion of shape fragments, which entail the validation of a given shape through the neighborhood of a node, along with the node's provenance and the rationale behind its validation.

Furthermore, trustworthiness also plays a role in inference services on the Semantic Web, as data inference must consider specifications related to trustworthiness and data must be evaluated for trustworthiness.
Important aspects for trustworthiness of data include~\cite{DBLP:journals/ws/ArtzG07}:
the origin of the data, 
trust already gained based on previous interactions, 
ratings assigned by policies of a system, and 
access controls and, in some cases, security and importance of information.
These aspects are realized in different systems.

In general, data provenance and trustworthiness of data on the Semantic Web have been addressed for RDF data \cite{DBLP:journals/ki/DividinoSSS09,DBLP:conf/semweb/FlourisFPTC09} as well as for OWL and rules in \cite{DBLP:journals/ki/DividinoSSS09}.
In addition, there are some recent approaches on supporting how-provenance for SPARQL queries~\cite{DBLP:journals/jacm/GeertsUKFC16,DBLP:journals/pvldb/HernandezGH21,DBLP:conf/www/Galarraga0KH23} with the goal of providing users with explanations on how the answers to their queries were derived from the underlying graphs.
Other work deals with access controls over distributed data on the Semantic Web~\cite{DBLP:conf/esws/GavriloaieNOSW04}.
Furthermore, there are approaches to computing trust values~\cite{DBLP:journals/jair/StoilosSPTH07} and informativeness of subgraphs~\cite{DBLP:conf/cikm/KasneciEW09}.
There are also digital signatures for graphs~\cite{DBLP:conf/ecai/BellomariniNS20}.
Analogous to digital signatures for documents, entire graphs or selected vertices and edges of a graph are provided with a digital signature to ensure the authenticity of the data and thus detect unauthorized modifications~\cite{DBLP:phd/dnb/Kasten16}.
In the approach for digital graph signatures developed by Kasten et\,al., graph data on the Web is supported in RDF format as well as in OWL~\cite{DBLP:conf/esws/KastenSS14}.
The digital graph signature is itself represented as a graph again and can thus be published together with the data on the Web.
The link between the signature graph and the signed graph is established by the named graph mechanism~\cite{DBLP:conf/esws/KastenSS14}, although other mechanisms are also possible.
Through this mechanism, it is possible to combine and nest signed graphs.
It is thus possible to re-sign already signed graphs together with other, new graph data, etc.
This makes it possible to build complex chains of trust between publishers of graph data and to be able to prove the origin of data~\cite{DBLP:conf/esws/KastenSS14,DBLP:phd/dnb/Kasten16}.

    
\section{Machine Learning on Knowledge Graphs}
\label{sec:machine-learning-on-knowledge-graphs}
\index{Semantic Web!Machine learning on knowledge graphs}

Machine learning on graphs is one of the most active fields of research in AI~\cite{hamilton-book}.
We will first consider graph embeddings for shallow representations of nodes, edges, and graphs.
Then we turn our attention to neural networks for graphs (GNNs), which can compute more complex representations.
Finally, we discuss developments in language models and their relationship to and integration with knowledge graphs.
    
\subsection{Graph Embeddings}
\index{Graph Embeddings}

Graph embeddings are used to compute representations of nodes, edges, and graphs for use in machine learning tasks, such as input for GNNs.
The basic idea behind embeddings is to use unsupervised learning to determine a representation of nodes, for example, via their neighborhood relationships in a high-dimensional space.

\subsubsection{Classical Graph Embeddings}
Classic approaches to the unsupervised calculation of graph embeddings include Node2Vec~\cite{node2vec}, which uses random graph traversal to calculate how similar the start node is to nodes in its neighborhood.
However, graph embeddings also include self-supervised aspects, such as in Deep Walk~\cite{deepwalk}, a generalization of Node2Vec using a weighted breadth-first versus depth-first search.
In Deep Walk, random traversals in the graph are also calculated starting from a node.
The embeddings are then calculated using, among other things, the skip-gram idea from the embedding model Word2Vec~\cite{word2vec} for texts. 
The latter provides for an element to be removed from a sequence in a self-supervised approach and predicted by a neural network.
In the context of Deep Walk, the sequences are the random traversals in the graph and the elements are the nodes in the graph.
An overview of classical graph embeddings can be found in Hamilton~\cite{hamilton-book} and Murphy~\cite{murphy-book}, among others.

\subsubsection{Embeddings for Knowledge Graphs}
Specific embedding models for knowledge graphs include RDF2Vec~\cite{rdf2vec} and a variety of approaches based on it.
In general, embeddings for knowledge graphs can be classified based on Biswas et al.~\cite{biswas_et_al:TGDK.1.1.4} into translation-based models based on a distance function between nodes, models that model probabilities and uncertainty, and approaches based on semantic similarity determined by tensor decomposition.

Examples include TransE~\cite{transe} as a translation-based model for representing nodes and edges in a common embedding space.
TransR~\cite{transr} extends TransE to support 1:N and other relationships.
It uses separate embedding spaces, i.\,e., a separate matrix for each relation~\cite{biswas_et_al:TGDK.1.1.4}, and therefore requires more memory.
While TransR is limited to the immediate neighborhood, PTransR extends this to include the encoding of paths.
TransG~\cite{transg} is based on a mixed model of Gaussian distributions to map different meanings of a relation such as \texttt{hasPart}.

While these embedding models operate on knowledge graphs, they do not map the specific semantics of Semantic Web standards such as RDFS and OWL.
One approach in this direction is RotateE~\cite{rotate}, which uses rotational transformations to map complex relationships in knowledge graphs~\cite{biswas_et_al:TGDK.1.1.4}, such as symmetric and antisymmetric relations, inversion, and composition.
Graph embedding models that can map parts of the semantics of OWL ontologies are Trans\-OWL~\cite{transowl} or OWL2Vec~\cite{owl2vec}. 
These map the semantics of description logic expressions such as the intersection of two classes, subclasses, etc. directly in the training of the embeddings using special loss functions.
TransOWL extends previous embedding models by encoding not only the directly observed graph, but also knowledge derived from conclusions based on available axioms. 
To do this, TransOWL integrates the OWL axioms directly into the loss function, effectively performing inferences during training. 
TransOWL is derived from TransE, while its variant TransROWL is based on TransR. 
In addition, there is also a version that extends TransOWL with equivalence axioms and inverse relations. 
Another model for OWL is OWL2Vec*, which uses random walks and word embeddings to encode the semantics of OWL ontologies.

\subsubsection{Discussion of Graph Embeddings}

Graph embeddings have a number of disadvantages~\cite{hamilton-book}.
First, the calculations are slow due to high redundancy.
Second, the embeddings of a graph cannot be easily transferred to new nodes or even new graphs without further learning.
Therefore, graph embeddings are inherently transductive, meaning they assume that the entire graph is available at training time and can be used for unsupervised (Node2Vec) or self-supervised learning (Deep Walk).
Therefore, the embeddings that have already been calculated cannot be transferred to new nodes.
The inefficiency caused by redundant calculations stems from the fact that a separate embedding must be determined for each node via random traversals.
In doing so, the same nodes are often visited multiple times for different embeddings.
It should also be mentioned that the primary goal of graph embeddings is to determine a suitable representation of graphs when dataset-specific features are not available or are to be enriched. 
Therefore, embeddings do not usually have a specific downstream task such as node classification.

\subsection{Neural Networks for Graphs}
\index{Neural networks for graphs}

Graph Neural Networks (GNNs) extend the concept of graph embeddings by enabling more complex representations to be calculated based on the node features.
In contrast to graph embeddings, GNNs are typically characterized by supervised or semi-supervised learning.
The origins of GNNs can be traced back at least to the work of Gori et al.~\cite{gorietal2005} and Scarselli et al.~\cite{scarsellietal2009}.

\subsubsection{Categories of Graph Neural Networks (GNNs)}
GNNs can be classified as isotropic and anisotropic, as well as transductive and inductive~\cite{DBLP:journals/nn/GalkeVFZHS23}.
The majority of GNNs are based on the principle of message passing, in which features of neighboring nodes are used to compute representations of the nodes.
In isotropic GNNs, all edges have the same influence on the calculation of node representations, while in anisotropic models, the edge weights and thus the representations of neighboring nodes are different or are learned.

As described above, transductive GNNs assume that the unlabeled nodes from the test set are also available at training time. 
These can be used in various calculations, for example for message passing, but have no influence on the error function of the model and are therefore not taken into account in the learning process using backpropagation.
Examples of GNNs are GCN (isotropic and transductive), which is motivated by spectral analysis in graph theory~\cite{gcn}.
GAT (anisotropic and inductive) learns the edge weights using attention~\cite{gat}.
GraphSAGE (isotropic and inductive) uses a sampler, among other things, and thus scales for large graphs.
When the hyperparameters are optimized, these models are powerful representatives of GNNs with very good performance~\cite{pitfalls,DBLP:journals/corr/abs-2406-08993-log2024}.

In classical GNNs such as GCN, message passing is used to efficiently calculate embeddings for all nodes in the graph simultaneously~\cite{hamilton-book}.
As mentioned above, GNNs are typically computed through supervised or semi-supervised learning and involve a downstream task such as node or graph classification, graph regression, and edge prediction (further applications are described in Murphy~\cite{murphy-book}).
Learning using GNNs is often referred to as semi-supervised because in many datasets and methods, only a few nodes per class are labeled with a gold standard label~\cite{hamilton-book}; for example, less than $5\%$ of the nodes have a label when training a GCN~\cite{gcn}.

\subsubsection{Oversquashing and Oversmoothing in GNNs}
One problem with classical GNNs is that long paths cannot be mapped, leading to oversquashing and oversmoothing~\cite{hamilton-book}.
Oversquashing refers to when the representations of distant nodes are ``squashed'' together over only a few edges, leading to a loss of important information for calculating the representation of the target node.
Oversmoothing refers to the incremental alignment of all representations in the GNN when, after many message-passing steps, the representations of the nodes lose their individual features.
These problems can be effectively addressed by inserting additional learnable weights between the GNN layers, as in the GCNII model~\cite{gcnii}.
Other approaches, such as Graph-MLP, no longer depend on message passing and apply contrastive learning~\cite{graph-mlp}.

\subsubsection{Homophilic and Heterophilic Graphs}

Another discussion surrounding GNNs concerns the distinction between homophilic and heterophilic graphs, i.\,e., whether neighboring nodes are highly likely to belong to the same class or to different classes~\cite{DBLP:conf/iclr/0001LST22-homophily-necessary}.
While models such as GCN assume that homophily is important for good performance, later work concludes that graphs can also be heterophilic as long as the neighborhood distribution of classes is consistent~\cite{DBLP:conf/iclr/0001LST22-homophily-necessary}.
A good overview of the most important GNNs can be found in Hamilton~\cite{hamilton-book} and Murphy~\cite{murphy-book}, among others.

\subsubsection{GNNs for Knowledge Graphs}

Specific neural networks for knowledge graphs include R-GCN~\cite{r-gcn}, an extension of GCN for learning relation-specific adjacency matrices.
Since the number of relations in knowledge graphs can be very large~\cite{DBLP:journals/dpd/GottronKS15}, R-GCN approximates the relation-specific adjacency matrices using a linear combination of fewer base matrices.
KGAT is a graph attention neural network based on a hybrid combination of user-item graphs and knowledge graphs~\cite{kgat}.
The goal is to model higher-order relationships such as user-book-author-user and user-book-genre-user, and it is used for recommendation tasks.

A detailed overview of neural networks for knowledge graphs can be found, e.\,g., in Kumar et al.~\cite{DBLP:journals/access/YeKSSW22-kg-survey}.
A discussion of future topics can be found in d'Amato et al.~\cite{DBLP:journals/tgdk/dAmatoMMS23}.

\section{Language Models and Knowledge Graphs}
\label{sec:language-models-and-knowledge-graphs}
\index{Language Models and Knowledge Graphs}
\index{Semantic Web!Language Models and Knowledge Graphs}

Language models based on neural networks, in particular the Transformer architecture~\cite{DBLP:conf/nips/VaswaniSPUJGKP17}, are often distinguished into encoder models (e.\,g., BERT~\cite{bert}), encoder-decoder models (such as T5~\cite{t5}), and decoder models (e.\,g., Llama~\cite{llama3} and GPT~\cite{gpt4}).
These differ in their capabilities and primary uses.
Encoder models are particularly suitable for classification~\cite{DBLP:conf/acl/GalkeS22}, while decoder models dominate the field of generative AI and chatbots~\cite{are-we-makin-progress}.
However, all models share characteristics that are summarized under the term \textit{foundation models}~\cite{foundation-models}.
These include pre-training using self-supervised learning, the use of a broad corpus as training data, and the associated applicability or adaptability to a variety of downstream tasks.
The size of the language models ranges from several hundred million parameters (BERT) to billions of parameters (Llama, GPT).
%
These are therefore also referred to as small language models versus large language models.

\subsection{Graph Transformer, Language Models, and Retrieval Augmented Generation}

Adaptations of the transformer architecture to graphs, such as graph transformer networks~\cite{graph-transformer-networks}, have already shown strong performance on downstream tasks in 2021.
However, initial approaches were not yet based on pre-training models and were therefore not transferable to different downstream tasks. 
One reason for this is that the features between the datasets are too specific and also have different dimensionalities.
Some early approaches to integrating knowledge graphs with language models have therefore aimed to encode the facts from knowledge graphs directly into the parameters of language models.
Interesting approaches in this direction include SKILL~\cite{DBLP:conf/naacl/MoiseevDAJ22-skill}, in which the language model T5 was trained on a textual representation of factual knowledge from knowledge graphs. 
SKILL shows that pre-training T5 on the Wikidata knowledge graph~\cite{DBLP:journals/cacm/VrandecicK14} leads to better downstream performance.
The knowledge graph ``disappears'' into the model parameters.
Changing the knowledge graph or deleting facts therefore becomes very laborous.
ATLAS~\cite{DBLP:journals/jmlr/IzacardLLHPSDJRG23-atlas} is a retrieval-augmented method (RAG) for integrating knowledge-intensive tasks into a language model with few training examples.
The goal of ATLAS is to learn facts and use them, e.\,g., for fact checking or asking trivia questions.
An overview of methods for editing knowledge graphs in language models can be found, e.\,g., in Cao et al.~\cite{DBLP:journals/ijautcomp/CaoLHS24}.

Due to the text-based nature of language models such as T5, it was necessary for the graphs to be represented in text form.
Features such as TF-IDF in GCN datasets cannot be well represented in so-called \textit{text-attributed graphs}.
Knowledge graphs in RDFS or OWL, on the other hand, are well suited due to their descriptive properties of nodes and edges.
A collection of text-based graphs can be found in TAGLAS~\cite{DBLP:journals/corr/abs-2406-14683-taglas}, among others.

\subsection{Graph Foundation Models}

Under the term \textit{graph foundation models}, there are a number of approaches for graphs that meet the above requirements.
The models differ in terms of their architecture, i.\,e., whether they are based on an existing language model, a GNN, or a combination of language model and GNN~\cite{graph-foundation-models-survey}. 
The Graph Language Model (GLM)~\cite{glm}, which is based on Google's T5 language model, belongs to the first category.
The attention mechanism of T5's transformer layer has been modified so that masking takes into account the adjacency of nodes. 
This modification of the attention mask also distinguishes GLM from the above-mentioned approaches such as SKILL.
For T5, the input graph is converted into a Levi graph and then processed as text, like any other prompt, in the T5 model. 
Due to the nature of GLM, it is suitable for processing both text and graphs natively, but is limited to text-attributed graphs.

Models such as PRODIGY, PSP, and GraphAny pursue an approach based on GNNs.
The first two, PRODIGY~\cite{prodigy} and PSP~\cite{psp}, can be understood as native graph-based foundation models that map a graph prompt in analogy to a text prompt. 
Based on a pre-trained model, similar to language models, several graphs can be given as examples (\textit{context learning}) along with the task to be solved.
Unlike PRODIGY, PSP works with prototypes that represent the graphs.
GraphAny~\cite{graphany} uses a combination of different linear GNNs based on the Simplified GCN Model (SGC)~\cite{sgc}.
In order to meet the requirement of being usable for various downstream tasks, GraphAny has a mechanism for handling different dimensionalities of node features in graph datasets.
GraphAny is very efficient due to the use of linear models.
In addition, training on a few examples is sufficient to achieve strong performance on other datasets with different features.
Approaches in the field of combining language models and GNNs include~\cite{opengraph,ofa,gofa}.

An overview of graph foundation models is provided by Liu et al.~\cite{DBLP:journals/corr/abs-2310-11829-graphfoundationmodels-survey}
and Fan et al.~\cite{DBLP:journals/corr/abs-2404-14928}.
	

\section{Applications}
\label{sec:userinterfaces}
\label{sec:examples}

With the increasing spread and use of semantic and linked data on the Web, the requirements for Semantic Web applications have increased at the same time as their application possibilities.
The general requirements for applications based on semantic data on the Web are given by their flexible and diverse representation and descriptions.
Applications that use data from relational databases or XML documents can start from a fixed schema. 
However, this cannot be assumed for data on the Web. 
Often, neither the data sources nor the type and amount of data in a source are fully known.
The dynamics of semantic data on the Web must be taken into account by applications accordingly, both when querying and aggregating data, and when visualizing data.
Thus, the real challenge of Semantic Web applications is to guarantee the best possible flexibility of the application to take into account the dynamics of data sources, data, and schemas during input, processing, and output.

In the following, selected examples of Semantic Web applications or application areas are presented. 
They illustrate how flexibility and quality of search, integration, aggregation, and presentation of data from the Web can be implemented. 
At the same time, they show the potential of Semantic Web applications.
First, uniform vocabularies and schemas are presented using the example of \textit{schema.org}.
These serve as a basis for semantic search to provide search engines with information about the meaning of web document content.
The search and integration of data from different sources is supported by \textit{Sig.ma}, a semantic web browser.
Other applications provide semantic search through other representation formalisms, e.\,g., \textit{Knowledge Graphs}).
Subsequently, the \textit{Facebook Graph-API}, an application programming interface (\textit{API}) to the Facebook (Knowledge) Graph, is introduced.

\subsection{Vocabularies and Schemas: Schema.org}

In HTML documents, the structure and composition of pages can be described with tags, but not the meaning of the information.
Vocabulary, schemas, and microdata can be used as mark-up in HTML documents to describe information about page content and its meaning in a way that search engines can process this information.

Schema.org\myurl{http://schema.org} is a collection of vocabularies and schemas to enrich HTML pages with additional information.
The vocabulary of \textit{Schema.org} includes a set of classes and their properties. A universal class ``thing'' is the most general, which is a kind of umbrella term for all classes.  
Other common classes are \textit{Organization}, \textit{Person}, \textit{Event}, and \textit{Place}.
Properties are used to describe classes in more detail.
For example, a person has the properties such as name, address, and date of birth.

In addition to vocabularies, Schema.org also specifies the use of HTML microdata, with the goal of representing data in HTML documents in as unambiguous a form as possible so that search engines can interpret it correctly.
An example of this is formats for unique dates and times, which can also describe intervals to indicate the duration of events.

Schema.org is supported by the search engines Bing, Google, and Yandex, among others.
There are extensions and libraries for various programming languages, including PHP, JavaScript, Ruby, and Python, to create web pages and web applications using vocabularies and microdata from Schema.org. 
Likewise, there are mappings from Schema.org vocabularies and microdata to RDFS.

\subsection{Semantic Search}

A classic web browser enables the display of web pages.
A semantic web browser goes one step further by additionally allowing the user to visualize the underlying information of individual pages, for example in the form of RDF metadata. 
Semantic Web browsers are also referred to as hyperdata browsers because they allow navigation between data while also allowing one to explore the connection to information about that data.
Thus, ordinary users can use and exploit Semantic Web data for their information search.

Sig.ma~\cite{key:sigma} was an application for (browsing) Semantic Web data, which may come from multiple distributed data sources.
Sig.ma provided an API for automatically integrating multiple data sources on the Web.
The requested data sources describe information in RDF.
A search in Sig.ma was initiated by a textual query from the user.
Entities such as people, places, or products can be searched for.
Results of a query are presented in aggregated form, that is, properties of the searched entity, such as a person, are presented in aggregated form from different data sources. 
For example, in a person search, information such as e-mail address, address, or current employer can be displayed.
In addition to the actual information, links to the underlying data sources are also displayed to allow users to navigate to refine their search.
Sig.ma also supported structured queries in which specific characteristics can be requested for an entity, such as contact information for a specific person.

Queries to data sources occur in parallel. 
The results from each data source in the form of RDF graphs are summarized by using properties of links in RDF data, such as \texttt{owl:sameAs}, or inverse-functional predicates.
When searching data sources, techniques such as indexes, logical inference, and heuristics are used for data aggregation.
OntoBroker\myurl{https://www.semafora-systems.com/ontobroker-and-ontostudio-x}~\cite{Decker1999} and OntoEdit~\cite{10.1007/3-540-48005-6_18} are ontology editors with search and inference systems for ontologies. 
Using OntoBroker, complex queries over distributed Semantic Web resources, e.\,g., represented in OWL, RDF, RDFS, SPARQL, and also F-Logic) can be efficiently processed.

\subsection{Knowledge Graphs and Wikidata}

There is an increasing number of knowledge bases and representations of structured data.
For example, the secondary database Wikidata\myurl{https://www.wikidata.org/wiki/Wikidata:Introduction/de}~\cite{DBLP:journals/cacm/VrandecicK14}.
A secondary database includes, in addition to the (actual) statements, relationships to their sources and other databases (called secondary information). 
Wikidata is a shared database between Wikipedia and Wikimedia. 
Wikidata mainly contains a collection of objects, which are represented as triples over the objects' properties and the corresponding values. 
Semantic MediaWiki\myurl{https://www.semantic-mediawiki.org/wiki/Semantic_MediaWiki} is an extension of MediaWiki. 
It serves as a flexible knowledge base and knowledge management system. 
Semantic MediaWiki extends a classic wiki with the ability to enrich content in a machine-readable way using semantic annotations. 
  
Another knowledge base was Freebase~\cite{DBLP:conf/sigmod/BollackerEPST08-freebase,DBLP:conf/aaai/BollackerCT07-freebase}, also an open and collaborative platform initiated in 2007 and acquired by Google in 2010.
The content from Freebase was taken from various sources, including parts from the MusicBrainz ontology mentioned earlier.
The success and widespread use of Wikidata prompted Google to migrate Freebase to Wikidata~\cite{DBLP:conf/www/TanonVSSP16}.
This strengthened the goal to develop a comprehensive, collaborative basis of structured data.

Google offers a semantic search function with Google Knowledge Graph\myurl{http://www.google.com/insidesearch/features/search/knowledge.html}$^,$\myurl{https://developers.google.com/knowledge-graph/}.
A knowledge graph, like an RDF graphs, is a set of triples representing links between entities.
This forms a semantic database.
Possible entity types are described on schema.org, among others.
If a search term occurs in a query, the corresponding entity is searched for in the knowledge graph. 
Starting from this entity, it is then possible to navigate to further entities by means of the links. 

\subsection{API-Access to Social Networks} 

A social network is essentially a graph in which connections are formed from users to other users, e.\,g., in the form of a friendship relationship or to events and groups.
Facebook's Graph API describes a programming interface to the Facebook Graph (called \textit{Open Graph}).
Within the graph, people, events, pages, and photos are represented as objects, with each object having a unique identifier.
For example, \url{https://graph.facebook.com/abba} is the identifier of ABBA's Facebook page.
There are also unique identifiers for the possible relationship types of an object, which allow navigating from one object to all connected objects with respect to a particular relationship.

The Graph API allows one to navigate the Facebook Graph and read objects, including their properties and relationships to other objects, as well as creating new objects in the Facebook Graph and deploying applications.
The API also supports requests for an object's metadata,
such as \textit{when} and \textit{by whom} an object was created.

\section{Impact for Practitioners}
\label{sec:praxis}
\label{sec:impact}

Linking and using graph data on the Web has become a widespread practice. 
Today, there is a large amount of open data in various formats and domains, such as bibliographic information management, bioinformatics, and e-government.
DBpedia is the central hub in this context, around which different datasets and domains are grouped (cf.~\cite{DBLP:journals/expert/Bizer09}).
This is illustrated, e.\,g., by the tremendous growth of the Linked Open Data Cloud\footnote{The growth of the Linked Open Data Cloud is documented at: \url{http://linkeddata.org/}.} since 2007.
Two of the latest notable supporters of graph-based data are online auctioneer eBay with their graph database\footnote{\url{https://github.com/eBay/akutan}} and the U.S. space agency NASA with the unification of internal distributed case databases as knowledge graphs\footnote{\url{https://blog.nuclino.com/why-nasa-converted-its-lessons-learned-database-into-a-knowledge-graph}}.
These and other success stories of the Semantic Web in industries and industry-scale knowledge graphs are described by Noy et al.~\cite{DBLP:journals/cacm/NoyGJNPT19}.
Further analyses and surveys arguing about the importance but also challenges of using graph data can be found in the literature like the 2020 survey of Sahu et al.~\cite{DBLP:journals/vldb/SahuMSLO20} and the 2021 reflection about the future of graphs by Sakr et\,al.~\cite{DBLP:journals/cacm/SakrBVIAAAABBDV21}.
The usefulness of knowledge graphs and semantic-based data modeling for complex systems is also discussed in the 2024 book by Abonyi et al.~\cite{AbonyiEtAl2024OntologiesInIndustry}.
The importance of graph databases is also reflected by the Forbes business magazine, which predicted in 2019 that graph databases would be the next mainstream database technology\footnote{\url{https://www.forbes.com/sites/cognitiveworld/2019/07/18/graph-databases-go-mainstream}}.

Regarding lightweight open graph data, Schema.org defines schemas for modeling data on web pages to provide information about the underlying data structures and meaning of the data.
Search engines can use this additional information to better analyze the content of web pages.
As mentioned above, Schema.org is supported by search engines such as Bing, Google, and Yandex. 
Studies on selected sources have shown that web pages among the top 10 results have up to 15~\% higher click-through rate\footnote{\url{http://developer.yahoo.net/blog/archives/2008/07/}}.
Other companies like BestBuy.com even report up to 30~\% higher click-through rates since adding semantic data to their websites (cf.~Section~\ref{sec:userinterfaces}) in 2009.
BestBuy.com uses the GoodRelations vocabulary\myurl{http://www.heppnetz.de/projects/goodrelations/} to describe online offers.
Similarly, Google uses semantic data from online commerce portals that use the GoodRelations vocabulary and takes it into account when searching\footnote{\url{http://www.ebusiness-unibw.org/wiki/GoodRelationsInGoogle}}. 

Another success is the publication of government data.
For example, the U.S. government makes government data publicly available with data.gov\myurl{http://www.data.gov/}, and U.S. Census\myurl{http://www.rdfabout.com/demo/census/} publishes statistical data about the United States.
In the UK, data.gov.uk\myurl{http://data.gov.uk} is a key part of a program to increase data transparency in the public sector.
The European Commission operates data.europa.eu\myurl{https://data.europa.eu/}, a European data portal with metadata about the member states.
Among others, it provides a SPARQL endpoint to access the data.

Finally, a strong growth of semantic biomedical data on the Web can be noted.
As part of Bio2RDF\myurl{http://bio2rdf.org/}, many bioinformatics databases have been linked.
Transinsight GmbH offers the knowledge-based search engine GoPubMed\myurl{http://www.gopubmed.org/} to find biomedical research articles.
Ontologies are used for searching.

Regarding more heavyweight ontologies in OWL, there has also been movement in recent years. 
In addition to numerous research-derived inference engines such as Pellet and Hermit mentioned above, inference mechanisms for OWL can now be found in commercial graph databases such as neo4j\footnote{\url{https://neo4j.com/blog/neo4j-rdf-graph-database-reasoning-engine/}}.
Furthermore, pattern-based core ontologies can also be found in software development workflows~\cite{DBLP:conf/semweb/SchonteichKS18}.
The development and use of core ontologies is part of a continuous delivery process that is used in practice.

\section{Summary and Outlook}

The Semantic Web consists of a variety of techniques that have been heavily influenced by long-term artificial intelligence research and its results.
The current state is also driven by an industry uptake under the umbrella term of Knowledge Graphs and reflected in various activities as described.
In summary, therefore, it can be observed that semantic data on the Web is having a real impact on commercial providers of products and services, as well as on governments and public administrations.

Despite all the research and industrial developments, the full potential of the Semantic Web has not yet been exploited.
Some important components of the Semantic Web architecture are still being explored, such as data provenance and trustworthiness.
Below, we describe three example directions for future work.
\begin{itemize}
    \item Neuro-symbolic systems: As mentioned in the introduction, we see as an important direction of future work the combination of symbolic AI and subsymbolic AI.
    By combining the strength of Large Language Models (LLM), i.\,e., generative AI, in processing and generating natural language text and accessing structured data and logical reasoning capabilities of the Semantic Web, a next step towards the vision of automated agents that perform complex planning tasks may be reached.
    An example is performing A* search with an LLM~\cite{DBLP:journals/corr/abs-2310-13227-toolchainstar}.
 Specifically, LLMs might comprehensively capture and acquire human knowledge~\cite{10.1145/3608966}, but current LLMs lack responding to simple questions of non-existing facts in their training data~\cite{10.1145/3608966}, may not contain all facts~\cite{DBLP:journals/corr/abs-2308-10168}, and thus return less accurate answers~\cite{Hou2023-fo}.
To leverage the distinct capabilities of both LLMs and the Semantic Web, the integration of neuro-symbolic systems appears to offer a viable solution~\cite{DBLP:journals/corr/abs-2306-08302}. 
Neuro-symbolic systems could also address the problem that LLMs' output is based on the most probable answer, which sometimes leads to wrong answers -- often referred to as ``hallucinations''~\cite{DBLP:journals/corr/abs-2302-04023,DBLP:conf/adbis/Hose23,DBLP:journals/corr/abs-2308-10168}. 
    \item Natural interfaces between machine and users: 
    A key to successful applications of the Semantic Web is intuitive user interfaces.
    Users must be offered applications that are intuitive and easy to use.
    This includes improving interfaces based on natural language for formulating queries and accessing structured data stored in SPARQL endpoints.
    Again, the use and deeper integration of LLMs with Knowledge Graphs shows a promising direction.
    \item Semantic Web components: There are still components of the architecture (see Section~\ref{sec:swarchitecture}) where active development and research are conducted.
    Most notably, there are crypto and trust. 
    Recent new W3C standards such as DID and Verifiable Credentials have been developed. 
    However, one can expect more work and development in this direction.
\end{itemize}

Finally, we like to point to existing literature discussing the future directions of research on the Semantic Web~\cite{DBLP:journals/corr/abs-2309-13939} and Knowledge Graphs~\cite{DBLP:journals/corr/abs-2308-14217}.
Breit et al. \cite{10.1145/3586163} conducted a survey on the fusion of Semantic Web and Machine Learning, exploring the opportunities arising from the convergence of these two paradigms.

\bibliography{references,references-grl}

@article{gofa,
	author       = {Lecheng Kong and
	Jiarui Feng and
	Hao Liu and
	Chengsong Huang and
	Jiaxin Huang and
	Yixin Chen and
	Muhan Zhang},
	title        = {{GOFA:} {A} Generative One-For-All Model for Joint Graph Language
	Modeling},
	journal      = {CoRR},
	volume       = {abs/2407.09709},
	year         = {2024},
	url          = {https://doi.org/10.48550/arXiv.2407.09709},
	doi          = {10.48550/ARXIV.2407.09709},
	eprinttype    = {arXiv},
	eprint       = {2407.09709},
	bibsource    = {dblp computer science bibliography, https://dblp.org}
}

@inproceedings{ofa,
author       = {Hao Liu and
Jiarui Feng and
Lecheng Kong and
Ningyue Liang and
Dacheng Tao and
Yixin Chen and
Muhan Zhang},
title        = {One For All: Towards Training One Graph Model For All Classification
Tasks},
booktitle    = {The Twelfth International Conference on Learning Representations,
{ICLR} 2024, Vienna, Austria, May 7-11, 2024},
publisher    = {OpenReview.net},
year         = {2024},
url          = {https://openreview.net/forum?id=4IT2pgc9v6},
bibsource    = {dblp computer science bibliography, https://dblp.org}
}

@inproceedings{opengraph,
author       = {Lianghao Xia and
Ben Kao and
Chao Huang},
editor       = {Yaser Al{-}Onaizan and
Mohit Bansal and
Yun{-}Nung Chen},
title        = {OpenGraph: Towards Open Graph Foundation Models},
booktitle    = {Findings of the Association for Computational Linguistics: {EMNLP}
2024, Miami, Florida, USA, November 12-16, 2024},
pages        = {2365--2379},
publisher    = {Association for Computational Linguistics},
year         = {2024},
url          = {https://aclanthology.org/2024.findings-emnlp.132},
bibsource    = {dblp computer science bibliography, https://dblp.org}
}

@article{graphany,
author       = {Jianan Zhao and
Hesham Mostafa and
Mikhail Galkin and
Michael M. Bronstein and
Zhaocheng Zhu and
Jian Tang},
title        = {GraphAny: {A} Foundation Model for Node Classification on Any Graph},
journal      = {CoRR},
volume       = {abs/2405.20445},
year         = {2024},
url          = {https://doi.org/10.48550/arXiv.2405.20445},
doi          = {10.48550/ARXIV.2405.20445},
eprinttype    = {arXiv},
eprint       = {2405.20445},
bibsource    = {dblp computer science bibliography, https://dblp.org}
}

@inproceedings{sgc,
author       = {Felix Wu and
Amauri H. Souza Jr. and
Tianyi Zhang and
Christopher Fifty and
Tao Yu and
Kilian Q. Weinberger},
editor       = {Kamalika Chaudhuri and
Ruslan Salakhutdinov},
title        = {Simplifying Graph Convolutional Networks},
booktitle    = {Proceedings of the 36th International Conference on Machine Learning,
{ICML} 2019, 9-15 June 2019, Long Beach, California, {USA}},
series       = {Proceedings of Machine Learning Research},
volume       = {97},
pages        = {6861--6871},
publisher    = {{PMLR}},
year         = {2019},
url          = {http://proceedings.mlr.press/v97/wu19e.html},
bibsource    = {dblp computer science bibliography, https://dblp.org}
}

@inproceedings{psp,
	author       = {Qingqing Ge and
	Zeyuan Zhao and
	Yiding Liu and
	Anfeng Cheng and
	Xiang Li and
	Shuaiqiang Wang and
	Dawei Yin},
	editor       = {Albert Bifet and
	Jesse Davis and
	Tomas Krilavicius and
	Meelis Kull and
	Eirini Ntoutsi and
	Indre Zliobaite},
	title        = {{PSP:} Pre-training and Structure Prompt Tuning for Graph Neural Networks},
	booktitle    = {Machine Learning and Knowledge Discovery in Databases. Research Track
	- European Conference, {ECML} {PKDD} 2024, Vilnius, Lithuania, September
	9-13, 2024, Proceedings, Part {V}},
	series       = {Lecture Notes in Computer Science},
	volume       = {14945},
	pages        = {423--439},
	publisher    = {Springer},
	year         = {2024},
	url          = {https://doi.org/10.1007/978-3-031-70362-1\_25},
	doi          = {10.1007/978-3-031-70362-1\_25},
	bibsource    = {dblp computer science bibliography, https://dblp.org}
}

@inproceedings{prodigy,
	author       = {Qian Huang and
	Hongyu Ren and
	Peng Chen and
	Gregor Krzmanc and
	Daniel Zeng and
	Percy Liang and
	Jure Leskovec},
	editor       = {Alice Oh and
	Tristan Naumann and
	Amir Globerson and
	Kate Saenko and
	Moritz Hardt and
	Sergey Levine},
	title        = {{PRODIGY:} Enabling In-context Learning Over Graphs},
	booktitle    = {Advances in Neural Information Processing Systems 36: Annual Conference
	on Neural Information Processing Systems 2023, NeurIPS 2023, New Orleans,
	LA, USA, December 10 - 16, 2023},
	year         = {2023},
	url          = {http://papers.nips.cc/paper\_files/paper/2023/hash/34dce0dc3121951dd0399ba02c0f0d06-Abstract-Conference.html},
	bibsource    = {dblp computer science bibliography, https://dblp.org}
}

@inproceedings{glm,
	author       = {Moritz Plenz and
	Anette Frank},
	editor       = {Lun{-}Wei Ku and
	Andre Martins and
	Vivek Srikumar},
	title        = {Graph Language Models},
	booktitle    = {Proceedings of the 62nd Annual Meeting of the Association for Computational
	Linguistics (Volume 1: Long Papers), {ACL} 2024, Bangkok, Thailand,
	August 11-16, 2024},
	pages        = {4477--4494},
	publisher    = {Association for Computational Linguistics},
	year         = {2024},
	url          = {https://doi.org/10.18653/v1/2024.acl-long.245},
	doi          = {10.18653/V1/2024.ACL-LONG.245},
	bibsource    = {dblp computer science bibliography, https://dblp.org}
}

@article{gpt4,
	author       = {OpenAI},
	title        = {{GPT-4} Technical Report},
	journal      = {CoRR},
	volume       = {abs/2303.08774},
	year         = {2023},
	url          = {https://doi.org/10.48550/arXiv.2303.08774},
	doi          = {10.48550/ARXIV.2303.08774},
	eprinttype    = {arXiv},
	eprint       = {2303.08774},
	bibsource    = {dblp computer science bibliography, https://dblp.org}
}

@article{t5,
	author       = {Colin Raffel and
	Noam Shazeer and
	Adam Roberts and
	Katherine Lee and
	Sharan Narang and
	Michael Matena and
	Yanqi Zhou and
	Wei Li and
	Peter J. Liu},
	title        = {Exploring the Limits of Transfer Learning with a Unified Text-to-Text
	Transformer},
	journal      = {J. Mach. Learn. Res.},
	volume       = {21},
	pages        = {140:1--140:67},
	year         = {2020},
	url          = {https://jmlr.org/papers/v21/20-074.html},
	bibsource    = {dblp computer science bibliography, https://dblp.org}
}

@article{llama3,
	author       = {Abhimanyu Dubey and
	Abhinav Jauhri and
	Abhinav Pandey and
	Abhishek Kadian and
	Ahmad Al{-}Dahle and
	{others} },
	title        = {The Llama 3 Herd of Models},
	journal      = {CoRR},
	volume       = {abs/2407.21783},
	year         = {2024},
	url          = {https://doi.org/10.48550/arXiv.2407.21783},
	doi          = {10.48550/ARXIV.2407.21783},
	eprinttype    = {arXiv},
	eprint       = {2407.21783},
	bibsource    = {dblp computer science bibliography, https://dblp.org}
}

@inproceedings{bert,
	author       = {Jacob Devlin and
	Ming{-}Wei Chang and
	Kenton Lee and
	Kristina Toutanova},
	editor       = {Jill Burstein and
	Christy Doran and
	Thamar Solorio},
	title        = {{BERT:} Pre-training of Deep Bidirectional Transformers for Language
	Understanding},
	booktitle    = {Proceedings of the 2019 Conference of the North American Chapter of
	the Association for Computational Linguistics: Human Language Technologies,
	{NAACL-HLT} 2019, Minneapolis, MN, USA, June 2-7, 2019, Volume 1 (Long
	and Short Papers)},
	pages        = {4171--4186},
	publisher    = {Association for Computational Linguistics},
	year         = {2019},
	url          = {https://doi.org/10.18653/v1/n19-1423},
	doi          = {10.18653/V1/N19-1423},
	bibsource    = {dblp computer science bibliography, https://dblp.org}
}

@inproceedings{DBLP:conf/nips/VaswaniSPUJGKP17,
	author       = {Ashish Vaswani and
	Noam Shazeer and
	Niki Parmar and
	Jakob Uszkoreit and
	Llion Jones and
	Aidan N. Gomez and
	Lukasz Kaiser and
	Illia Polosukhin},
	editor       = {Isabelle Guyon and
	Ulrike von Luxburg and
	Samy Bengio and
	Hanna M. Wallach and
	Rob Fergus and
	S. V. N. Vishwanathan and
	Roman Garnett},
	title        = {Attention is All you Need},
	booktitle    = {Advances in Neural Information Processing Systems 30: Annual Conference
	on Neural Information Processing Systems 2017, December 4-9, 2017,
	Long Beach, CA, {USA}},
	pages        = {5998--6008},
	year         = {2017},
	url          = {https://proceedings.neurips.cc/paper/2017/hash/3f5ee243547dee91fbd053c1c4a845aa-Abstract.html},
	bibsource    = {dblp computer science bibliography, https://dblp.org}
}

@inproceedings{r-gcn,
	author       = {Michael Sejr Schlichtkrull and
	Thomas N. Kipf and
	Peter Bloem and
	Rianne van den Berg and
	Ivan Titov and
	Max Welling},
	editor       = {Aldo Gangemi and
	Roberto Navigli and
	Maria{-}Esther Vidal and
	Pascal Hitzler and
	Rapha{\"{e}}l Troncy and
	Laura Hollink and
	Anna Tordai and
	Mehwish Alam},
	title        = {Modeling Relational Data with Graph Convolutional Networks},
	booktitle    = {The Semantic Web - 15th International Conference, {ESWC} 2018, Heraklion,
	Crete, Greece, June 3-7, 2018, Proceedings},
	series       = {Lecture Notes in Computer Science},
	volume       = {10843},
	pages        = {593--607},
	publisher    = {Springer},
	year         = {2018},
	url          = {https://doi.org/10.1007/978-3-319-93417-4\_38},
	doi          = {10.1007/978-3-319-93417-4\_38},
	bibsource    = {dblp computer science bibliography, https://dblp.org}
}

@article{graph-mlp,
	author       = {Yang Hu and
	Haoxuan You and
	Zhecan Wang and
	Zhicheng Wang and
	Erjin Zhou and
	Yue Gao},
	title        = {Graph-MLP: Node Classification without Message Passing in Graph},
	journal      = {CoRR},
	volume       = {abs/2106.04051},
	year         = {2021},
	url          = {https://arxiv.org/abs/2106.04051},
	eprinttype    = {arXiv},
	eprint       = {2106.04051},
	bibsource    = {dblp computer science bibliography, https://dblp.org}
}

@inproceedings{gcnii,
	author       = {Ming Chen and
	Zhewei Wei and
	Zengfeng Huang and
	Bolin Ding and
	Yaliang Li},
	title        = {Simple and Deep Graph Convolutional Networks},
	booktitle    = {Proceedings of the 37th International Conference on Machine Learning,
	{ICML} 2020, 13-18 July 2020, Virtual Event},
	series       = {Proceedings of Machine Learning Research},
	volume       = {119},
	pages        = {1725--1735},
	publisher    = {{PMLR}},
	year         = {2020},
	url          = {http://proceedings.mlr.press/v119/chen20v.html},
	bibsource    = {dblp computer science bibliography, https://dblp.org}
}

@article{pitfalls,
	author       = {Oleksandr Shchur and
	Maximilian Mumme and
	Aleksandar Bojchevski and
	Stephan G{\"{u}}nnemann},
	title        = {Pitfalls of Graph Neural Network Evaluation},
	journal      = {CoRR},
	volume       = {abs/1811.05868},
	year         = {2018},
	url          = {http://arxiv.org/abs/1811.05868},
	eprinttype    = {arXiv},
	eprint       = {1811.05868},
	bibsource    = {dblp computer science bibliography, https://dblp.org}
}

@inproceedings{gat,
	author       = {Petar Velickovic and
	Guillem Cucurull and
	Arantxa Casanova and
	Adriana Romero and
	Pietro Li{\`{o}} and
	Yoshua Bengio},
	title        = {Graph Attention Networks},
	booktitle    = {6th International Conference on Learning Representations, {ICLR} 2018,
	Vancouver, BC, Canada, April 30 - May 3, 2018, Conference Track Proceedings},
	publisher    = {OpenReview.net},
	year         = {2018},
	url          = {https://openreview.net/forum?id=rJXMpikCZ},
	bibsource    = {dblp computer science bibliography, https://dblp.org}
}

@INPROCEEDINGS{gorietal2005,
author={Gori, M. and Monfardini, G. and Scarselli, F.},
booktitle={Proceedings. 2005 IEEE International Joint Conference on Neural Networks, 2005.}, 
title={A new model for learning in graph domains}, 
year={2005},
volume={2},
number={},
pages={729-734 vol. 2},
doi={10.1109/IJCNN.2005.1555942}}

@article{scarsellietal2009,
author       = {Franco Scarselli and
Marco Gori and
Ah Chung Tsoi and
Markus Hagenbuchner and
Gabriele Monfardini},
title        = {The Graph Neural Network Model},
journal      = {{IEEE} Trans. Neural Networks},
volume       = {20},
number       = {1},
pages        = {61--80},
year         = {2009},
url          = {https://doi.org/10.1109/TNN.2008.2005605},
doi          = {10.1109/TNN.2008.2005605},
bibsource    = {dblp computer science bibliography, https://dblp.org}
}

@article{owl2vec,
	author       = {Jiaoyan Chen and
	Pan Hu and
	Ernesto Jim{\'{e}}nez{-}Ruiz and
	Ole Magnus Holter and
	Denvar Antonyrajah and
	Ian Horrocks},
	title        = {OWL2Vec*: embedding of {OWL} ontologies},
	journal      = {Mach. Learn.},
	volume       = {110},
	number       = {7},
	pages        = {1813--1845},
	year         = {2021},
	url          = {https://doi.org/10.1007/s10994-021-05997-6},
	doi          = {10.1007/S10994-021-05997-6},
	bibsource    = {dblp computer science bibliography, https://dblp.org}
}

@inproceedings{transowl,
	author       = {Claudia d'Amato and
	Nicola Flavio Quatraro and
	Nicola Fanizzi},
	editor       = {Ruben Verborgh and
	Katja Hose and
	Heiko Paulheim and
	Pierre{-}Antoine Champin and
	Maria Maleshkova and
	{\'{O}}scar Corcho and
	Petar Ristoski and
	Mehwish Alam},
	title        = {Injecting Background Knowledge into Embedding Models for Predictive
	Tasks on Knowledge Graphs},
	booktitle    = {The Semantic Web - 18th International Conference, {ESWC} 2021, Virtual
	Event, June 6-10, 2021, Proceedings},
	series       = {Lecture Notes in Computer Science},
	volume       = {12731},
	pages        = {441--457},
	publisher    = {Springer},
	year         = {2021},
	url          = {https://doi.org/10.1007/978-3-030-77385-4\_26},
	doi          = {10.1007/978-3-030-77385-4\_26},
	bibsource    = {dblp computer science bibliography, https://dblp.org}
}

@inproceedings{rotate,
	author       = {Zhiqing Sun and
	Zhi{-}Hong Deng and
	Jian{-}Yun Nie and
	Jian Tang},
	title        = {RotatE: Knowledge Graph Embedding by Relational Rotation in Complex
	Space},
	booktitle    = {7th International Conference on Learning Representations, {ICLR} 2019,
	New Orleans, LA, USA, May 6-9, 2019},
	publisher    = {OpenReview.net},
	year         = {2019},
	url          = {https://openreview.net/forum?id=HkgEQnRqYQ},
	bibsource    = {dblp computer science bibliography, https://dblp.org}
}

@inproceedings{transg,
	author       = {Han Xiao and
	Minlie Huang and
	Xiaoyan Zhu},
	title        = {TransG : {A} Generative Model for Knowledge Graph Embedding},
	booktitle    = {Proceedings of the 54th Annual Meeting of the Association for Computational
	Linguistics, {ACL} 2016, August 7-12, 2016, Berlin, Germany, Volume
	1: Long Papers},
	publisher    = {The Association for Computer Linguistics},
	year         = {2016},
	url          = {https://doi.org/10.18653/v1/p16-1219},
	doi          = {10.18653/V1/P16-1219},
	bibsource    = {dblp computer science bibliography, https://dblp.org}
}

@inproceedings{transr,
	author       = {Yankai Lin and
	Zhiyuan Liu and
	Maosong Sun and
	Yang Liu and
	Xuan Zhu},
	editor       = {Blai Bonet and
	Sven Koenig},
	title        = {Learning Entity and Relation Embeddings for Knowledge Graph Completion},
	booktitle    = {Proceedings of the Twenty-Ninth {AAAI} Conference on Artificial Intelligence,
	January 25-30, 2015, Austin, Texas, {USA}},
	pages        = {2181--2187},
	publisher    = {{AAAI} Press},
	year         = {2015},
	url          = {https://doi.org/10.1609/aaai.v29i1.9491},
	doi          = {10.1609/AAAI.V29I1.9491},
	bibsource    = {dblp computer science bibliography, https://dblp.org}
}

@inproceedings{transe,
	author       = {Antoine Bordes and
	Nicolas Usunier and
	Alberto Garc{\'{\i}}a{-}Dur{\'{a}}n and
	Jason Weston and
	Oksana Yakhnenko},
	editor       = {Christopher J. C. Burges and
	L{\'{e}}on Bottou and
	Zoubin Ghahramani and
	Kilian Q. Weinberger},
	title        = {Translating Embeddings for Modeling Multi-relational Data},
	booktitle    = {Advances in Neural Information Processing Systems 26: 27th Annual
	Conference on Neural Information Processing Systems 2013. Proceedings
	of a meeting held December 5-8, 2013, Lake Tahoe, Nevada, United States},
	pages        = {2787--2795},
	year         = {2013},
	url          = {https://proceedings.neurips.cc/paper/2013/hash/1cecc7a77928ca8133fa24680a88d2f9-Abstract.html},
	bibsource    = {dblp computer science bibliography, https://dblp.org}
}

@inproceedings{rdf2vec,
	author       = {Petar Ristoski and
	Heiko Paulheim},
	editor       = {Paul Groth and
	Elena Simperl and
	Alasdair J. G. Gray and
	Marta Sabou and
	Markus Kr{\"{o}}tzsch and
	Freddy L{\'{e}}cu{\'{e}} and
	Fabian Fl{\"{o}}ck and
	Yolanda Gil},
	title        = {RDF2Vec: {RDF} Graph Embeddings for Data Mining},
	booktitle    = {The Semantic Web - {ISWC} 2016 - 15th International Semantic Web Conference,
	Kobe, Japan, October 17-21, 2016, Proceedings, Part {I}},
	series       = {Lecture Notes in Computer Science},
	volume       = {9981},
	pages        = {498--514},
	year         = {2016},
	url          = {https://doi.org/10.1007/978-3-319-46523-4\_30},
	doi          = {10.1007/978-3-319-46523-4\_30},
	bibsource    = {dblp computer science bibliography, https://dblp.org}
}

@inproceedings{word2vec,
	author       = {Tom{\'{a}}s Mikolov and
	Ilya Sutskever and
	Kai Chen and
	Gregory S. Corrado and
	Jeffrey Dean},
	editor       = {Christopher J. C. Burges and
	L{\'{e}}on Bottou and
	Zoubin Ghahramani and
	Kilian Q. Weinberger},
	title        = {Distributed Representations of Words and Phrases and their Compositionality},
	booktitle    = {Advances in Neural Information Processing Systems 26: 27th Annual
	Conference on Neural Information Processing Systems 2013. Proceedings
	of a meeting held December 5-8, 2013, Lake Tahoe, Nevada, United States},
	pages        = {3111--3119},
	year         = {2013},
	url          = {https://proceedings.neurips.cc/paper/2013/hash/9aa42b31882ec039965f3c4923ce901b-Abstract.html},
	bibsource    = {dblp computer science bibliography, https://dblp.org}
}

@inproceedings{deepwalk,
	author       = {Bryan Perozzi and
	Rami Al{-}Rfou and
	Steven Skiena},
	editor       = {Sofus A. Macskassy and
	Claudia Perlich and
	Jure Leskovec and
	Wei Wang and
	Rayid Ghani},
	title        = {DeepWalk: online learning of social representations},
	booktitle    = {The 20th {ACM} {SIGKDD} International Conference on Knowledge Discovery
	and Data Mining, {KDD} '14, New York, NY, {USA} - August 24 - 27,
	2014},
	pages        = {701--710},
	publisher    = {{ACM}},
	year         = {2014},
	url          = {https://doi.org/10.1145/2623330.2623732},
	doi          = {10.1145/2623330.2623732},
	bibsource    = {dblp computer science bibliography, https://dblp.org}
}

@inproceedings{node2vec,
	author       = {Aditya Grover and
	Jure Leskovec},
	editor       = {Balaji Krishnapuram and
	Mohak Shah and
	Alexander J. Smola and
	Charu C. Aggarwal and
	Dou Shen and
	Rajeev Rastogi},
	title        = {node2vec: Scalable Feature Learning for Networks},
	booktitle    = {Proceedings of the 22nd {ACM} {SIGKDD} International Conference on
	Knowledge Discovery and Data Mining, San Francisco, CA, USA, August
	13-17, 2016},
	pages        = {855--864},
	publisher    = {{ACM}},
	year         = {2016},
	url          = {https://doi.org/10.1145/2939672.2939754},
	doi          = {10.1145/2939672.2939754},
	bibsource    = {dblp computer science bibliography, https://dblp.org}
}

@article{DBLP:journals/corr/abs-2404-14928,
	author       = {Wenqi Fan and
	Shijie Wang and
	Jiani Huang and
	Zhikai Chen and
	Yu Song and
	Wenzhuo Tang and
	Haitao Mao and
	Hui Liu and
	Xiaorui Liu and
	Dawei Yin and
	Qing Li},
	title        = {Graph Machine Learning in the Era of Large Language Models (LLMs)},
	journal      = {CoRR},
	volume       = {abs/2404.14928},
	year         = {2024},
	url          = {https://doi.org/10.48550/arXiv.2404.14928},
	doi          = {10.48550/ARXIV.2404.14928},
	eprinttype    = {arXiv},
	eprint       = {2404.14928},
	bibsource    = {dblp computer science bibliography, https://dblp.org}
}

@inproceedings{kgat,
	author       = {Xiang Wang and
	Xiangnan He and
	Yixin Cao and
	Meng Liu and
	Tat{-}Seng Chua},
	editor       = {Ankur Teredesai and
	Vipin Kumar and
	Ying Li and
	R{\'{o}}mer Rosales and
	Evimaria Terzi and
	George Karypis},
	title        = {{KGAT:} Knowledge Graph Attention Network for Recommendation},
	booktitle    = {Proceedings of the 25th {ACM} {SIGKDD} International Conference on
	Knowledge Discovery {\&} Data Mining, {KDD} 2019, Anchorage, AK,
	USA, August 4-8, 2019},
	pages        = {950--958},
	publisher    = {{ACM}},
	year         = {2019},
	url          = {https://doi.org/10.1145/3292500.3330989},
	doi          = {10.1145/3292500.3330989},
	bibsource    = {dblp computer science bibliography, https://dblp.org}
}

@article{DBLP:journals/corr/abs-2310-11829-graphfoundationmodels-survey,
	author       = {Jiawei Liu and
	Cheng Yang and
	Zhiyuan Lu and
	Junze Chen and
	Yibo Li and
	Mengmei Zhang and
	Ting Bai and
	Yuan Fang and
	Lichao Sun and
	Philip S. Yu and
	Chuan Shi},
	title        = {Towards Graph Foundation Models: {A} Survey and Beyond},
	journal      = {CoRR},
	volume       = {abs/2310.11829},
	year         = {2023},
	url          = {https://doi.org/10.48550/arXiv.2310.11829},
	doi          = {10.48550/ARXIV.2310.11829},
	eprinttype    = {arXiv},
	eprint       = {2310.11829},
	bibsource    = {dblp computer science bibliography, https://dblp.org}
}

@article{DBLP:journals/corr/abs-2406-14683-taglas,
	author       = {Jiarui Feng and
	Hao Liu and
	Lecheng Kong and
	Yixin Chen and
	Muhan Zhang},
	title        = {{TAGLAS:} An atlas of text-attributed graph datasets in the era of
	large graph and language models},
	journal      = {CoRR},
	volume       = {abs/2406.14683},
	year         = {2024},
	url          = {https://doi.org/10.48550/arXiv.2406.14683},
	doi          = {10.48550/ARXIV.2406.14683},
	eprinttype    = {arXiv},
	eprint       = {2406.14683},
	bibsource    = {dblp computer science bibliography, https://dblp.org}
}

@article{DBLP:journals/ijautcomp/CaoLHS24,
	author       = {Boxi Cao and
	Hongyu Lin and
	Xianpei Han and
	Le Sun},
	title        = {The Life Cycle of Knowledge in Big Language Models: {A} Survey},
	journal      = {Mach. Intell. Res.},
	volume       = {21},
	number       = {2},
	pages        = {217--238},
	year         = {2024},
	url          = {https://doi.org/10.1007/s11633-023-1416-x},
	doi          = {10.1007/S11633-023-1416-X},
	bibsource    = {dblp computer science bibliography, https://dblp.org}
}

@inproceedings{DBLP:conf/acl/GalkeS22,
	author       = {Lukas Galke and
	Ansgar Scherp},
	editor       = {Smaranda Muresan and
	Preslav Nakov and
	Aline Villavicencio},
	title        = {Bag-of-Words vs. Graph vs. Sequence in Text Classification: Questioning
	the Necessity of Text-Graphs and the Surprising Strength of a Wide
	{MLP}},
	booktitle    = {Proceedings of the 60th Annual Meeting of the Association for Computational
	Linguistics (Volume 1: Long Papers), {ACL} 2022, Dublin, Ireland,
	May 22-27, 2022},
	pages        = {4038--4051},
	publisher    = {Association for Computational Linguistics},
	year         = {2022},
	url          = {https://doi.org/10.18653/v1/2022.acl-long.279},
	doi          = {10.18653/V1/2022.ACL-LONG.279},
	bibsource    = {dblp computer science bibliography, https://dblp.org}
}

@article{DBLP:journals/dpd/GottronKS15,
	author       = {Thomas Gottron and
	Malte Knauf and
	Ansgar Scherp},
	title        = {Analysis of schema structures in the Linked Open Data graph based
	on unique subject URIs, pay-level domains, and vocabulary usage},
	journal      = {Distributed Parallel Databases},
	volume       = {33},
	number       = {4},
	pages        = {515--553},
	year         = {2015},
	url          = {https://doi.org/10.1007/s10619-014-7143-0},
	doi          = {10.1007/S10619-014-7143-0},
	bibsource    = {dblp computer science bibliography, https://dblp.org}
}

@article{DBLP:journals/corr/abs-2406-08993-log2024,
	author       = {Yuankai Luo and
	Lei Shi and
	Xiao{-}Ming Wu},
	title        = {Classic GNNs are Strong Baselines: Reassessing GNNs for Node Classification},
	journal      = {CoRR},
	volume       = {abs/2406.08993},
	year         = {2024},
	url          = {https://doi.org/10.48550/arXiv.2406.08993},
	doi          = {10.48550/ARXIV.2406.08993},
	eprinttype    = {arXiv},
	eprint       = {2406.08993},
	bibsource    = {dblp computer science bibliography, https://dblp.org}
}

@article{DBLP:journals/jmlr/IzacardLLHPSDJRG23-atlas,
	author       = {Gautier Izacard and
	Patrick S. H. Lewis and
	Maria Lomeli and
	Lucas Hosseini and
	Fabio Petroni and
	Timo Schick and
	Jane Dwivedi{-}Yu and
	Armand Joulin and
	Sebastian Riedel and
	Edouard Grave},
	title        = {Atlas: Few-shot Learning with Retrieval Augmented Language Models},
	journal      = {J. Mach. Learn. Res.},
	volume       = {24},
	pages        = {251:1--251:43},
	year         = {2023},
	url          = {https://jmlr.org/papers/v24/23-0037.html},
	bibsource    = {dblp computer science bibliography, https://dblp.org}
}

@inproceedings{DBLP:conf/naacl/MoiseevDAJ22-skill,
	author       = {Fedor Moiseev and
	Zhe Dong and
	Enrique Alfonseca and
	Martin Jaggi},
	editor       = {Marine Carpuat and
	Marie{-}Catherine de Marneffe and
	Iv{\'{a}}n Vladimir Meza Ru{\'{\i}}z},
	title        = {{SKILL:} Structured Knowledge Infusion for Large Language Models},
	booktitle    = {Proceedings of the 2022 Conference of the North American Chapter of
	the Association for Computational Linguistics: Human Language Technologies,
	{NAACL} 2022, Seattle, WA, United States, July 10-15, 2022},
	pages        = {1581--1588},
	publisher    = {Association for Computational Linguistics},
	year         = {2022},
	url          = {https://doi.org/10.18653/v1/2022.naacl-main.113},
	doi          = {10.18653/V1/2022.NAACL-MAIN.113},
	bibsource    = {dblp computer science bibliography, https://dblp.org}
}

@article{graph-foundation-models-survey,
	author       = {Jiawei Liu and
	Cheng Yang and
	Zhiyuan Lu and
	Junze Chen and
	Yibo Li and
	Mengmei Zhang and
	Ting Bai and
	Yuan Fang and
	Lichao Sun and
	Philip S. Yu and
	Chuan Shi},
	title        = {Towards Graph Foundation Models: {A} Survey and Beyond},
	journal      = {CoRR},
	volume       = {abs/2310.11829},
	year         = {2023},
	url          = {https://doi.org/10.48550/arXiv.2310.11829},
	doi          = {10.48550/ARXIV.2310.11829},
	eprinttype    = {arXiv},
	eprint       = {2310.11829},
	bibsource    = {dblp computer science bibliography, https://dblp.org}
}

@inproceedings{graph-transformer-networks,
	author       = {Seongjun Yun and
	Minbyul Jeong and
	Raehyun Kim and
	Jaewoo Kang and
	Hyunwoo J. Kim},
	editor       = {Hanna M. Wallach and
	Hugo Larochelle and
	Alina Beygelzimer and
	Florence d'Alch{\'{e}}{-}Buc and
	Emily B. Fox and
	Roman Garnett},
	title        = {Graph Transformer Networks},
	booktitle    = {Advances in Neural Information Processing Systems 32: Annual Conference
	on Neural Information Processing Systems 2019, NeurIPS 2019, December
	8-14, 2019, Vancouver, BC, Canada},
	pages        = {11960--11970},
	year         = {2019},
	url          = {https://proceedings.neurips.cc/paper/2019/hash/9d63484abb477c97640154d40595a3bb-Abstract.html},
	bibsource    = {dblp computer science bibliography, https://dblp.org}
}

@article{foundation-models,
	author       = {Rishi Bommasani and
	Drew A. Hudson and
	Ehsan Adeli and
	Russ B. Altman and
	Simran Arora and
	{et al.}},
	title        = {On the Opportunities and Risks of Foundation Models},
	journal      = {CoRR},
	volume       = {abs/2108.07258},
	year         = {2021},
	url          = {https://arxiv.org/abs/2108.07258},
	eprinttype    = {arXiv},
	eprint       = {2108.07258},
	bibsource    = {dblp computer science bibliography, https://dblp.org}
}

@article{are-we-makin-progress,
	author       = 
	{Lukas Galke and
	Ansgar Scherp and 
	Andor Diera and
	Bao Xin Lin and
	Bhakti Khera and
	Tim Meuser and
	Tushar Singhal and
	Fabian Karl},
	title        = {A Review of Methods for Single-label and
	Multi-label Text Classification},
	journal      = {CoRR},
	volume       = {abs/2204.03954},
	year         = {2022},
	url          = {https://doi.org/10.48550/arXiv.2204.03954},
	doi          = {10.48550/ARXIV.2204.03954},
	eprinttype    = {arXiv},
	eprint       = {2204.03954},
	bibsource    = {dblp computer science bibliography, https://dblp.org}
}

@inproceedings{DBLP:conf/iclr/0001LST22-homophily-necessary,
	author       = {Yao Ma and
	Xiaorui Liu and
	Neil Shah and
	Jiliang Tang},
	title        = {Is Homophily a Necessity for Graph Neural Networks?},
	booktitle    = {The Tenth International Conference on Learning Representations, {ICLR}
	2022, Virtual Event, April 25-29, 2022},
	publisher    = {OpenReview.net},
	year         = {2022},
	url          = {https://openreview.net/forum?id=ucASPPD9GKN},
	bibsource    = {dblp computer science bibliography, https://dblp.org}
}

@article{DBLP:journals/access/YeKSSW22-kg-survey,
	author       = {Zi Ye and
	Yogan Jaya Kumar and
	Goh Ong Sing and
	Fengyan Song and
	Junsong Wang},
	title        = {A Comprehensive Survey of Graph Neural Networks for Knowledge Graphs},
	journal      = {{IEEE} Access},
	volume       = {10},
	pages        = {75729--75741},
	year         = {2022},
	url          = {https://doi.org/10.1109/ACCESS.2022.3191784},
	doi          = {10.1109/ACCESS.2022.3191784},
	bibsource    = {dblp computer science bibliography, https://dblp.org}
}

@article{DBLP:journals/nn/GalkeVFZHS23,
	author       = {Lukas Galke and
	Iacopo Vagliano and
	Benedikt Franke and
	Tobias Zielke and
	Marcel Hoffmann and
	Ansgar Scherp},
	title        = {Lifelong learning on evolving graphs under the constraints of imbalanced
	classes and new classes},
	journal      = {Neural Networks},
	volume       = {164},
	pages        = {156--176},
	year         = {2023},
	url          = {https://doi.org/10.1016/j.neunet.2023.04.022},
	doi          = {10.1016/J.NEUNET.2023.04.022},
	bibsource    = {dblp computer science bibliography, https://dblp.org}
}

@Article{biswas_et_al:TGDK.1.1.4,
	author =	{Biswas, Russa and Kaffee, Lucie-Aim\'{e}e and Cochez, Michael and Dumbrava, Stefania and Jendal, Theis E. and Lissandrini, Matteo and Lopez, Vanessa and Menc{\'\i}a, Eneldo Loza and Paulheim, Heiko and Sack, Harald and Vakaj, Edlira Kalemi and de Melo, Gerard},
	title =	{{Knowledge Graph Embeddings: Open Challenges and Opportunities}},
	journal =	{Transactions on Graph Data and Knowledge},
	pages =	{4:1--4:32},
	year =	{2023},
	volume =	{1},
	number =	{1},
	publisher =	{Schloss Dagstuhl -- Leibniz-Zentrum f{\"u}r Informatik},
	address =	{Dagstuhl, Germany},
	URL =		{https://drops.dagstuhl.de/entities/document/10.4230/TGDK.1.1.4},
	URN =		{urn:nbn:de:0030-drops-194783},
	doi =		{10.4230/TGDK.1.1.4}
}

@book{hamilton-book,
	author       = {William L. Hamilton},
	title        = {Graph Representation Learning},
	series       = {Synthesis Lectures on Artificial Intelligence and Machine Learning},
	publisher    = {Morgan {\&} Claypool Publishers},
	year         = {2020},
	url          = {https://doi.org/10.2200/S01045ED1V01Y202009AIM046},
	doi          = {10.2200/S01045ED1V01Y202009AIM046},
	isbn         = {978-3-031-00460-5},
	bibsource    = {dblp computer science bibliography, https://dblp.org}
}

@inproceedings{gcn,
	author       = {Thomas N. Kipf and
	Max Welling},
	title        = {Semi-Supervised Classification with Graph Convolutional Networks},
	booktitle    = {5th International Conference on Learning Representations, {ICLR} 2017,
	Toulon, France, April 24-26, 2017, Conference Track Proceedings},
	publisher    = {OpenReview.net},
	year         = {2017},
	url          = {https://openreview.net/forum?id=SJU4ayYgl},
	bibsource    = {dblp computer science bibliography, https://dblp.org}
}

@book{murphy-book,
	author = "Kevin P. Murphy",
	title = "Probabilistic Machine Learning: An introduction",
	publisher = "MIT Press",
	year = 2022,
	url = "http://probml.github.io/book1"
}

@article{DBLP:journals/tgdk/dAmatoMMS23,
  author       = {Claudia d'Amato and
                  Louis Mahon and
                  Pierre Monnin and
                  Giorgos Stamou},
  title        = {Machine Learning and Knowledge Graphs: Existing Gaps and Future Research
                  Challenges},
  journal      = {{TGDK}},
  volume       = {1},
  number       = {1},
  pages        = {8:1--8:35},
  year         = {2023},
  url          = {https://doi.org/10.4230/TGDK.1.1.8},
  doi          = {10.4230/TGDK.1.1.8},
  bibsource    = {dblp computer science bibliography, https://dblp.org}
}

@inproceedings{DBLP:conf/esws/Hartig11,
  author       = {Olaf Hartig},
  title        = "{Zero-Knowledge Query Planning for an Iterator Implementation of Link Traversal Based Query Execution}",
  booktitle    = {{ESWC} {(1)}},
  series       = {Lecture Notes in Computer Science},
  volume       = {6643},
  pages        = {154--169},
  publisher    = {Springer},
  year         = {2011}
}

@book{DBLP:series/synthesis/2017Gayo,
  author       = {Jos{\'{e}} Emilio Labra Gayo and
                  Eric Prud'hommeaux and
                  Iovka Boneva and
                  Dimitris Kontokostas},
  title        = {Validating {RDF} Data},
  series       = {Synthesis Lectures on the Semantic Web: Theory and Technology},
  publisher    = {Morgan {\&} Claypool Publishers},
  year         = {2017},
  url          = {https://doi.org/10.2200/S00786ED1V01Y201707WBE016},
  doi          = {10.2200/S00786ED1V01Y201707WBE016},
  isbn         = {978-3-031-79477-3}
}

@article{calvanese2017ontop,
  author       = {Diego Calvanese and
                  Benjamin Cogrel and
                  Sarah Komla{-}Ebri and
                  Roman Kontchakov and
                  Davide Lanti and
                  Martin Rezk and
                  Mariano Rodriguez{-}Muro and
                  Guohui Xiao},
  title        = {Ontop: Answering {SPARQL} queries over relational databases},
  journal      = {Semantic Web},
  volume       = {8},
  number       = {3},
  pages        = {471--487},
  year         = {2017},
  url          = {https://doi.org/10.3233/SW-160217},
  doi          = {10.3233/SW-160217}
}

@article{DBLP:journals/ws/SequedaM13,
  author       = {Juan F. Sequeda and
                  Daniel P. Miranker},
  title        = {Ultrawrap: {SPARQL} execution on relational data},
  journal      = {J. Web Semant.},
  volume       = {22},
  pages        = {19--39},
  year         = {2013}
}

@inproceedings{DBLP:conf/www/PriyatnaCS14,
  author       = {Freddy Priyatna and
                  {\'{O}}scar Corcho and
                  Juan F. Sequeda},
  title        = {Formalisation and experiences of R2RML-based {SPARQL} to {SQL} query
                  translation using morph},
  booktitle    = {23rd International World Wide Web Conference, {WWW} '14, Seoul, Republic
                  of Korea, April 7-11, 2014},
  pages        = {479--490},
  publisher    = {{ACM}},
  year         = {2014},
  url          = {https://doi.org/10.1145/2566486.2567981},
  doi          = {10.1145/2566486.2567981}
}

@article{chaves2020enhancing,
  author       = {David Chaves{-}Fraga and
                  Edna Ruckhaus and
                  Freddy Priyatna and
                  Maria{-}Esther Vidal and
                  {\'{O}}scar Corcho},
  title        = {{Enhancing virtual ontology based access over tabular data with Morph-CSV}},
  journal      = {Semantic Web},
  volume       = {12},
  number       = {6},
  pages        = {869--902},
  year         = {2021},
  url          = {https://doi.org/10.3233/SW-210432},
  doi          = {10.3233/SW-210432} 
}

@inproceedings{mami2019squerall,
  title={{Squerall: Virtual ontology-based access to heterogeneous and large data sources}},
  author={Mami, Mohamed Nadjib and Graux, Damien and Scerri, Simon and Jabeen, Hajira and Auer, S{\"o}ren and Lehmann, Jens},
  booktitle={International Semantic Web Conference},
  pages={229--245},
  year={2019},
  organization={Springer}
}

@inproceedings{oo2022rmlstreamer,
  title={RMLStreamer-SISO: an RDF stream generator from streaming heterogeneous data},
  author={Oo, Sitt Min and Haesendonck, Gerald and De Meester, Ben and Dimou, Anastasia},
  booktitle={International Semantic Web Conference},
  pages={697--713},
  year={2022},
  organization={Springer}
}

@article{10.1145/3586163,
author = {Breit, Anna and Waltersdorfer, Laura and Ekaputra, Fajar J. and Sabou, Marta and Ekelhart, Andreas and Iana, Andreea and Paulheim, Heiko and Portisch, Jan and Revenko, Artem and Teije, Annette Ten and Van Harmelen, Frank},
title = {Combining Machine Learning and Semantic Web: A Systematic Mapping Study},
year = {2023},
issue_date = {December 2023},
publisher = {Association for Computing Machinery},
address = {New York, NY, USA},
volume = {55},
number = {14s},
issn = {0360-0300},
url = {https://doi.org/10.1145/3586163},
doi = {10.1145/3586163},
journal = {ACM Comput. Surv.},
month = {jul},
articleno = {313},
numpages = {41}
}

@article{DBLP:journals/corr/abs-2306-08302,
  author       = {Shirui Pan and
                  Linhao Luo and
                  Yufei Wang and
                  Chen Chen and
                  Jiapu Wang and
                  Xindong Wu},
  title        = {Unifying Large Language Models and Knowledge Graphs: {A} Roadmap},
  journal      = {CoRR},
  volume       = {abs/2306.08302},
  year         = {2023},
  doi          = {10.48550/arXiv.2306.08302}
}

@incollection{DBLP:reference/sp/HorrocksP11,
  author       = {Ian Horrocks and
                  Peter F. Patel{-}Schneider},
  editor       = {John Domingue and
                  Dieter Fensel and
                  James A. Hendler},
  title        = {{KR} and Reasoning on the Semantic Web: {OWL}},
  booktitle    = {Handbook of Semantic Web Technologies},
  pages        = {365--398},
  publisher    = {Springer},
  year         = {2011},
  url          = {https://doi.org/10.1007/978-3-540-92913-0\_9},
  doi          = {10.1007/978-3-540-92913-0\_9},
  bibsource    = {dblp computer science bibliography, https://dblp.org}
}

@ARTICLE{Hou2023-fo,
  title    = "From answers to insights: Unveiling the strengths and limitations
              of {ChatGPT} and Biomedical Knowledge Graphs",
  author   = "Hou, Yu and Yeung, Jeremy and Xu, Hua and Su, Chang and Wang, Fei
              and Zhang, Rui",
  journal  = "medRxiv",
  month    =  jun,
  year     =  2023,
  language = "en"
}

@article{10.1145/3608966,
author = {Denning, Peter J.},
title = {The Smallness of Large Language Models},
year = {2023},
issue_date = {September 2023},
publisher = {Association for Computing Machinery},
address = {New York, NY, USA},
volume = {66},
number = {9},
issn = {0001-0782},
url = {https://doi.org/10.1145/3608966},
doi = {10.1145/3608966},
journal = {Commun. ACM},
month = {aug},
pages = {24–27},
numpages = {4}
}

@inproceedings{AbuAebAgl23,
  author       = {Ghadeer Abuoda and Christian Aebeloe and Daniele Dell'Aglio and Arthur Keen and Katja Hose},
  title        = "{StarBench: Benchmarking RDF-star Triplestores}",
  booktitle    = {QuWeDa icw. ISWC 2023},
  series       = {{CEUR} Workshop Proceedings},
  volume       = {3565},
  publisher    = {CEUR-WS.org},
  year         = {2023}
}

@book{AbonyiEtAl2024OntologiesInIndustry,
  editor = {János Abonyi and László Nagy and Tamás Ruppert},
  title = {Ontology-Based Development of Industry 4.0 and 5.0 Solutions for Smart Manufacturing and Production: Knowledge Graph and Semantic Based Modeling and Optimization of Complex Systems},
  publisher = {Springer},
  year = {2024}
  }

@inproceedings{DBLP:conf/ecai/BellomariniNS20,
  author       = {Luigi Bellomarini and
                  Markus Nissl and
                  Emanuel Sallinger},
  title        = {Blockchains as Knowledge Graphs - Blockchains for Knowledge Graphs
                  (Vision Paper)},
  booktitle    = {Proceedings of the International Workshop on Knowledge Representation
                  and Representation Learning co-located with the 24th European Conference
                  on Artificial Intelligence {(ECAI} 2020), Virtual Event, September,
                  2020},
  series       = {{CEUR} Workshop Proceedings},
  volume       = {3020},
  pages        = {43--51},
  publisher    = {CEUR-WS.org},
  year         = {2020}
}

@inproceedings{DBLP:conf/rr/KhandelwalJK11,
  author       = {Ankesh Khandelwal and
                  Ian Jacobi and
                  Lalana Kagal},
  editor       = {Sebastian Rudolph and
                  Claudio Gutierrez},
  title        = {Linked Rules: Principles for Rule Reuse on the Web},
  booktitle    = {Web Reasoning and Rule Systems - 5th International Conference, {RR}
                  2011, Galway, Ireland, August 29-30, 2011. Proceedings},
  series       = {Lecture Notes in Computer Science},
  volume       = {6902},
  pages        = {108--123},
  publisher    = {Springer},
  year         = {2011},
  doi          = {10.1007/978-3-642-23580-1\_9}
}

@article{DBLP:journals/corr/abs-2309-13939,
  author       = {Irene Celino and
                  Heiko Paulheim},
  title        = {The Time Traveler's Guide to Semantic Web Research: Analyzing Fictitious
                  Research Themes in the {ESWC} "Next 20 Years" Track},
  journal      = {CoRR},
  volume       = {abs/2309.13939},
  year         = {2023},
  doi          = {10.48550/arXiv.2309.13939},
}

@book{DBLP:books/sp/GriesS93,
  author       = {David Gries and
                  Fred B. Schneider},
  title        = {A Logical Approach to Discrete Math},
  series       = {Texts and Monographs in Computer Science},
  publisher    = {Springer},
  year         = {1993},
  url          = {https://doi.org/10.1007/978-1-4757-3837-7},
  doi          = {10.1007/978-1-4757-3837-7},
  isbn         = {0-387-94115-0}
}

@inproceedings{DBLP:conf/esws/XiaoHBRGC18,
  author       = {Guohui Xiao and
                  Dag Hovland and
                  Dimitris Bilidas and
                  Martin Rezk and
                  Martin Giese and
                  Diego Calvanese},
  title        = {Efficient Ontology-Based Data Integration with Canonical IRIs},
  booktitle    = {The Semantic Web - 15th International Conference, {ESWC} 2018, Heraklion,
                  Crete, Greece, June 3-7, 2018, Proceedings},
  series       = {Lecture Notes in Computer Science},
  volume       = {10843},
  pages        = {697--713},
  publisher    = {Springer},
  year         = {2018},
  doi          = {10.1007/978-3-319-93417-4\_45}
}

@article{DBLP:journals/semweb/Debattista0AC18,
  author       = {Jeremy Debattista and
                  Christoph Lange and
                  S{\"{o}}ren Auer and
                  Dominic Cortis},
  title        = {Evaluating the quality of the {LOD} cloud: An empirical investigation},
  journal      = {Semantic Web},
  volume       = {9},
  number       = {6},
  pages        = {859--901},
  year         = {2018},
  doi          = {10.3233/SW-180306}
}

@Article{Zaveri2015surveyQuality,
  author    = {Zaveri, Amrapali and Rula, Anisa and Maurino, Andrea and Pietrobon, Ricardo and Lehmann, Jens and Auer, S{\"o}ren},
  journal   = {Semantic Web Journal},
  title     = {{Quality assessment for linked data: A survey}},
  year      = {2015},
  month     = mar,
  number    = {1},
  pages     = {63--93},
  volume    = {7},
  comment   = {coins 'intrinsic dimension' for data quality},
  doi       = {10.3233/SW-150175},
  publisher = {IOS Press},
  url       = {http://www.semantic-web-journal.net/system/files/swj556.pdf},
}

@article{DBLP:journals/corr/abs-2308-14217,
  author       = {Xin Luna Dong},
  title        = {Generations of Knowledge Graphs: The Crazy Ideas and the Business Impact},
  journal      = {CoRR},
  volume       = {abs/2308.14217},
  year         = {2023},
  doi          = {10.48550/arXiv.2308.14217}
}

@misc{SHACL,
    author = {Knublauch, Holger and Kontokostas, Dimitris},
    title = {Shapes Constraint Language (SHACL)},
    howpublished = {W3C Recommendation},
    year = 2017,
    url = {https://www.w3.org/TR/2017/REC-shacl-20170720/}
}

@misc{SHEX,
    author = {Eric Prudhommeaux and Iovka Boneva and Jose Emilio Labra Gayo and Gregg Kellogg},
    title = {SHEX - SHAPE EXPRESSIONS},
    year = 2018,
    url = {https://shex.io/shex-semantics/index.html}
}

@InProceedings{Corman2018,
    author="Julien Corman
        and Juan L. Reutter
        and Ognjen Savkovi{\'c}",
    title="Semantics and {V}alidation of Recursive {SHACL}",
    booktitle="International Semantic Web Conference {ISWC}",
    year="2018",
    publisher="Springer",
    pages="318--336"
}

@InProceedings{Corman2019,
    author = "Julien Corman
        and Fernando Florenzano
        and Juan L. Reutter
        and Ognjen Savkovi{\'c}",
    title = "Validating {SHACL} {C}onstraints over a {SPARQL} {E}ndpoint",
    booktitle="International Semantic Web Conference {ISWC}",
    year = "2019",
    publisher="Springer",
    pages = "145--163"
}

@inproceedings{Andresel2020,
    author = {Andreşel, Medina and Corman, Julien and Ortiz, Magdalena and Reutter, Juan L. and Savković, Ognjen and Šimkus, Mantas},
    year = {2020},
    pages = {1570--1580},
    title = {Stable Model Semantics for Recursive SHACL},
    booktitle = {ACM--The Web Conference}
}

@inproceedings{DBLP:conf/esws/RohdeV23,
  author       = {Philipp D. Rohde and
                  Maria{-}Esther Vidal},
  title        = {Towards Certified Distributed Query Processing},
  booktitle    = {Joint Proceedings of the {ESWC} 2023 Workshops and Tutorials co-located
                  with 20th European Semantic Web Conference {(ESWC} 2023), Hersonissos,
                  Greece, May 28-29, 2023},
  series       = {{CEUR} Workshop Proceedings},
  volume       = {3443},
  publisher    = {CEUR-WS.org},
  year         = {2023}
}

@inproceedings{DBLP:conf/edbt/DelvaDJB23,
  author       = {Thomas Delva and
                  Anastasia Dimou and
                  Maxime Jakubowski and
                  Jan Van den Bussche},
  title        = {Data Provenance for {SHACL}},
  booktitle    = {Proceedings 26th International Conference on Extending Database Technology,
                  {EDBT} 2023, Ioannina, Greece, March 28-31, 2023},
  pages        = {285--297},
  publisher    = {OpenProceedings.org},
  year         = {2023},
  doi          = {10.48786/edbt.2023.23}
}

@inproceedings{DBLP:conf/vldb/Rohde21,
  author       = {Philipp D. Rohde},
  editor       = {Philip A. Bernstein and
                  Tilmann Rabl},
  title        = {{SHACL} Constraint Validation during {SPARQL} Query Processing},
  booktitle    = {Proceedings of the {VLDB} 2021 PhD Workshop co-located with the 47th
                  International Conference on Very Large Databases {(VLDB} 2021), Copenhagen,
                  Denmark, August 16, 2021},
  series       = {{CEUR} Workshop Proceedings},
  volume       = {2971},
  publisher    = {CEUR-WS.org},
  year         = {2021}
}

@article{Motik2009,
    author = {Motik, Boris and Horrocks, Ian and Sattler, Ulrike},
    title = {Bridging the Gap between OWL and Relational Databases},
    year = "2009",
    volume = {7},
    number = {2},
    journal = {Web Semantics: Science, Services and Agents on the World Wide Web},
    pages = {74-89},
    numpages = {16},
    publisher = {Elsevier Science Publishers B. V.},
    address = {NLD}
}

@InProceedings{Motik2007,
    author = {Motik, Boris and Horrocks, Ian and Sattler, Ulrike},
    title = {Adding Integrity Constraints to OWL},
    booktitle = "OWLED 2007 - OWL: Experiences and Directions",
    year = "2007",
    publisher = "CEUR Workshop Proceedings (CEUR-WS.org)",
    address = {Aachen},
    volume = "258"
}

@inproceedings{DBLP:conf/edbt/LausenMS08,
  author       = {Georg Lausen and
                  Michael Meier and
                  Michael Schmidt},
  editor       = {Alfons Kemper and
                  Patrick Valduriez and
                  Noureddine Mouaddib and
                  Jens Teubner and
                  Mokrane Bouzeghoub and
                  Volker Markl and
                  Laurent Amsaleg and
                  Ioana Manolescu},
  title        = {SPARQLing constraints for {RDF}},
  booktitle    = {{EDBT} 2008, 11th International Conference on Extending Database Technology,
                  Nantes, France, March 25-29, 2008, Proceedings},
  series       = {{ACM} International Conference Proceeding Series},
  volume       = {261},
  pages        = {499--509},
  publisher    = {{ACM}},
  year         = {2008},
  doi          = {10.1145/1353343.1353404}
}

@incollection{DBLP:books/sp/22/0002LB22,
  author       = {Christoph Lange and
                  J{\"{o}}rg Langkau and
                  Sebastian R. Bader},
  editor       = {Boris Otto and
                  Michael ten Hompel and
                  Stefan Wrobel},
  title        = {The {IDS} Information Model: {A} Semantic Vocabulary for Sovereign
                  Data Exchange},
  booktitle    = {Designing Data Spaces: The Ecosystem Approach to Competitive Advantage},
  pages        = {111--127},
  publisher    = {Springer},
  year         = {2022},
  doi          = {10.1007/978-3-030-93975-5\_7}
}

@book{DBLP:series/synthesis/2021Hogan,
  author       = {Aidan Hogan and
                  Eva Blomqvist and
                  Michael Cochez and
                  Claudia d'Amato and
                  Gerard de Melo and
                  Claudio Gutierrez and
                  Sabrina Kirrane and
                  Jos{\'{e}} Emilio Labra Gayo and
                  Roberto Navigli and
                  Sebastian Neumaier and
                  Axel{-}Cyrille Ngonga Ngomo and
                  Axel Polleres and
                  Sabbir M. Rashid and
                  Anisa Rula and
                  Lukas Schmelzeisen and
                  Juan Sequeda and
                  Steffen Staab and
                  Antoine Zimmermann},
  title        = {Knowledge Graphs},
  series       = {Synthesis Lectures on Data, Semantics, and Knowledge},
  publisher    = {Morgan {\&} Claypool Publishers},
  year         = {2021},
  doi          = {10.2200/S01125ED1V01Y202109DSK022}
}

@inproceedings{DBLP:conf/aaai/TaoSBM10,
  author       = {Jiao Tao and
                  Evren Sirin and
                  Jie Bao and
                  Deborah L. McGuinness},
  editor       = {Maria Fox and
                  David Poole},
  title        = {Integrity Constraints in {OWL}},
  booktitle    = {Proceedings of the Twenty-Fourth {AAAI} Conference on Artificial Intelligence,
                  {AAAI} 2010, Atlanta, Georgia, USA, July 11-15, 2010},
  pages        = {1443--1448},
  publisher    = {{AAAI} Press},
  year         = {2010},
  doi          = {10.1609/aaai.v24i1.7525}
}

@inproceedings{DBLP:conf/www/FigueraRV21,
  author       = {M{\'{o}}nica Figuera and
                  Philipp D. Rohde and
                  Maria{-}Esther Vidal},
  editor       = {Jure Leskovec and
                  Marko Grobelnik and
                  Marc Najork and
                  Jie Tang and
                  Leila Zia},
  title        = {Trav-SHACL: Efficiently Validating Networks of {SHACL} Constraints},
  booktitle    = {{WWW} '21: The Web Conference 2021, Virtual Event / Ljubljana, Slovenia,
                  April 19-23, 2021},
  pages        = {3337--3348},
  publisher    = {{ACM} / {IW3C2}},
  year         = {2021},
  doi          = {10.1145/3442381.3449877}
}

@inproceedings{DBLP:conf/semweb/CarrollBHS04,
  author       = {Jeremy J. Carroll and
                  Christian Bizer and
                  Patrick J. Hayes and
                  Patrick Stickler},
  editor       = {Jennifer Golbeck and
                  Piero A. Bonatti and
                  Wolfgang Nejdl and
                  Daniel Olmedilla and
                  Marianne Winslett},
  title        = {Semantic Web Publishing using Named Graphs},
  booktitle    = {Proceedings of the ISWC*04 Workshop on Trust, Security, and Reputation
                  on the Semantic Web, Hiroshima, Japan, November 7, 2004},
  series       = {{CEUR} Workshop Proceedings},
  volume       = {127},
  publisher    = {CEUR-WS.org},
  year         = {2004},
  url          = {https://ceur-ws.org/Vol-127/paper2.pdf},
  bibsource    = {dblp computer science bibliography, https://dblp.org}
}

@article{DBLP:journals/ao/Uschold15,
  author       = {Michael Uschold},
  title        = {Ontology and database schema: What's the difference?},
  journal      = {Appl. Ontology},
  volume       = {10},
  number       = {3-4},
  pages        = {243--258},
  year         = {2015},
  doi          = {10.3233/AO-150158}
}

@article{DBLP:journals/sigmod/UscholdG04,
  author       = {Michael Uschold and
                  Michael Gr{\"{u}}ninger},
  title        = {Ontologies and Semantics for Seamless Connectivity},
  journal      = {{SIGMOD} Rec.},
  volume       = {33},
  number       = {4},
  pages        = {58--64},
  year         = {2004},
  url          = {https://doi.org/10.1145/1041410.1041420},
  doi          = {10.1145/1041410.1041420}
}

@inproceedings{DBLP:conf/dagstuhl/McGuinness03,
  author       = {Deborah L. McGuinness},
  editor       = {Dieter Fensel and
                  James A. Hendler and
                  Henry Lieberman and
                  Wolfgang Wahlster},
  title        = {Ontologies Come of Age},
  booktitle    = {Spinning the Semantic Web: Bringing the World Wide Web to Its Full
                  Potential [outcome of a Dagstuhl seminar]},
  pages        = {171--194},
  publisher    = {{MIT} Press},
  year         = {2003}
}

@Inbook{Cheatham2017,
author="Cheatham, Michelle
and Pesquita, Catia",
editor="Zomaya, Albert Y.
and Sakr, Sherif",
title="Semantic Data Integration",
bookTitle="Handbook of Big Data Technologies",
year="2017",
publisher="Springer International Publishing",
address="Cham",
pages="263--305",
isbn="978-3-319-49340-4",
doi="10.1007/978-3-319-49340-4_8",
url="https://doi.org/10.1007/978-3-319-49340-4_8"
}

@inproceedings{DBLP:conf/iceis/GalkinAVS17,
  author       = {Mikhail Galkin and
                  S{\"{o}}ren Auer and
                  Maria{-}Esther Vidal and
                  Simon Scerri},
  editor       = {Slimane Hammoudi and
                  Michal Smialek and
                  Olivier Camp and
                  Joaquim Filipe},
  title        = {Enterprise Knowledge Graphs: {A} Semantic Approach for Knowledge Management
                  in the Next Generation of Enterprise Information Systems},
  booktitle    = {{ICEIS} 2017 - Proceedings of the 19th International Conference on
                  Enterprise Information Systems, Volume 2, Porto, Portugal, April 26-29,
                  2017},
  pages        = {88--98},
  publisher    = {SciTePress},
  year         = {2017},
  doi          = {10.5220/0006325200880098}
}

@article{DBLP:journals/ws/TrivelaSCS15,
  author       = {Despoina Trivela and
                  Giorgos Stoilos and
                  Alexandros Chortaras and
                  Giorgos B. Stamou},
  title        = {Optimising resolution-based rewriting algorithms for {OWL} ontologies},
  journal      = {J. Web Semant.},
  volume       = {33},
  pages        = {30--49},
  year         = {2015},
  doi          = {10.1016/j.websem.2015.02.001}
}

@inproceedings{DBLP:conf/aaai/GlimmKKS15,
  author       = {Birte Glimm and
                  Yevgeny Kazakov and
                  Ilianna Kollia and
                  Giorgos B. Stamou},
  editor       = {Blai Bonet and
                  Sven Koenig},
  title        = {Lower and Upper Bounds for {SPARQL} Queries over {OWL} Ontologies},
  booktitle    = {Proceedings of the Twenty-Ninth {AAAI} Conference on Artificial Intelligence,
                  January 25-30, 2015, Austin, Texas, {USA}},
  pages        = {109--115},
  publisher    = {{AAAI} Press},
  year         = {2015},
  doi          = {10.1609/aaai.v29i1.9192}
}

@book{DBLP:series/synthesis/2020Hamilton,
  author       = {William L. Hamilton},
  title        = {Graph Representation Learning},
  series       = {Synthesis Lectures on Artificial Intelligence and Machine Learning},
  publisher    = {Morgan {\&} Claypool Publishers},
  year         = {2020},
  url          = {https://doi.org/10.2200/S01045ED1V01Y202009AIM046},
  doi          = {10.2200/S01045ED1V01Y202009AIM046},
  bibsource    = {dblp computer science bibliography, https://dblp.org}
}

@inproceedings{csimcsek2019rocketrml,
  title={{RocketRML-A NodeJS implementation of a use-case specific RML mapper}},
  author={{\c{S}}im{\c{s}}ek, Umutcan and K{\"a}rle, Elias and Fensel, Dieter},
  booktitle={Proceeding of the First International Workshop on Knowledge Graph Building},
  year={2019}
}

@article{TimBernersLee2001,
author = {Tim Berners-Lee and James Hendler and Orla Lassila},
title = "{The Semantic Web}",
journal = {Scientific American},
pages = {1-4},
year  = {2001},
}

@article{Noy19,
title	= "{Industry-scale Knowledge Graphs: Lessons and Challenges}",
author	= {Natasha Noy and Yuqing Gao and Anshu Jain and Anant Narayanan and Alan Patterson and Jamie Taylor},
year	= {2019},
URL	= {https://cacm.acm.org/magazines/2019/8/238342-industry-scale-knowledge-graphs/fulltext},
journal	= {Communications of the ACM},
pages	= {36-43},
volume	= {62 (8)}
}

@article{DBLP:journals/pvldb/RabbaniLH23,
  author       = {Kashif Rabbani and
                  Matteo Lissandrini and
                  Katja Hose},
  title        = "{Extraction of Validating Shapes from very large Knowledge Graphs}",
  journal      = {Proc. {VLDB} Endow.},
  volume       = {16},
  number       = {5},
  pages        = {1023--1032},
  year         = {2023}
}

@inproceedings{DBLP:conf/sigmod/RabbaniLH23,
  author       = {Kashif Rabbani and
                  Matteo Lissandrini and
                  Katja Hose},
  title        = "{{SHACTOR:} Improving the Quality of Large-Scale Knowledge Graphs with
                  Validating Shapes}",
  booktitle    = {{SIGMOD} Conference Companion},
  pages        = {151--154},
  publisher    = {{ACM}},
  year         = {2023}
}

@inproceedings{DBLP:conf/emnlp/CabotN21,
  author       = {Pere{-}Llu{\'{\i}}s Huguet Cabot and
                  Roberto Navigli},
  editor       = {Marie{-}Francine Moens and
                  Xuanjing Huang and
                  Lucia Specia and
                  Scott Wen{-}tau Yih},
  title        = {{REBEL:} Relation Extraction By End-to-end Language generation},
  booktitle    = {Findings of the Association for Computational Linguistics: {EMNLP}
                  2021, Virtual Event / Punta Cana, Dominican Republic, 16-20 November,
                  2021},
  pages        = {2370--2381},
  publisher    = {Association for Computational Linguistics},
  year         = {2021},
  url          = {https://doi.org/10.18653/v1/2021.findings-emnlp.204},
  doi          = {10.18653/V1/2021.FINDINGS-EMNLP.204},
  bibsource    = {dblp computer science bibliography, https://dblp.org}
}

@inproceedings{DBLP:conf/edbt/RabbaniLH21,
  author       = {Kashif Rabbani and
                  Matteo Lissandrini and
                  Katja Hose},
  title        = "{Optimizing {SPARQL} Queries using Shape Statistics}",
  booktitle    = {{EDBT}},
  pages        = {505--510},
  publisher    = {OpenProceedings.org},
  year         = {2021}
}

@inproceedings{DBLP:conf/www/RabbaniLH22,
  author       = {Kashif Rabbani and
                  Matteo Lissandrini and
                  Katja Hose},
  title        = "{{SHACL} and ShEx in the Wild: {A} Community Survey on Validating Shapes
                  Generation and Adoption}",
  booktitle    = {{WWW} (Companion Volume)},
  pages        = {260--263},
  publisher    = {{ACM}},
  year         = {2022}
}

@inproceedings{DBLP:conf/owled/JuppBS08,
  author       = {Simon Jupp and
                  Sean Bechhofer and
                  Robert Stevens},
  editor       = {Catherine Dolbear and
                  Alan Ruttenberg and
                  Ulrike Sattler},
  title        = {{SKOS} with {OWL:} Don't be Full-ish!},
  booktitle    = {Proceedings of the Fifth {OWLED} Workshop on {OWL:} Experiences and
                  Directions, collocated with the 7th International Semantic Web Conference
                  (ISWC-2008), Karlsruhe, Germany, October 26-27, 2008},
  series       = {{CEUR} Workshop Proceedings},
  volume       = {432},
  publisher    = {CEUR-WS.org},
  year         = {2008},
  url          = {https://ceur-ws.org/Vol-432/owled2008eu\_submission\_22.pdf},
  bibsource    = {dblp computer science bibliography, https://dblp.org}
}

@article{DBLP:journals/sLogica/BaaderS01,
  author       = {Franz Baader and
                  Ulrike Sattler},
  title        = {An Overview of Tableau Algorithms for Description Logics},
  journal      = {Stud Logica},
  volume       = {69},
  number       = {1},
  pages        = {5--40},
  year         = {2001},
  url          = {https://doi.org/10.1023/A:1013882326814},
  doi          = {10.1023/A:1013882326814},
  bibsource    = {dblp computer science bibliography, https://dblp.org}
}

@article{DBLP:journals/corr/abs-2302-04023,
  author       = {Yejin Bang and
                  Samuel Cahyawijaya and
                  Nayeon Lee and
                  Wenliang Dai and
                  Dan Su and
                  Bryan Wilie and
                  Holy Lovenia and
                  Ziwei Ji and
                  Tiezheng Yu and
                  Willy Chung and
                  Quyet V. Do and
                  Yan Xu and
                  Pascale Fung},
  title        = "{A Multitask, Multilingual, Multimodal Evaluation of ChatGPT on Reasoning, Hallucination, and Interactivity}",
  journal      = {CoRR},
  volume       = {abs/2302.04023},
  year         = {2023}
}

@inproceedings{DBLP:conf/adbis/Hose23,
  author       = {Katja Hose},
  title        = {Knowledge Engineering in the Era of Artificial Intelligence},
  booktitle    = {{ADBIS}},
  series       = {Lecture Notes in Computer Science},
  volume       = {13985},
  pages        = {3--15},
  publisher    = {Springer},
  year         = {2023}
}

@inproceedings{DBLP:conf/semweb/KelesH19,
  author       = {Ilkcan Keles and
                  Katja Hose},
  title        = "{Skyline Queries over Knowledge Graphs}",
  booktitle    = {{ISWC}},
  series       = {Lecture Notes in Computer Science},
  volume       = {11778},
  pages        = {293--310},
  publisher    = {Springer},
  year         = {2019}
}

@inproceedings{DBLP:conf/esws/ChengH22,
  author       = {Sijin Cheng and
                  Olaf Hartig},
  title        = "{Towards Query Processing over Heterogeneous Federations of {RDF} Data Sources}",
  booktitle    = {{ESWC} (Satellite Events)},
  series       = {Lecture Notes in Computer Science},
  volume       = {13384},
  pages        = {57--62},
  publisher    = {Springer},
  year         = {2022}
}

@inproceedings{DBLP:conf/www/HelingA22,
  author       = {Lars Heling and
                  Maribel Acosta},
  title        = "{Federated {SPARQL} Query Processing over Heterogeneous Linked Data Fragments}",
  booktitle    = {{WWW}},
  pages        = {1047--1057},
  publisher    = {{ACM}},
  year         = {2022}
}

@inproceedings{DBLP:conf/semweb/MontoyaAH18a,
  author       = {Gabriela Montoya and
                  Christian Aebeloe and
                  Katja Hose},
  title        = "{Towards Efficient Query Processing over Heterogeneous {RDF} Interfaces}",
  booktitle    = {{ISWC} (Best Workshop Papers)},
  series       = {Studies on the Semantic Web},
  volume       = {36},
  pages        = {39--53},
  publisher    = {{IOS} Press},
  year         = {2018}
}

@inproceedings{DBLP:conf/semweb/MontoyaKH19,
  author       = {Gabriela Montoya and
                  Ilkcan Keles and
                  Katja Hose},
  title        = "{Analysis of the Effect of Query Shapes on Performance over {LDF} Interfaces}",
  booktitle    = {QuWeDa@ISWC},
  series       = {{CEUR} Workshop Proceedings},
  volume       = {2496},
  pages        = {51--66},
  publisher    = {CEUR-WS.org},
  year         = {2019}
}

@inproceedings{DBLP:conf/semweb/HansenLGLTH20,
  author       = {Emil Riis Hansen and
                  Matteo Lissandrini and
                  Agneta Ghose and
                  S{\o}ren L{\o}kke and
                  Christian Thomsen and
                  Katja Hose},
  title        = "{Transparent Integration and Sharing of Life Cycle Sustainability Data with Provenance}",
  booktitle    = {{ISWC}},
  series       = {Lecture Notes in Computer Science},
  volume       = {12507},
  pages        = {378--394},
  publisher    = {Springer},
  year         = {2020}
}

@inproceedings{DBLP:conf/esws/IbragimovHPZ15,
  author       = {Dilshod Ibragimov and
                  Katja Hose and
                  Torben Bach Pedersen and
                  Esteban Zim{\'{a}}nyi},
  title        = "{Processing Aggregate Queries in a Federation of {SPARQL} Endpoints}",
  booktitle    = {{ESWC}},
  series       = {Lecture Notes in Computer Science},
  volume       = {9088},
  pages        = {269--285},
  publisher    = {Springer},
  year         = {2015}
}

@inproceedings{DBLP:conf/semweb/GalarragaMH17,
  author       = {Luis Gal{\'{a}}rraga and
                  Kim Ahlstr{\o}m Meyn Mathiassen and
                  Katja Hose},
  title        = "{QBOAirbase: The European Air Quality Database as an {RDF} Cube}",
  booktitle    = {{ISWC} (Posters, Demos {\&} Industry Tracks)},
  series       = {{CEUR} Workshop Proceedings},
  volume       = {1963},
  publisher    = {CEUR-WS.org},
  year         = {2017}
}

@inproceedings{Chaves-FragaEIC19,
  author       = {David Chaves{-}Fraga and
                  Kemele M. Endris and
                  Enrique Iglesias and
                  {\'{O}}scar Corcho and
                  Maria{-}Esther Vidal},
  title        = {What Are the Parameters that Affect the Construction of a Knowledge
                  Graph?},
  booktitle    = {On the Move to Meaningful Internet Systems: {OTM} 2019 Conferences
                  - Confederated International Conferences: CoopIS, ODBASE, C{\&}TC
                  2019, Rhodes, Greece, October 21-25, 2019, Proceedings},
  series       = {Lecture Notes in Computer Science},
  volume       = {11877},
  pages        = {695--713},
  publisher    = {Springer},
  year         = {2019},
  url          = {https://doi.org/10.1007/978-3-030-33246-4\_43},
  doi          = {10.1007/978-3-030-33246-4\_43}
}

@inproceedings{iglesias2020sdm,
  author       = {Enrique Iglesias and
                  Samaneh Jozashoori and
                  David Chaves{-}Fraga and
                  Diego Collarana and
                  Maria{-}Esther Vidal},
  title        = {SDM-RDFizer: An {RML} Interpreter for the Efficient Creation of {RDF}
                  Knowledge Graphs},
  journal      = {CoRR},
  volume       = {abs/2008.07176},
  year         = {2020},
  url          = {https://arxiv.org/abs/2008.07176}
}

@article{arenas2022morph,
  title   = {{Morph-KGC: Scalable knowledge graph materialization with mapping partitions}},
  author  = {Arenas-Guerrero, Julián and Chaves-Fraga, David and Toledo, Jhon and Pérez, María S. and Corcho, Oscar},
  journal = {Semantic Web},
  year    = {2022},
  doi     = {10.3233/SW-223135}
}

@article{chebotko2009semantics,
  author       = {Artem Chebotko and
                  Shiyong Lu and
                  Farshad Fotouhi},
  title        = {Semantics preserving {SPARQL-to-SQL} translation},
  journal      = {Data Knowl. Eng.},
  volume       = {68},
  number       = {10},
  pages        = {973--1000},
  year         = {2009},
  url          = {https://doi.org/10.1016/j.datak.2009.04.001},
  doi          = {10.1016/j.datak.2009.04.001}
}

@misc{w3c-did,
  author= {Manu Sporny and Amy Guy and Markus Sabadello and Drummond Reed and Manu Sporny and Dave Longley and Markus Sabadello and Drummond Reed and Orie Steele and Christopher Allen},
  title = {Decentralized Identifiers ({DIDs}) v1.0: Core architecture, data model, and representations},
  howpublished = {\url{https://www.w3.org/TR/did-core/}},
  year = {2012}
}

@inproceedings{lenzerini2002data,
  author       = {Maurizio Lenzerini},
  title        = {Data Integration: {A} Theoretical Perspective},
  booktitle    = {Proceedings of the Twenty-first {ACM} {SIGACT-SIGMOD-SIGART} Symposium
                  on Principles of Database Systems, June 3-5, Madison, Wisconsin, {USA}},
  pages        = {233--246},
  publisher    = {{ACM}},
  year         = {2002},
  url          = {https://doi.org/10.1145/543613.543644},
  doi          = {10.1145/543613.543644}
}

@inproceedings{DimouSCVMW14,
  author       = {Anastasia Dimou and
                  Miel Vander Sande and
                  Pieter Colpaert and
                  Ruben Verborgh and
                  Erik Mannens and
                  Rik Van de Walle},
  editor       = {Christian Bizer and
                  Tom Heath and
                  S{\"{o}}ren Auer and
                  Tim Berners{-}Lee},
  title        = {{RML:} {A} Generic Language for Integrated {RDF} Mappings of Heterogeneous
                  Data},
  booktitle    = {Proceedings of the Workshop on Linked Data on the Web co-located with
                  the 23rd International World Wide Web Conference {(WWW} 2014), Seoul,
                  Korea, April 8, 2014},
  series       = {{CEUR} Workshop Proceedings},
  volume       = {1184},
  publisher    = {CEUR-WS.org},
  year         = {2014},
  url          = {https://ceur-ws.org/Vol-1184/ldow2014\_paper\_01.pdf}
}

@article{DBLP:journals/ws/AsscheDHHMD23,
  author       = {Dylan Van Assche and
                  Thomas Delva and
                  Gerald Haesendonck and
                  Pieter Heyvaert and
                  Ben De Meester and
                  Anastasia Dimou},
  title        = {Declarative {RDF} graph generation from heterogeneous (semi-)structured
                  data: {A} systematic literature review},
  journal      = {J. Web Semant.},
  volume       = {75},
  pages        = {100753},
  year         = {2023},
  url          = {https://doi.org/10.1016/j.websem.2022.100753},
  doi          = {10.1016/j.websem.2022.100753}
}

@article{DBLP:journals/semweb/AisoposJNPRSIVM23,
  author       = {Fotis Aisopos and
                  Samaneh Jozashoori and
                  Emetis Niazmand and
                  Disha Purohit and
                  Ariam Rivas and
                  Ahmad Sakor and
                  Enrique Iglesias and
                  Dimitrios Vogiatzis and
                  Ernestina Menasalvas and
                  Alejandro Rodr{\'{\i}}guez Gonz{\'{a}}lez and
                  Guillermo Vigueras and
                  Daniel G{\'{o}}mez{-}Bravo and
                  Maria Torrente and
                  Roberto Hern{\'{a}}ndez L{\'{o}}pez and
                  Mariano Provencio Pulla and
                  Athanasios Dalianis and
                  Anna Triantafillou and
                  Georgios Paliouras and
                  Maria{-}Esther Vidal},
  title        = {Knowledge graphs for enhancing transparency in health data ecosystems},
  journal      = {Semantic Web},
  volume       = {14},
  number       = {5},
  pages        = {943--976},
  year         = {2023}
}

@inproceedings{DBLP:conf/semweb/MontoyaSH17,
  author       = {Gabriela Montoya and
                  Hala Skaf{-}Molli and
                  Katja Hose},
  title        = "{The Odyssey Approach for Optimizing Federated {SPARQL} Queries}",
  booktitle    = {{ISWC}},
  series       = {Lecture Notes in Computer Science},
  volume       = {10587},
  pages        = {471--489},
  publisher    = {Springer},
  year         = {2017}
}

@inproceedings{DBLP:conf/www/Galarraga0KH23,
  author       = {Luis Gal{\'{a}}rraga and
                  Daniel Hern{\'{a}}ndez and
                  Anas Katim and
                  Katja Hose},
  title        = {Visualizing How-Provenance Explanations for {SPARQL} Queries},
  booktitle    = {{WWW} (Companion Volume)},
  pages        = {212--216},
  publisher    = {{ACM}},
  year         = {2023}
}

@inproceedings{DBLP:conf/www/AzzamAMKPH21,
  author       = {Amr Azzam and
                  Christian Aebeloe and
                  Gabriela Montoya and
                  Ilkcan Keles and
                  Axel Polleres and
                  Katja Hose},
  title        = "{WiseKG: Balanced Access to Web Knowledge Graphs}",
  booktitle    = {{WWW} '21: The Web Conference 2021, Virtual Event / Ljubljana, Slovenia,
                  April 19-23, 2021},
  pages        = {1422--1434},
  publisher    = {{ACM} / {IW3C2}},
  year         = {2021}
}

@inproceedings{DBLP:conf/f-ic/MinierSM19,
  author       = {Thomas Minier and
                  Hala Skaf{-}Molli and
                  Pascal Molli},
  editor       = {Nathalie Hernandez},
  title        = {SaGe : pr{\'{e}}emption Web pour les services publics d'{\'{e}}valuation
                  de requ{\^{e}}tes {SPARQL}},
  booktitle    = {{IC} 2019: 30es Journ{\'{e}}es francophones d'Ing{\'{e}}nierie
                  des Connaissances (Proceedings of the 30th French Knowledge Engineering
                  Conference), Toulouse, France, July 2-4, 2019},
  pages        = {141},
  year         = {2019}
}

@article{DBLP:journals/corr/HartigA16,
  author       = {Olaf Hartig and
                  Carlos Buil Aranda},
  title        = {brTPF: Bindings-Restricted Triple Pattern Fragments (Extended Preprint)},
  journal      = {CoRR},
  volume       = {abs/1608.08148},
  year         = {2016}
}

@inproceedings{DBLP:conf/www/AzzamFABP20,
  author       = {Amr Azzam and
                  Javier D. Fern{\'{a}}ndez and
                  Maribel Acosta and
                  Martin Beno and
                  Axel Polleres},
  editor       = {Yennun Huang and
                  Irwin King and
                  Tie{-}Yan Liu and
                  Maarten van Steen},
  title        = {{SMART-KG:} Hybrid Shipping for {SPARQL} Querying on the Web},
  booktitle    = {{WWW} '20: The Web Conference 2020, Taipei, Taiwan, April 20-24, 2020},
  pages        = {984--994},
  publisher    = {{ACM} / {IW3C2}},
  year         = {2020}
}

@inproceedings{DBLP:conf/semweb/AcostaV15,
  author       = {Maribel Acosta and
                  Maria{-}Esther Vidal},
  title        = {Networks of Linked Data Eddies: An Adaptive Web Query Processing Engine
                  for {RDF} Data},
  booktitle    = {The Semantic Web - {ISWC} 2015 - 14th International Semantic Web Conference,
                  Bethlehem, PA, USA, October 11-15, 2015, Proceedings, Part {I}},
  series       = {Lecture Notes in Computer Science},
  volume       = {9366},
  pages        = {111--127},
  publisher    = {Springer},
  year         = {2015}
}

@inproceedings{DBLP:conf/semweb/AcostaVLCR11,
  author       = {Maribel Acosta and
                  Maria{-}Esther Vidal and
                  Tomas Lampo and
                  Julio Castillo and
                  Edna Ruckhaus},
  editor       = {Lora Aroyo and
                  Chris Welty and
                  Harith Alani and
                  Jamie Taylor and
                  Abraham Bernstein and
                  Lalana Kagal and
                  Natasha Fridman Noy and
                  Eva Blomqvist},
  title        = {{ANAPSID:} An Adaptive Query Processing Engine for {SPARQL} Endpoints},
  booktitle    = {The Semantic Web - {ISWC} 2011 - 10th International Semantic Web Conference,
                  Bonn, Germany, October 23-27, 2011, Proceedings, Part {I}},
  series       = {Lecture Notes in Computer Science},
  volume       = {7031},
  pages        = {18--34},
  publisher    = {Springer},
  year         = {2011}
}

@incollection{DBLP:series/lncs/EndrisVG20,
  author       = {Kemele M. Endris and
                  Maria{-}Esther Vidal and
                  Damien Graux},
  editor       = {Valentina Janev and
                  Damien Graux and
                  Hajira Jabeen and
                  Emanuel Sallinger},
  title        = {Federated Query Processing},
  booktitle    = {Knowledge Graphs and Big Data Processing},
  series       = {Lecture Notes in Computer Science},
  volume       = {12072},
  pages        = {73--86},
  publisher    = {Springer},
  year         = {2020},
  url          = {https://doi.org/10.1007/978-3-030-53199-7\_5},
  doi          = {10.1007/978-3-030-53199-7\_5}
}

@inproceedings{DBLP:conf/pods/ScheufeleM97,
  author    = {Wolfgang Scheufele and
               Guido Moerkotte},
  editor    = {Alberto O. Mendelzon and
               Z. Meral {\"{O}}zsoyoglu},
  title     = {On the Complexity of Generating Optimal Plans with Cross Products},
  booktitle = {Proceedings of the Sixteenth {ACM} {SIGACT-SIGMOD-SIGART} Symposium
               on Principles of Database Systems},
  year      = {1997}
}

@book{DBLP:books/crc/linked2014,
  editor       = {Andreas Harth and
                  Katja Hose and
                  Ralf Schenkel},
  title        = "{Linked Data Management}",
  publisher    = {Chapman and Hall/CRC},
  year         = {2014}
}

@incollection{DBLP:reference/bdt/HartigHS19,
  author       = {Olaf Hartig and
                  Katja Hose and
                  Juan F. Sequeda},
  title        = "{Linked Data Management}",
  booktitle    = {Encyclopedia of Big Data Technologies},
  publisher    = {Springer},
  year         = {2019}
}

@article{DBLP:journals/tods/IbarakiK84,
  author    = {Toshihide Ibaraki and
               Tiko Kameda},
  title     = {On the Optimal Nesting Order for Computing N-Relational Joins},
  journal   = {{ACM} Trans. Database Syst.},
  volume    = {9},
  number    = {3},
  pages     = {482--502},
  year      = {1984}
}

@inproceedings{DBLP:conf/sigmod/HoseS12,
  author       = {Katja Hose and
                  Ralf Schenkel},
  title        = "{Towards benefit-based {RDF} source selection for {SPARQL} queries}",
  booktitle    = {{SWIM}},
  pages        = {2},
  publisher    = {{ACM}},
  year         = {2012}
}

@article{DBLP:journals/tlsdkcs/EndrisGLMVA18,
  author       = {Kemele M. Endris and
                  Mikhail Galkin and
                  Ioanna Lytra and
                  Mohamed Nadjib Mami and
                  Maria{-}Esther Vidal and
                  S{\"{o}}ren Auer},
  title        = {Querying Interlinked Data by Bridging {RDF} Molecule Templates},
  journal      = {Trans. Large Scale Data Knowl. Centered Syst.},
  volume       = {39},
  pages        = {1--42},
  year         = {2018}
}

@article{DBLP:journals/tlsdkcs/VidalCAMP16,
  author       = {Maria{-}Esther Vidal and
                  Sim{\'{o}}n Castillo and
                  Maribel Acosta and
                  Gabriela Montoya and
                  Guillermo Palma},
  title        = {On the Selection of {SPARQL} Endpoints to Efficiently Execute Federated
                  {SPARQL} Queries},
  journal      = {Trans. Large Scale Data Knowl. Centered Syst.},
  volume       = {25},
  pages        = {109--149},
  year         = {2016}
}

@inproceedings{DBLP:conf/esws/VidalRLMSP10,
  author       = {Maria{-}Esther Vidal and
                  Edna Ruckhaus and
                  Tomas Lampo and
                  Amad{\'{\i}}s Mart{\'{\i}}nez and
                  Javier Sierra and
                  Axel Polleres},
  title        = {Efficiently Joining Group Patterns in {SPARQL} Queries},
  booktitle    = {The Extended Semantic
                  Web Conference, {ESWC}},
  year         = {2010}
}

@inproceedings{DBLP:conf/semweb/SchwarteHHSS11,
  author       = {Andreas Schwarte and
                  Peter Haase and
                  Katja Hose and
                  Ralf Schenkel and
                  Michael Schmidt},
  title        = {FedX: Optimization Techniques for Federated Query Processing on Linked
                  Data},
  booktitle    = {The Semantic Web - {ISWC} 2011 - 10th International Semantic Web Conference,
                  Bonn, Germany, October 23-27, 2011, Proceedings, Part {I}},
  series       = {Lecture Notes in Computer Science},
  volume       = {7031},
  pages        = {601--616},
  publisher    = {Springer},
  year         = {2011}
}

@article{Wiederhold92,
  author    = {Gio Wiederhold},
  title     = {Mediators in the Architecture of Future Information Systems},
  journal   = {{IEEE} Computer},
  volume    = {25},
  number    = {3},
  pages     = {38--49},
  year      = {1992}
}

@inproceedings{ZadorozhnyRVUB02,
  author    = {Vladimir Zadorozhny and
               Louiqa Raschid and
               Maria{-}Esther Vidal and
               Tolga Urhan and
               Laura Bright},
  title     = {Efficient evaluation of queries in a mediator for WebSources},
  booktitle = {Proceedings of the 2002 {ACM} {SIGMOD} International Conference on
               Management of Data, Madison, Wisconsin, USA, June 3-6, 2002},
  pages     = {85--96},
  year      = {2002}
}

@inproceedings{DBLP:conf/semweb/VerborghHMHVSCCMW14a,
  author       = {Ruben Verborgh and
                  Olaf Hartig and
                  Ben De Meester and
                  Gerald Haesendonck and
                  Laurens De Vocht and
                  Miel Vander Sande and
                  Richard Cyganiak and
                  Pieter Colpaert and
                  Erik Mannens and
                  Rik Van de Walle},
  editor       = {Matthew Horridge and
                  Marco Rospocher and
                  Jacco van Ossenbruggen},
  title        = {Low-Cost Queryable Linked Data through Triple Pattern Fragments},
  booktitle    = {Proceedings of the {ISWC} 2014 Posters {\&} Demonstrations Track
                  a track within the 13th International Semantic Web Conference, {ISWC}
                  2014, Riva del Garda, Italy, October 21, 2014},
  series       = {{CEUR} Workshop Proceedings},
  volume       = {1272},
  pages        = {13--16},
  publisher    = {CEUR-WS.org},
  year         = {2014}
}

@article{DBLP:journals/corr/abs-2002-09172,
  author       = {Christian Aebeloe and
                  Ilkcan Keles and
                  Gabriela Montoya and
                  Katja Hose},
  title        = "{Star Pattern Fragments: Accessing Knowledge Graphs through Star Patterns}",
  volume       = {abs/2002.09172},
  year         = {2020},
  url          = {https://arxiv.org/abs/2002.09172}
}

@article{DBLP:journals/ws/VerborghSHHVMHC16,
  author       = {Ruben Verborgh and
                  Miel Vander Sande and
                  Olaf Hartig and
                  Joachim Van Herwegen and
                  Laurens De Vocht and
                  Ben De Meester and
                  Gerald Haesendonck and
                  Pieter Colpaert},
  title        = {Triple Pattern Fragments: {A} low-cost knowledge graph interface for
                  the Web},
  journal      = {J. Web Semant.},
  volume       = {37-38},
  pages        = {184--206},
  year         = {2016},
  url          = {https://doi.org/10.1016/j.websem.2016.03.003}
}

@incollection{DBLP:books/degruyter/20/ScherpG20,
  author       = {Ansgar Scherp and
                  Gerd Gr{\"{o}}ner},
  editor       = {G{\"{u}}nther G{\"{o}}rz and
                  Ute Schmid and
                  Tanya Braun},
  title        = {Semantic Web},
  booktitle    = {Handbuch der K{\"{u}}nstlichen Intelligenz, 6. Auflage},
  pages        = {783--816},
  publisher    = {De Gruyter},
  year         = {2020},
  url          = {https://doi.org/10.1515/9783110659948-018},
  doi          = {10.1515/9783110659948-018},
  bibsource    = {dblp computer science bibliography, https://dblp.org}
}

@incollection{DBLP:books/ol/13/GronerSS13,
  author       = {Gerd Gr{\"{o}}ner and
                  Ansgar Scherp and
                  Steffen Staab},
  editor       = {G{\"{u}}nther G{\"{o}}rz and
                  Josef Schneeberger and
                  Ute Schmid},
  title        = {Semantic Web},
  booktitle    = {Handbuch der K{\"{u}}nstlichen Intelligenz, 5. Auflage},
  pages        = {585--612},
  publisher    = {Oldenbourg Wissenschaftsverlag},
  year         = {2013},
  url          = {https://doi.org/10.1524/9783486719796.585},
  doi          = {10.1524/9783486719796.585},
  bibsource    = {dblp computer science bibliography, https://dblp.org}
}

@inproceedings{DBLP:conf/esws/KastenSS14,
  author    = {Andreas Kasten and
               Ansgar Scherp and
               Peter Schau{\ss}},
  title     = {A Framework for Iterative Signing of Graph Data on the Web},
  booktitle = {Extended Semntic Web Conference},
  no_pages     = {146--160},
  publisher = {Springer},
  year      = {2014},
}

@phdthesis{DBLP:phd/dnb/Kasten16,
  author    = {Andreas Kasten},
  title     = {Secure semantic web data management: confidentiality, integrity, and
               compliant availability in open and distributed networks},
  school    = {University of Koblenz and Landau, Germany},
  year      = {2016},
  url       = {https://kola.opus.hbz-nrw.de/frontdoor/index/index/docId/1393},
  urn       = {urn:nbn:de:kola-13939},
  bibsource = {dblp computer science bibliography, https://dblp.org}
}

@inproceedings{DBLP:conf/semweb/SchonteichKS18,
  author    = {Falko Sch{\"{o}}nteich and
               Andreas Kasten and
               Ansgar Scherp},
  title     = {A Pattern-Based Core Ontology for Product Lifecycle Management based
               on {DUL}},
  booktitle = {Workshop on Ontology Design and Patterns},
  no_pages     = {92--106},
  year      = {2018},
  series    = {{CEUR} Workshop Proceedings},
  volume    = {2195},
  publisher = {CEUR-WS.org},
  url       = {http://ceur-ws.org/Vol-2195},
}

@article{emf-mtap,
  author    = {Ansgar Scherp and
               Thomas Franz and
               Carsten Saathoff and
               Steffen Staab},
  title     = {A core ontology on events for representing occurrences in
               the real world},
  journal   = {Multimedia Tools Appl.},
  volume    = {58},
  number    = {2},
  year      = {2012},
  pages     = {293-331},
  ee        = {http://dx.doi.org/10.1007/s11042-010-0667-z},
  bibsource = {DBLP, http://dblp.uni-trier.de}
}

@inproceedings{www-m3o,
  author    = {Carsten Saathoff and
               Ansgar Scherp},
  title     = {Unlocking the semantics of multimedia presentations in the
               web with the multimedia metadata ontology},
  no_pages     = {831-840},
  booktitle     = {International Conference on World Wide Web},
  publisher = {ACM},
  year      = {2010},
}

@article{Oberle:2007:DES:1290204.1290385,
 author = {Oberle, Daniel and Ankolekar, Anupriya and Hitzler, Pascal and Cimiano, Philipp and Sintek, Michael and Kiesel, Malte and Mougouie, Babak and Baumann, Stephan and Vembu, Shankar and Romanelli, Massimo and Buitelaar, Paul and Engel, Ralf and Sonntag, Daniel and Reithinger, Norbert and Loos, Berenike and Zorn, Hans-Peter and Micelli, Vanessa and Porzel, Robert and Schmidt, Christian and Weiten, Moritz and Burkhardt, Felix and Zhou, Jianshen},
 title = {{DOLCE ergo SUMO: On foundational and domain models in the SmartWeb Integrated Ontology (SWIntO)}},
 journal = {Web Semant.},
 issue_date = {September, 2007},
 volume = {5},
 number = {3},
 month = sep,
 year = {2007},
 issn = {1570-8268},
 pages = {156--174},
 numpages = {19},
 url = {http://dx.doi.org/10.1016/j.websem.2007.06.002},
 doi = {10.1016/j.websem.2007.06.002},
 acmid = {1290385},
 publisher = {Elsevier Science Publishers B. V.},
 address = {Amsterdam, The Netherlands, The Netherlands},
}

@article{ScherpEtAlDesigningCoreOntologiesIOS2011,
  author    = {Ansgar Scherp and Carsten Saathoff and Thomas Franz and Steffen Staab},
  title     = {Designing core ontologies},
  journal   = {Applied Ontology},
  volume    = {6},
  number    = {3},
  year      = {2011},
  pages     = {177-221},
  ee        = {http://dx.doi.org/10.3233/AO-2011-0096},
  bibsource = {DBLP, http://dblp.uni-trier.de}
}

@INBOOK{BorgoMasoloDOLCE2009,
  chapter = {Foundational choices in {DOLCE}},
  title = {Handbook on Ontologies},
  publisher = {Springer},
  year = {2009},
  author = {Stefano Borgo and Claudio Masolo},
  edition = {2nd}
}

@inbook{Euzenat2007a,
  chapter = {Classifications of ontology matching techniques},
  title = {Ontology matching},
  publisher = {Springer},
  year = {2007},
  author = {J\'er\^ome Euzenat and Pavel Shvaiko},
  isbn = {3-540-49611-4},
  language = {english},
  page = {341}
}

@INBOOK{GangemiPresuttiODP2009,
  chapter = {Ontology Design Patterns},
  title = {Handbook on Ontologies},
  publisher = {Springer},
  year = {2009},
  author = {Aldo Gangemi and Valentina Presutti},
}

@article{DBLP:journals/vldb/SahuMSLO20,
  author       = {Siddhartha Sahu and
                  Amine Mhedhbi and
                  Semih Salihoglu and
                  Jimmy Lin and
                  M. Tamer {\"{O}}zsu},
  title        = {The ubiquity of large graphs and surprising challenges of graph processing:
                  extended survey},
  journal      = {{VLDB} J.},
  volume       = {29},
  number       = {2-3},
  pages        = {595--618},
  year         = {2020},
  url          = {https://doi.org/10.1007/s00778-019-00548-x},
  doi          = {10.1007/S00778-019-00548-X},
  bibsource    = {dblp computer science bibliography, https://dblp.org}
}

@article{DBLP:journals/cacm/SakrBVIAAAABBDV21,
  author       = {Sherif Sakr and
                  Angela Bonifati and
                  Hannes Voigt and
                  Alexandru Iosup and
                  Khaled Ammar and
                  Renzo Angles and
                  Walid G. Aref and
                  Marcelo Arenas and
                  Maciej Besta and
                  Peter A. Boncz and
                  Khuzaima Daudjee and
                  Emanuele Della Valle and
                  Stefania Dumbrava and
                  Olaf Hartig and
                  Bernhard Haslhofer and
                  Tim Hegeman and
                  Jan Hidders and
                  Katja Hose and
                  Adriana Iamnitchi and
                  Vasiliki Kalavri and
                  Hugo Kapp and
                  Wim Martens and
                  M. Tamer {\"{O}}zsu and
                  Eric Peukert and
                  Stefan Plantikow and
                  Mohamed Ragab and
                  Matei Ripeanu and
                  Semih Salihoglu and
                  Christian Schulz and
                  Petra Selmer and
                  Juan F. Sequeda and
                  Joshua Shinavier and
                  G{\'{a}}bor Sz{\'{a}}rnyas and
                  Riccardo Tommasini and
                  Antonino Tumeo and
                  Alexandru Uta and
                  Ana Lucia Varbanescu and
                  Hsiang{-}Yun Wu and
                  Nikolay Yakovets and
                  Da Yan and
                  Eiko Yoneki},
  title        = {The future is big graphs: a community view on graph processing systems},
  journal      = {Commun. {ACM}},
  volume       = {64},
  number       = {9},
  pages        = {62--71},
  year         = {2021},
  url          = {https://doi.org/10.1145/3434642},
  doi          = {10.1145/3434642},
  bibsource    = {dblp computer science bibliography, https://dblp.org}
}

@article{DBLP:journals/cacm/NoyGJNPT19,
  author       = {Natalya Fridman Noy and
                  Yuqing Gao and
                  Anshu Jain and
                  Anant Narayanan and
                  Alan Patterson and
                  Jamie Taylor},
  title        = {Industry-scale knowledge graphs: lessons and challenges},
  journal      = {Commun. {ACM}},
  volume       = {62},
  number       = {8},
  pages        = {36--43},
  year         = {2019},
  url          = {https://doi.org/10.1145/3331166},
  doi          = {10.1145/3331166},
  bibsource    = {dblp computer science bibliography, https://dblp.org}
}

@article{DBLP:journals/corr/abs-2305-04676,
  author       = {Milena Trajanoska and
                  Riste Stojanov and
                  Dimitar Trajanov},
  title        = {Enhancing Knowledge Graph Construction Using Large Language Models},
  journal      = {CoRR},
  volume       = {abs/2305.04676},
  year         = {2023},
  url          = {https://doi.org/10.48550/arXiv.2305.04676},
  doi          = {10.48550/arXiv.2305.04676},
  eprinttype    = {arXiv},
  eprint       = {2305.04676},
  bibsource    = {dblp computer science bibliography, https://dblp.org}
}

@article{DBLP:journals/eswa/HabernalK13,
  author       = {Ivan Habernal and
                  Miloslav Konop{\'{\i}}k},
  title        = {{SWSNL:} Semantic Web Search Using Natural Language},
  journal      = {Expert Syst. Appl.},
  volume       = {40},
  number       = {9},
  pages        = {3649--3664},
  year         = {2013},
  url          = {https://doi.org/10.1016/j.eswa.2012.12.070},
  doi          = {10.1016/j.eswa.2012.12.070},
  bibsource    = {dblp computer science bibliography, https://dblp.org}
}

@BOOK{OberleMiddleware2006,
  title = {Semantic Management of Middleware},
  publisher = {Springer},
  year = {2006},
  author = {Daniel Oberle}
}

@incollection{GuarinoOberleStaabWhatIsAnOntology2009,
  title = {What is an Ontology?},
  booktitle = {Handbook on Ontologies},
  publisher = {Springer},
  year = {2009},
  editor = {Steffen Staab and Ruder Studer},
  author = {Daniel Oberle and Nicola Guarino and Steffen Staab},
  edition = {2nd}
}

@phdthesis{DBLP:phd/dnb/Blume22,
  author       = {Till Blume},
  title        = {Semantic structural graph summaries for evolving and distributed graphs},
  school       = {University of Ulm, Germany},
  year         = {2022},
  url          = {https://nbn-resolving.org/urn:nbn:de:bsz:289-oparu-46050-1},
  urn          = {urn:nbn:de:bsz:289-oparu-46050-1},
  bibsource    = {dblp computer science bibliography, https://dblp.org}
}

@article{DBLP:journals/corr/abs-2003-02320,
  author       = {Aidan Hogan and
                  Eva Blomqvist and
                  Michael Cochez and
                  Claudia d'Amato and
                  Gerard de Melo and
                  Claudio Gutierrez and
                  Jos{\'{e}} Emilio Labra Gayo and
                  Sabrina Kirrane and
                  Sebastian Neumaier and
                  Axel Polleres and
                  Roberto Navigli and
                  Axel{-}Cyrille Ngonga Ngomo and
                  Sabbir M. Rashid and
                  Anisa Rula and
                  Lukas Schmelzeisen and
                  Juan F. Sequeda and
                  Steffen Staab and
                  Antoine Zimmermann},
  title        = {Knowledge Graphs},
  journal      = {CoRR},
  volume       = {abs/2003.02320},
  year         = {2020},
  url          = {https://arxiv.org/abs/2003.02320},
  eprinttype    = {arXiv},
  eprint       = {2003.02320},
  bibsource    = {dblp computer science bibliography, https://dblp.org}
}

@incollection{DBLP:books/sp/10/Wood10,
  author       = {David Wood},
  editor       = {David Wood},
  title        = {Reliable and Persistent Identification of Linked Data Elements},
  booktitle    = {Linking Enterprise Data},
  pages        = {149--173},
  publisher    = {Springer},
  year         = {2010},
  url          = {https://doi.org/10.1007/978-1-4419-7665-9\_8},
  doi          = {10.1007/978-1-4419-7665-9\_8},
  bibsource    = {dblp computer science bibliography, https://dblp.org}
}

@article{DBLP:journals/pvldb/NeumannW08,
  author       = {Thomas Neumann and
                  Gerhard Weikum},
  title        = {{RDF-3X:} a RISC-style engine for {RDF}},
  journal      = {Proc. {VLDB} Endow.},
  volume       = {1},
  number       = {1},
  pages        = {647--659},
  year         = {2008},
  doi          = {10.14778/1453856.1453927}
}

@inproceedings{DBLP:conf/rweb/HoseSTW11,
  author       = {Katja Hose and
                  Ralf Schenkel and
                  Martin Theobald and
                  Gerhard Weikum},
  title        = "{Database Foundations for Scalable {RDF} Processing}",
  booktitle    = {Reasoning Web},
  series       = {Lecture Notes in Computer Science},
  volume       = {6848},
  pages        = {202--249},
  publisher    = {Springer},
  year         = {2011}
}

@inproceedings{DBLP:conf/www/StockerSBKR08,
  author       = {Markus Stocker and
                  Andy Seaborne and
                  Abraham Bernstein and
                  Christoph Kiefer and
                  Dave Reynolds},
  editor       = {Jinpeng Huai and
                  Robin Chen and
                  Hsiao{-}Wuen Hon and
                  Yunhao Liu and
                  Wei{-}Ying Ma and
                  Andrew Tomkins and
                  Xiaodong Zhang},
  title        = {{SPARQL} basic graph pattern optimization using selectivity estimation},
  booktitle    = {Proceedings of the 17th International Conference on World Wide Web,
                  {WWW} 2008, Beijing, China, April 21-25, 2008},
  pages        = {595--604},
  publisher    = {{ACM}},
  year         = {2008},
  doi          = {10.1145/1367497.1367578}
}

@inproceedings{DBLP:conf/semweb/FerradaBH20,
  author       = {Sebasti{\'{a}}n Ferrada and
                  Benjamin Bustos and
                  Aidan Hogan},
  title        = {Extending {SPARQL} with Similarity Joins},
  booktitle    = {{ISWC} {(1)}},
  series       = {Lecture Notes in Computer Science},
  volume       = {12506},
  pages        = {201--217},
  publisher    = {Springer},
  year         = {2020}
}

@inproceedings{DBLP:conf/kcap/LoustaunauH21,
  author       = {Alberto Moya Loustaunau and
                  Aidan Hogan},
  title        = {Predicting {SPARQL} Query Dynamics},
  booktitle    = {{K-CAP}},
  pages        = {161--168},
  publisher    = {{ACM}},
  year         = {2021}
}

@inproceedings{DBLP:conf/edbt/Gubichev014,
  author       = {Andrey Gubichev and
                  Thomas Neumann},
  title        = {Exploiting the query structure for efficient join ordering in {SPARQL} queries},
  booktitle    = {{EDBT}},
  pages        = {439--450},
  publisher    = {OpenProceedings.org},
  year         = {2014}
}

@book{GomezPerezFernandezLopezOntologicalEngineering,
  author = {Asunci{\'o}n G{\'o}mez-P{\'e}rez and Mariano F{\'e}rn{\'a}ndez L{\'o}pez and Oscar Corcho},
  publisher = {Springer},
  title = {Ontological engineering},
  year = {2004},
  isbn = {1-85233-551-3},
}

@article{DBLP:journals/corr/abs-2310-13227-toolchainstar,
  author       = {Yuchen Zhuang and
                  Xiang Chen and
                  Tong Yu and
                  Saayan Mitra and
                  Victor Bursztyn and
                  Ryan A. Rossi and
                  Somdeb Sarkhel and
                  Chao Zhang},
  title        = {ToolChain*: Efficient Action Space Navigation in Large Language Models
                  with A* Search},
  journal      = {CoRR},
  volume       = {abs/2310.13227},
  year         = {2023},
  url          = {https://doi.org/10.48550/arXiv.2310.13227},
  doi          = {10.48550/ARXIV.2310.13227},
  eprinttype    = {arXiv},
  eprint       = {2310.13227},
  bibsource    = {dblp computer science bibliography, https://dblp.org}
}

@book{euzenat2007b,
        author          =       {J\'er\^ome Euzenat and Pavel Shvaiko},
        title           =       {Ontology matching},
        no_language        =       {english},
        page            =       341,
        publisher       =       {Springer},
        no_address         =       {Heidelberg (DE)},
        year            =       2007,
        no_isbn            =       {3-540-49611-4}}

@book{Ehrig07,
  no_address = {Berlin},
  author = {Marc Ehrig},
  no_interhash = {f296f88c9bd60a4a804bbcfbd1bc8d82},
  no_intrahash = {c496a12e80eb3820227e14bfbb98c75e},
  publisher = {Springer},
  series = {Semantic Web and Beyond},
  title = {Ontology Alignment: Bridging the Semantic Gap},
  volume = 4,
  year = 2007,
  no_keywords = {ai springer v1002 book semantic web},
  no_file = {Amazon Search inside:http\://www.amazon.de/gp/reader/038732805X/:URL},
  no_isbn = {978-0-387-32805-8},
  no_biburl = {http://www.bibsonomy.org/bibtex/2c496a12e80eb3820227e14bfbb98c75e/flint63},
  no_doi = {10.1007/978-0-387-36501-5},
  no_abstract = {A large number of information systems use many different individual schemas to represent data. Semantically linking these schemas is a necessary precondition to establish interoperability between agents and services. Consequently, ontology alignment and mapping for data integration has become central to building a world-wide semantic web. The book introduces novel methods and approaches for semantic integration. In addition to developing new methods for ontology alignment, the author provides extensive explanations of up-to-date case studies. The topic of this book, coupled with the application-focused methodology, will appeal to professionals from a number of different domains.}
}

@inproceedings{Blomqvist09,
  author    = {Eva Blomqvist},
  title     = {OntoCase-Automatic Ontology Enrichment Based on Ontology
               Design Patterns},
  booktitle = {International Semantic Web Conference},
  year      = {2009},
  pages     = {65-80},
  no_ee        = {http://dx.doi.org/10.1007/978-3-642-04930-9_5},
  no_crossref  = {DBLP:conf/semweb/2009},
  no_bibsource = {DBLP, http://dblp.uni-trier.de}
}

@inproceedings{RendleS06,
  author    = {Steffen Rendle and
               Lars Schmidt-Thieme},
  title     = {Object Identification with Constraints},
  year      = {2006},
  no_pages     = {1026-1031},
  booktitle     = {International Conference on Data Mining},
  publisher = {IEEE}, 
}

@inproceedings{YagoWWW07,
 author = {Suchanek, Fabian M. and Kasneci, Gjergji and Weikum, Gerhard},
 title = {Yago: a core of semantic knowledge},
 booktitle = {International Conference on World Wide Web},
 year = {2007},
 no_pages = {697--706},
 publisher = {ACM},
}

@inproceedings{HalpinP09,
  author    = {Harry Halpin and
               Valentina Presutti},
  title     = {An Ontology of Resources: Solving the Identity Crisis},
  booktitle = {European Semantic Web Conference},
  year      = {2009},
  pages     = {521-534},
  ee        = {http://dx.doi.org/10.1007/978-3-642-02121-3_39},
}

@inproceedings{PallikaEtAlCoreference2007,
 author = {Kanani, Pallika and McCallum, Andrew and Pal, Chris},
 title = {Improving author coreference by resource-bounded information gathering from the web},
 booktitle = {Conference on Artifical intelligence},
 year = {2007},
 no_pages = {429--434},
 publisher = {Morgan Kaufmann},
}

@inproceedings{WickEtAlCoreference,
  author    = {Michael L. Wick and
               Aron Culotta and
               Khashayar Rohanimanesh and
               Andrew McCallum},
  title     = {An Entity Based Model for Coreference Resolution},
  booktitle = {SIAM International Conference on Data Mining},
  year      = {2009},
  pages     = {365-376},
  ee        = {http://www.siam.org/proceedings/datamining/2009/dm09_036_wickm.pdf},
  bibsource = {DBLP, http://dblp.uni-trier.de}
}

@inproceedings{WickKDD2008,
  author = {Wick, Michael L. and Rohanimanesh, Khashayar and Schultz, Karl and McCallum, Andrew},
  title = {A unified approach for schema matching, coreference and canonicalization},
  booktitle = {International Conference on Knowledge Discovery and Data Mining},
  year = {2008},
  no_pages = {722--730},
  publisher = {ACM},
}

@inproceedings{GlaserLOD2009,
 author = {Glaser, H. and Jaffri, A. and Millard, I.},
 title = {Managing Co-reference on the Semantic Web},
 booktitle = {WWW2009 Workshop: Linked Data on the Web},
 year = {2009},
}

@INPROCEEDINGS{DBLP:conf/semweb/WilkinsonSKR03,
  author = {Kevin Wilkinson and Craig Sayers and Harumi A. Kuno and Dave Reynolds},
  title = {Efficient {RDF} Storage and Retrieval in {Jena}2},
  booktitle = {International Workshop on Semantic Web and Databases},
  year = {2003},
  no_pages = {131-150},
}

@ARTICLE{DBLP:journals/ws/ArtzG07,
  author = {Donovan Artz and Yolanda Gil},
  title = {{A Survey of Trust in Computer Science and the Semantic Web}},
  journal = {J. Web Sem.},
  year = {2007},
  volume = {5},
  pages = {58-71},
  number = {2},
  bibsource = {DBLP, http://dblp.uni-trier.de},
  ee = {http://dx.doi.org/10.1016/j.websem.2007.03.002}
}

@INPROCEEDINGS{key:dbpedia,
  author = {Auer, S. and Bizer, C. and Kobilarov, G. and Lehmann, J. and Cyganiak,
	R. and Ives, Z.},
  title = {{DBpedia: A Nucleus for a Web of Open Data}},
  booktitle = {Semantic Web Conference and Asian Semantic Web Conference},
  year = {2008},
  pages = {722--735},
  month = {November},
  doi = {10.1007/978-3-540-76298-0_52},
  journal = {The Semantic Web},
  location = {Busan, Korea},
  url = {http://dx.doi.org/10.1007/978-3-540-76298-0_52}
}

@TECHREPORT{rfc1630,
  author = {T. Berners-Lee},
  title = {{Universal Resource Identifiers in WWW: A Unifying Syntax for the
	Expression of Names and Addresses of Objects on the Network as used
	in the World-Wide Web}},
  institution = {Internet Engineering Task Force},
  year = {1994},
  type = {RFC},
  number = {1630},
  month = jun,
  days = {15},
  pages = {28},
  url = {http://www.rfc-editor.org/rfc/rfc1630.txt}
}

@TECHREPORT{rfc1738,
    series =    {Request for Comments},
    number =    1738,
    howpublished =  {RFC 1738},
    publisher = {RFC Editor},
    doi =       {10.17487/RFC1738},
    url =       {https://www.rfc-editor.org/info/rfc1738},
    author =    {Tim Berners-Lee and Larry M Masinter and Mark P. McCahill},
    title =     {{Uniform Resource Locators (URL)}},
    pagetotal = 25,
    year =      1994,
    month =     dec,
}

@TECHREPORT{rfc3986,
    series =    {Request for Comments},
    number =    3986,
    howpublished =  {RFC 3986},
    publisher = {RFC Editor},
    doi =       {10.17487/RFC3986},
    url =       {https://www.rfc-editor.org/info/rfc3986},
        author =    {Tim Berners-Lee and Roy T. Fielding and Larry M Masinter},
    title =     {{Uniform Resource Identifier (URI): Generic Syntax}},
    pagetotal = 61,
    year =      2005,
    month =     jan,
}

@ARTICLE{DBLP:journals/expert/Bizer09,
  author = {Christian Bizer},
  title = {{The Emerging Web of Linked Data}},
  journal = {IEEE Intelligent Systems},
  year = {2009},
  volume = {24},
  pages = {87-92},
  number = {5},
  bibsource = {DBLP, http://dblp.uni-trier.de},
  ee = {http://doi.ieeecomputersociety.org/10.1109/MIS.2009.102}
}

@INPROCEEDINGS{iswc-Broekstra02sesame,
  author = {Jeen Broekstra and Arjohn Kampman and Frank Van Harmelen},
  title = {{Sesame: A Generic Architecture for Storing and Querying RDF and
	RDF Schema}},
  booktitle = {International Semantic Web Conference},
  year = {2002},
  pages = {54--68},
  publisher = {Springer}
}

@ARTICLE{DBLP:journals/ki/DividinoSSS09,
  author = {Renata Queiroz Dividino and Simon Schenk and Sergej Sizov and Steffen
	Staab},
  title = {{Provenance, Trust, Explanations - and all that other Meta Knowledge}},
  journal = {KI},
  year = {2009},
  volume = {23},
  pages = {24-30},
  number = {2},
  bibsource = {DBLP, http://dblp.uni-trier.de},
  ee = {http://www.kuenstliche-intelligenz.de/fileadmin/template/main/archiv/pdf/ki2009-02_page24_web_teaser.pdf}
}

@TECHREPORT{rfc3987,
  author = {M. Duerst and M. Suignard},
  title = {Internationalized Resource Identifiers {(IRIs)}},
  institution = {Internet Engineering Task Force},
  year = {2005},
  type = {RFC},
  number = {3987},
  month = jan,
  url = {http://www.rfc-editor.org/rfc/rfc3987.txt}
}

@article{DBLP:journals/jacm/GeertsUKFC16,
  author       = {Floris Geerts and
                  Thomas Unger and
                  Grigoris Karvounarakis and
                  Irini Fundulaki and
                  Vassilis Christophides},
  title        = {Algebraic Structures for Capturing the Provenance of {SPARQL} Queries},
  journal      = {J. {ACM}},
  volume       = {63},
  number       = {1},
  pages        = {7:1--7:63},
  year         = {2016},
  url          = {https://doi.org/10.1145/2810037},
  doi          = {10.1145/2810037},
  bibsource    = {dblp computer science bibliography, https://dblp.org}
}

@article{DBLP:journals/pvldb/HernandezGH21,
  author       = {Daniel Hern{\'{a}}ndez and
                  Luis Gal{\'{a}}rraga and
                  Katja Hose},
  title        = "{Computing How-Provenance for {SPARQL} Queries via Query Rewriting}",
  journal      = {Proc. {VLDB} Endow.},
  volume       = {14},
  number       = {13},
  pages        = {3389--3401},
  year         = {2021},
  url          = {http://www.vldb.org/pvldb/vol14/p3389-galarraga.pdf},
  doi          = {10.14778/3484224.3484235},
  bibsource    = {dblp computer science bibliography, https://dblp.org}
}

@INPROCEEDINGS{DBLP:conf/semweb/FlourisFPTC09,
  author = {Giorgos Flouris and Irini Fundulaki and Panagiotis Pediaditis and
	Yannis Theoharis and Vassilis Christophides},
  title = {{Coloring RDF Triples to Capture Provenance}},
  booktitle = {International Semantic Web Conference},
  year = {2009},
  volume = {5823},
  series = {LNCS},
  pages = {196--212},
  publisher = {Springer},
  isbn = {978-3-642-04929-3},
  location = {Heidelberg},
  no_editor = {Abraham Bernstein and David R. Karger and Tom Heath and Lee Feigenbaum
	and Diana Maynard and Enrico Motta and Krishnaprasad Thirunarayan}
}

@INPROCEEDINGS{DBLP:conf/esws/GavriloaieNOSW04,
  author = {Rita Gavriloaie and Wolfgang Nejdl and Daniel Olmedilla and Kent
	E. Seamons and Marianne Winslett},
  title = {{No Registration Needed: How to Use Declarative Policies and Negotiation
	to Access Sensitive Resources on the Semantic Web}},
  booktitle = {European Semantic Web Symposium},
  year = {2004},
  volume = {3053},
  series = {LNCS},
  pages = {342-356},
  publisher = {Springer},
  bibsource = {DBLP, http://dblp.uni-trier.de},
  ee = {http://springerlink.metapress.com/openurl.asp?genre=article{\&}issn=0302-9743{\&}volume=3053{\&}spage=342}
}

@TECHREPORT{key:owl2,
  no_author = {},
  title = {{{OWL} 2 Web Ontology Language Document Overview }},
  institution = {W3C},
  year = {2009},
  type = {{W3C} Recommendation},
  month = { October },
  note = {http://www.w3.org/TR/owl2-overview/},
  day = {27}
}

@ARTICLE{DBLP:journals/ai/HeinsohnKNP94,
  author = {Jochen Heinsohn and Daniel Kudenko and Bernhard Nebel and Hans-J{\"u}rgen
	Profitlich},
  title = {{An Empirical Analysis of Terminological Representation Systems}},
  journal = {Artif. Intell.},
  year = {1994},
  volume = {68},
  pages = {367-397},
  number = {2},
  bibsource = {DBLP, http://dblp.uni-trier.de}
}

@ARTICLE{DBLP:journals/insk/JanikSS11,
  author = {Maciej Janik and Ansgar Scherp and Steffen Staab},
  title = {{The Semantic Web: Collective Intelligence on the Web}},
  journal = {Informatik Spektrum},
  year = {2011},
  volume = {34},
  pages = {469-483},
  number = {5},
  bibsource = {DBLP, http://dblp.uni-trier.de},
  ee = {http://dx.doi.org/10.1007/s00287-011-0535-x}
}

@INPROCEEDINGS{DBLP:conf/cikm/KasneciEW09,
  author = {Gjergji Kasneci and Shady Elbassuoni and Gerhard Weikum},
  title = {{MING: Mining Informative Entity Relationship Subgraphs}},
  booktitle = {Information and Knowledge Management},
  year = {2009},
  no_pages = {1653-1656},
  publisher = {ACM},
  bibsource = {DBLP, http://dblp.uni-trier.de},
  ee = {http://doi.acm.org/10.1145/1645953.1646196}
}

@INPROCEEDINGS{key:oem,
  author = {Papakonstantinou, Yannis and Garcia-Molina, Hector and Widom, Jennifer},
  title = {{Object Exchange Across Heterogeneous Information Sources}},
  booktitle = {Data Engineering},
  year = {1995},
  pages = {251--260},
  address = {Washington, DC, USA},
  publisher = {IEEE Computer Society},
  isbn = {0-8186-6910-1}
}

@ARTICLE{Raimond2009IMR,
  author = {Yves Raimond and Christopher Sutton and Mark B. Sandler},
  title = {{Interlinking Music-Related Data on the Web}},
  journal = {IEEE MultiMedia},
  year = {2009},
  volume = {16},
  pages = {52-63},
  number = {2},
  bibsource = {DBLP, http://dblp.uni-trier.de},
  ee = {http://doi.ieeecomputersociety.org/10.1109/MMUL.2009.29}
}

@ARTICLE{DBLP:journals/jair/StoilosSPTH07,
  author = {Giorgos Stoilos and Giorgos B. Stamou and Jeff Z. Pan and Vassilis
	Tzouvaras and Ian Horrocks},
  title = {{Reasoning with Very Expressive Fuzzy Description Logics}},
  journal = {J. Artif. Intell. Res.},
  year = {2007},
  volume = {30},
  pages = {273-320},
  bibsource = {DBLP, http://dblp.uni-trier.de},
  ee = {http://www.jair.org/papers/paper2279.html}
}

@INPROCEEDINGS{key:sigma,
  author = {Giovanni Tummarello and Richard Cyganiak and Michele Catasta and
	Szymon Danielczyk and Stefan Decker},
  title = {{Sig.ma: live views on the Web of Data}},
  booktitle = {Semantic Web Challenge 2009 at the 8th International Semantic Web
	Conference (ISWC2009)},
  year = {2009}
}

@PROCEEDINGS{DBLP:conf/dlog/2003handbook,
  title = {{The Description Logic Handbook: Theory, Implementation, and Applications}},
  year = {2003},
  editor = {Franz Baader and Diego Calvanese and Deborah L. McGuinness and Daniele
	Nardi and Peter F. Patel-Schneider},
  publisher = {Cambridge University Press},
  bibsource = {DBLP, http://dblp.uni-trier.de},
  booktitle = {Description Logic Handbook},
  isbn = {0-521-78176-0}
}

@MISC{W3CR2RML,
  author = {Souripriya Das and Seema Sundara and Richard Cyganiak},
  title = {{R2RML: RDB to RDF Mapping Language}},
  year = {2012},
  note = {https://www.w3.org/TR/r2rml/},
}

@MISC{W3TURTLE, 
  author = {David Beckett and Tim Berners-Lee and Eric Prud'hommeaux and Gavin Carothers},
  title = {{Terse RDF Triple Language}},
  year = {2014},
  note = {http://www.w3.org/TR/turtle/}
}

@inproceedings{DBLP:conf/www/TanonVSSP16,
  author    = {Thomas Pellissier Tanon and
               Denny Vrandecic and
               Sebastian Schaffert and
               Thomas Steiner and
               Lydia Pintscher},
  title     = {From Freebase to Wikidata: The Great Migration},
  booktitle = {International Conference on World Wide Web},
  no_pages     = {1419--1428},
  year      = {2016},
  publishr = {ACM},
}

@article{DBLP:journals/cacm/VrandecicK14,
  author    = {Denny Vrandecic and
               Markus Kr{\"{o}}tzsch},
  title     = {Wikidata: a free collaborative knowledgebase},
  journal   = {Commun. {ACM}},
  volume    = {57},
  number    = {10},
  pages     = {78--85},
  year      = {2014},
  url       = {https://doi.org/10.1145/2629489},
  doi       = {10.1145/2629489},
  bibsource = {dblp computer science bibliography, https://dblp.org}
}

@article{DBLP:journals/ki/GlimmS16a,
  author    = {Birte Glimm and
               Heiner Stuckenschmidt},
  title     = {15 Years of Semantic Web: An Incomplete Survey},
  journal   = {{KI}},
  volume    = {30},
  number    = {2},
  pages     = {117--130},
  year      = {2016},
  url       = {https://doi.org/10.1007/s13218-016-0424-1},
  doi       = {10.1007/s13218-016-0424-1},
  bibsource = {dblp computer science bibliography, https://dblp.org}
}

@inproceedings{DBLP:conf/semweb/MotikSS04,
  author    = {Boris Motik and
               Ulrike Sattler and
               Rudi Studer},
  title     = {Query Answering for {OWL-DL} with Rules},
  booktitle = {International Semantic Web Conference},
  no_pages     = {549--563},
  year      = {2004},
  publisher = {Springer},
}

@article{DBLP:journals/jar/GlimmHMSW14,
  author    = {Birte Glimm and
               Ian Horrocks and
               Boris Motik and
               Giorgos Stoilos and
               Zhe Wang},
  title     = {HermiT: An {OWL} 2 Reasoner},
  journal   = {J. Autom. Reasoning},
  volume    = {53},
  number    = {3},
  pages     = {245--269},
  year      = {2014},
  url       = {https://doi.org/10.1007/s10817-014-9305-1},
  doi       = {10.1007/s10817-014-9305-1},
  bibsource = {dblp computer science bibliography, https://dblp.org}
}

@Inbook{Decker1999,
  author="Decker, Stefan and Erdmann, Michael and Fensel, Dieter and Studer, Rudi",
  chapter="Ontobroker: Ontology Based Access to Distributed and Semi-Structured Information",
  title="Database Semantics: Semantic Issues in Multimedia Systems",
  year="1999",
  publisher="Springer",
  no_pages="351--369",
}

@InProceedings{10.1007/3-540-48005-6_18,
  author="Sure, York and Erdmann, Michael and Angele, Juergen and Staab, Steffen and Studer, Rudi and Wenke, Dirk",
  title="{OntoEdit: Collaborative Ontology Development for the Semantic Web}",
  booktitle="International Semantic Web Conference",
  year="2002", 
  publisher="Springer",
  no_pages="221--235",
}

@article{DBLP:journals/tods/PerezAG09,
  author       = {Jorge P{\'{e}}rez and
                  Marcelo Arenas and
                  Claudio Gutierrez},
  title        = {Semantics and complexity of {SPARQL}},
  journal      = {{ACM} Trans. Database Syst.},
  volume       = {34},
  number       = {3},
  pages        = {16:1--16:45},
  year         = {2009},
  url          = {https://doi.org/10.1145/1567274.1567278},
  doi          = {10.1145/1567274.1567278}
}

@inproceedings{DBLP:conf/i-semantics/GalkinEACVA17,
  author       = {Mikhail Galkin and
                  Kemele M. Endris and
                  Maribel Acosta and
                  Diego Collarana and
                  Maria{-}Esther Vidal and
                  S{\"{o}}ren Auer},
  editor       = {Rinke Hoekstra and
                  Catherine Faron{-}Zucker and
                  Tassilo Pellegrini and
                  Victor de Boer},
  title        = {SMJoin: {A} Multi-way Join Operator for {SPARQL} Queries},
  booktitle    = {Proceedings of the 13th International Conference on Semantic Systems,
                  SEMANTiCS 2017, Amsterdam, The Netherlands, September 11-14, 2017},
  pages        = {104--111},
  publisher    = {{ACM}},
  year         = {2017},
  url          = {https://doi.org/10.1145/3132218.3132220},
  doi          = {10.1145/3132218.3132220},
  
}

@article{DBLP:journals/corr/abs-2308-10168,
  author       = {Kai Sun and
                  Yifan Ethan Xu and
                  Hanwen Zha and
                  Yue Liu and
                  Xin Luna Dong},
  title        = {Head-to-Tail: How Knowledgeable are Large Language Models (LLM)? {A.K.A.}
                  Will LLMs Replace Knowledge Graphs?},
  journal      = {CoRR},
  volume       = {abs/2308.10168},
  year         = {2023},
  url          = {https://doi.org/10.48550/arXiv.2308.10168},
  doi          = {10.48550/arXiv.2308.10168}
}

@inproceedings{DBLP:conf/sigmod/BollackerEPST08-freebase,
  author       = {Kurt D. Bollacker and
                  Colin Evans and
                  Praveen K. Paritosh and
                  Tim Sturge and
                  Jamie Taylor},
  editor       = {Jason Tsong{-}Li Wang},
  title        = {Freebase: a collaboratively created graph database for structuring
                  human knowledge},
  booktitle    = {Proceedings of the {ACM} {SIGMOD} International Conference on Management
                  of Data, {SIGMOD} 2008, Vancouver, BC, Canada, June 10-12, 2008},
  pages        = {1247--1250},
  publisher    = {{ACM}},
  year         = {2008},
  url          = {https://doi.org/10.1145/1376616.1376746},
  doi          = {10.1145/1376616.1376746},
  bibsource    = {dblp computer science bibliography, https://dblp.org}
}

@inproceedings{DBLP:conf/aaai/BollackerCT07-freebase,
  author       = {Kurt D. Bollacker and
                  Robert P. Cook and
                  Patrick Tufts},
  title        = {Freebase: {A} Shared Database of Structured General Human Knowledge},
  booktitle    = {Proceedings of the Twenty-Second {AAAI} Conference on Artificial Intelligence,
                  July 22-26, 2007, Vancouver, British Columbia, Canada},
  pages        = {1962--1963},
  publisher    = {{AAAI} Press},
  year         = {2007},
  url          = {http://www.aaai.org/Library/AAAI/2007/aaai07-355.php},
  bibsource    = {dblp computer science bibliography, https://dblp.org}
}

\end{document}